\documentclass{article} 
\usepackage{iclr2020_conference,times}
\usepackage{hyperref}
\usepackage{url}

\usepackage{from_neurips}

\usepackage{amsmath,amsfonts,bm}

\def\eqref#1{equation~\ref{#1}}

\def\1{\bm{1}}

\DeclareMathAlphabet{\mathsfit}{\encodingdefault}{\sfdefault}{m}{sl}
\SetMathAlphabet{\mathsfit}{bold}{\encodingdefault}{\sfdefault}{bx}{n}

\usepackage{todonotes}

\newcommand{\hide}[1]{}

\title{NAS-Bench-1Shot1: \\Benchmarking and Dissecting \\One-shot Neural Architecture Search}

\author{Arber Zela$^1$$^*$, Julien Siems$^{1}$\thanks{Equal contribution} , \& Frank Hutter$^{1,2}$ \\
$^1$Department of Computer Science, University of Freiburg\\
$^2$Bosch Center for Artificial Intelligence\\
\texttt{\{zelaa, siemsj, fh\}@cs.uni-freiburg.de}
}

\iclrfinalcopy

\begin{document}

\maketitle

\begin{abstract}
One-shot neural architecture search (NAS) has played a crucial role in making NAS methods computationally feasible in practice.
Nevertheless, there is still a lack of understanding on how these weight-sharing algorithms exactly work due to the many factors controlling the dynamics of the process.
%
In order to allow a scientific study of these components, 
we introduce a general framework for one-shot NAS that can be instantiated to many recently-introduced variants and introduce a general benchmarking framework that draws on the recent large-scale tabular benchmark NAS-Bench-101 for cheap anytime evaluations of one-shot NAS methods. 
%
To showcase the framework, we compare several state-of-the-art one-shot NAS methods, examine how sensitive they are to their hyperparameters and how they can be improved by tuning their hyperparameters, and compare their performance to that of blackbox optimizers for NAS-Bench-101.
\end{abstract}
\section{Introduction}
\label{sec: introduction}

While neural architecture search (NAS) has attracted a lot of attention due to the effectiveness in automatically designing state-of-the-art neural networks~\citep{zoph-iclr17, zoph-arXiv18, real17a, real-arXiv18a}, the focus has recently shifted to making the search process more efficient~\citep{Pham18,Elsken19,darts,Xie18,Cai19,casale2019}. 
The most crucial concept which led to a reduction in search costs to the order of a single function evaluation is certainly the weight-sharing paradigm: Training only a single large architecture (the \textit{one-shot} model) subsuming all the possible architectures in the search space~\citep{brock2018smash,Pham18}.

Despite the great advancements of these methods, the exact results of many NAS papers are often hard to reproduce~\citep{li2019random,sciuto19,Yang2020NAS}. This is a result of several factors, such as unavailable original implementations, differences in the employed search spaces, training or evaluation pipelines, hyperparameter settings, and even pseudorandom number seeds~\citep{best_practices}.
%
One solution to guard against these problems would be a common library of NAS methods that provides primitives to construct different algorithm variants, similar to what as RLlib~\citep{rllib} offers for the field of reinforcement learning. Our paper makes a first step into this direction.

Furthermore, experiments in NAS can be computationally extremely costly, making it virtually impossible to perform proper scientific evaluations with many repeated runs to draw statistically robust conclusions.
To address this issue, \citet{ying19a} introduced NAS-Bench-101, a large tabular benchmark with 423k unique 
cell architectures, trained and fully evaluated using a one-time extreme amount of compute power (several months on thousands of TPUs), which now allows to cheaply simulate an arbitrary number of runs of NAS methods, even on a laptop. NAS-Bench-101 enabled a comprehensive benchmarking of many discrete NAS optimizers~\citep{zoph-iclr17, real-arXiv18a}, using the exact same settings. However, the discrete nature of this benchmark does not allow to directly benchmark one-shot NAS optimizers~\citep{Pham18, darts, Xie18, Cai19}.
In this paper, we introduce the first method for making this possible.

Specifically, after providing some background (Section \ref{sec:background}), we make the following contributions:

\begin{enumerate}[leftmargin=*]
    \item  We introduce \emph{NAS-Bench-1Shot1}, 
    a novel benchmarking framework that allows us to reuse the extreme amount of compute time that went into generating NAS-Bench-101~\citep{ying19a} to cheaply benchmark one-shot NAS methods. Our mapping between search space representations is novel to the best of our knowledge and it allows querying the performance of found architectures from one-shot NAS methods, contrary to what is claimed by \citet{ying19a}.
    Specifically, it allows us to follow the full trajectory of architectures found by arbitrary one-shot NAS methods at each search epoch without the need for retraining them individually, allowing for a careful and statistically sound analysis (Section\mbox{ \ref{sec:methodology}}).
    \item We introduce a general framework for one-shot NAS methods that can be instantiated to many recent one-shot NAS variants, enabling fair head-to-head evaluations based on a single code base (Section \ref{sec:framework}).
    \item We use the above to compare several state-of-the-art one-shot NAS methods,
    assess the correlation between their one-shot model performance and final test performance,
    examine how sensitive they are to their hyperparameters,
    and compare their performance to that of black-box optimizers used in NAS-Bench-101 (Section \ref{sec:experiments}).
\end{enumerate}

We provide our open-source implementation\footnote{\href{https://github.com/automl/nasbench-1shot1}{https://github.com/automl/nasbench-1shot1}}, which we expect will also facilitate the reproducibility and benchmarking of other one-shot NAS methods in the future.
\section{Background and Related Work}
\label{sec:background}

\subsection{NAS-Bench-101}\label{sec: NAS-Bench}
NAS-Bench-101 \citep{ying19a} is a database of an exhaustive evaluation of all architectures in a constrained cell-structured space on CIFAR-10 \citep{krizhevsky-tech09a}. Each cell is represented as a directed acyclic graph (DAG) where the nodes represent operation choices and the edges represent the information flow through the neural network (see also Figure \ref{fig:sec2:overview_graphic} and Section \ref{subsec: search_spaces}). To limit the number of architectures in the search space, the authors used the following constraints on the cell: 3 operations in the operation set $\mathcal{O}$ = \{3x3 convolution, 1x1 convolution,  3x3 max-pool\}, at most 7 nodes (this includes input and output node, therefore 5 choice nodes) and at most 9 edges.

These constraints, and exploiting symmetries, reduced the search space to 423k unique valid architectures. Each architecture was trained from scratch three times to also obtain a measure of variance. In addition, each architecture was trained for 4, 12, 36 and 108 epochs; for our analysis, we mainly used the results for models trained for 108 epochs, if not stated otherwise.


\subsection{NAS-Bench-102}\label{sec: NAS-Bench-102}
Concurrently to this work, ~\citet{dong20} released NAS-Bench-102, which is another NAS benchmark that, differently from NAS-Bench-101, enables the evaluation of weight-sharing NAS methods. Their search space consists of a total of 15,625 architectures, which is exhaustively evaluated on 3 image classification datasets. Similarly to \citet{zela19} and this work, \citet{dong20} found that architectural overfitting occurs for DARTS for all their datasets.

While NAS-Bench-102 and this work go towards the same direction, they differ in many ways:
\begin{enumerate}[leftmargin=*]
    \item They use extensive computation to create a \emph{new} benchmark (with 15.625 architectures), while we devise a novel reformulation to reuse the even much more extensive computation of the NAS-Bench-101 dataset (~120 TPU years) to create three new one-shot search spaces with the larges one containing 363.648 architectures. This required zero additional computational cost.
    \item We show that it \emph{is} possible to reuse the graph representation in NAS-Bench-101 to run one-shot NAS methods; this requires changes to the one-shot search space, but allows a mapping which can be used for architecture evaluation.
    \item They evaluate their search space on 3 image classification datasets, while we introduce 3 different search spaces (as sub-spaces of NAS-Bench-101) with growing complexity.
\end{enumerate}

\subsection{One-shot Neural Architecture Search}\label{subsec: oneshotNAS}
The NAS problem can be defined as searching for the optimal operation (e.g. in terms of validation error of architectures) out of the operation set $\mathcal{O}$ in each node of the DAG and for the best connectivity pattern between these nodes.

Designing architectures for optimized accuracy or to comply with resource constraints led to significant breakthroughs on many standard benchmarks \citep{Pham18, zoph-iclr17, brock2018smash, darts, Cai19, Elsken19}. 
While early methods were computationally extremely expensive \citep{zoph-iclr17}, the weight-sharing paradigm~\citep{brock2018smash,Pham18} led to a significant increase in search efficiency. Here, the weights of the operations in each architecture are shared in a supermodel (the so-called \emph{one-shot model} or convolutional neural fabric~\citep{SaxenaV16}), 
which contains an exponential number of sub-networks, each of which represents a discrete architecture. Architectures whose sub-networks share components (nodes/edges)
also share the weights for these components' operations; therefore, in analogy to DropOut~\citep{srivastava-jmlr14a}, training one architecture implicitly also trains (parts of) an exponential number of related architectures.
%
There are a variety of methods on how to conduct NAS by means of the one-shot model~\citep{brock2018smash, Pham18, bender_icml:2018, darts, li2019random} (see also Appendix \ref{sec:app:optimizers}), but the final problem is to find the optimal sub-network in this one-shot model.

The weight sharing method was used to great effect in DARTS~\citep{darts}, where it allows a gradient based optimization of both the architectural and the one-shot weights. Subsequent work on DARTS has addressed further lowering the computational and the memory requirements \citep{dong2019search, xu2019pcdarts, Cai19, casale2019}.

One fundamental drawback of the weight sharing method is the fact that the architecture search typically takes place in a lower fidelity model (e.g., using less cells and/or cheaper operations): the so-called \emph{proxy model}. After the search, a discrete architecture is derived from the proxy model which is then trained with more parameters --- a stage often referred to as \emph{architecture evaluation}. This poses the question whether the architecture found in the proxy model is also a good architecture in the bigger model,
a question studied by several recent works \citep{bender_icml:2018, sciuto19}.

\section{A general framework for benchmarking one-shot NAS}
\label{sec:methodology}

\begin{figure}[t]
\centering
\includegraphics[width=1.0\textwidth]{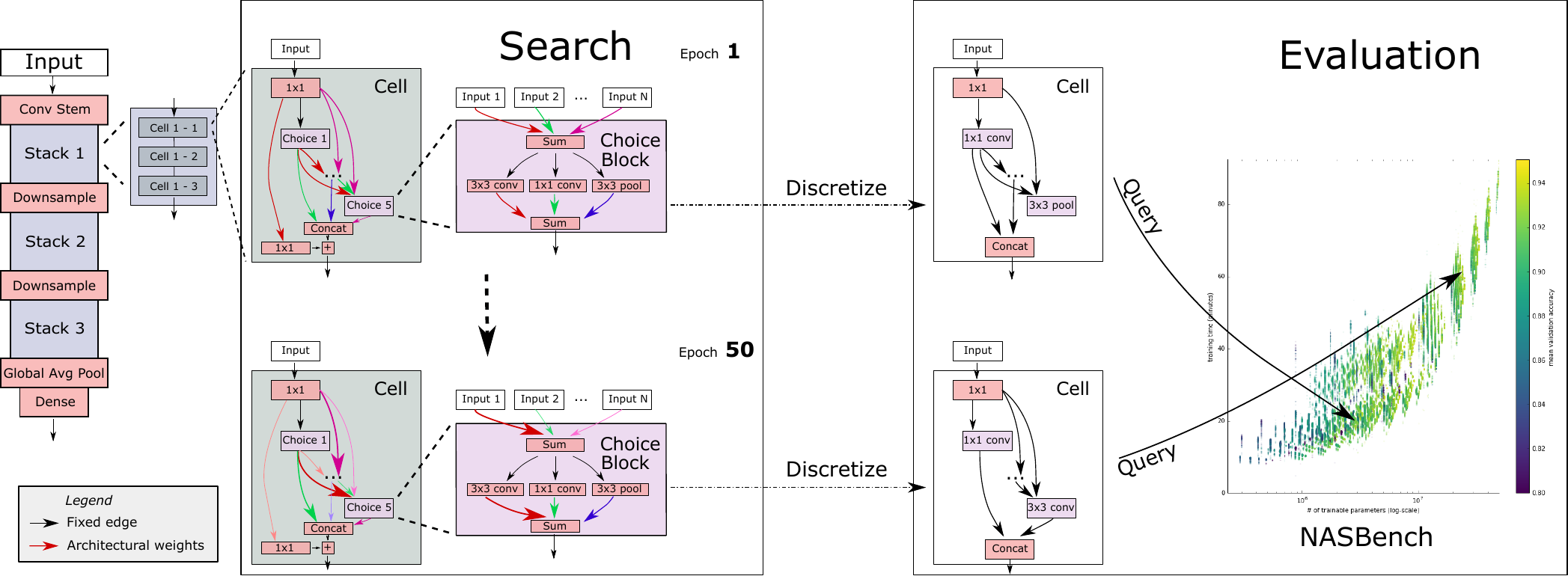}
\caption{Overview of the NAS-Bench-1Shot1 analysis strategy. 
The one-shot model we construct only contains discrete architectures that are elements of NAS-Bench-101~\citep{ying19a}.
The cell architecture chosen is similar to that of \citet{bender_icml:2018}, with each choice block containing an operation decision. Note that NAS-Bench-101 does not contain a separate reduction cell type. Plot on the right from~\citet{ying19a} (Best viewed in color).
}
\label{fig:sec2:overview_graphic}
\end{figure}

We will now introduce our framework for cheaply benchmarking the anytime performance of one-shot NAS methods.
Our main analysis strategy is the following: First, we run the search procedure of various methods and save the architecture weights of the one-shot models for each epoch. Second, we find the discrete architecture at each epoch and query it in NAS-Bench-101. The last step is not trivial due to the different representations of the search space used in NAS-Bench-101 and standard one-shot methods. \citet{ying19a} state that one-shot methods cannot be directly evaluated on NAS-Bench-101. In the following sections we present a mapping between these different search space representations, which eventually enable us to evaluate one-shot methods on NAS-Bench-101. To the best of our knowledge this is a novel contribution of this paper.

\subsection{Search space representation}
\label{subsec: search_spaces}

In order to carry out the analysis we propose in this work, we had to construct a search space that only contains discrete architectures that are also contained in NAS-Bench-101. This allows us to look up any discrete architectures' performance in NAS-Bench-101 when the larger model is trained from scratch. Unfortunately, this is non-trivial since the NAS-Bench-101 space does not match the typical space used in one-shot NAS methods. We separately consider the various parts of the search space.

\textbf{Network-Level Topology.} In terms of network-level topology, our search spaces closely resemble the models which were evaluated in NAS-Bench-101. We used the same macro architecture as in NAS-Bench-101, i.e., 3 stacked blocks with a max-pooling operation in-between, where each block consists of 3 stacked cells (see Figure~\ref{fig:sec2:overview_graphic}).
While our final evaluation models exactly follow NAS-Bench-101 in order to be able to reuse its evaluations, our one-shot model only has 16 initial convolution filters, rather than the 128 used in NAS-Bench-101. This is a common practice to accelerate NAS and used similarly in, e.g., \citet{darts}. 

\textbf{Cell-Level Topology.} The cell-level structure is represented as a DAG, where the input node is the output of a previous cell or the convolutional stem (projected using a $1\times 1$ convolution in order to scale the channel counts), and the output node consists of firstly concatenating the outputs of all intermediate nodes and adding to that the projected output of the input node. In order to have the operation choices still in the intermediate nodes of the DAG, we adapt the \textit{choice block} motif from \citet{bender_icml:2018} as depicted in Figure~\ref{fig:sec2:overview_graphic}. The edges connecting input, output nodes and choice blocks represent only the information flow in the graph. To have a large and expressive enough search space(s), we introduce the following architectural weights in the DAG edges:
\begin{itemize}[leftmargin=*]
    \item $\alpha^{i,j}$ to edges connecting nodes $i<j$ to choice block $j$. The input of choice block $j$ is then computed as $I^j = \sum_{i<j} \frac{\exp(\alpha^{i,j})}{\sum_{i^{\prime} < j}\exp(\alpha^{i^{\prime},j})} x^i$, where $x^i$ is the output tensor of node $i$ (either input node or choice block).
    \item $\gamma^{j,k}$ to the edges connecting the input node or choice blocks $j<k$ to the output node $k$ of the cell, where the corresponding output feature maps of the choice blocks are concatenated and then added to the projected input of the cell: $O^k = \frac{\exp(\gamma^{1,k})}{\sum_{j^{\prime} < k}\exp(\gamma^{j^{\prime},k})} x^1 + \oplus_{1<j<k} \frac{\exp(\gamma^{j,k})}{\sum_{j^{\prime} < k}\exp(\gamma^{j^{\prime},k})} x^j$, where $\oplus$ is the concatenation operator and $x^1$ the input node tensor.
\end{itemize}

Note that the non-linearity applied to the edge weights varies depending on the NAS optimizer used; e.g. for GDAS~\citep{dong2019search} and SNAS~\citep{Xie18} it would be a Gumbel-Softmax~\citep{gumbel} instead. 

\textbf{Choice Blocks.} As in \citet{bender_icml:2018}, each choice block inside the cell can select between the operations in the operations set $\mathcal{O}$ of NAS-Bench-101. In order to find the optimal operation in each choice block via gradient-based one-shot NAS methods, we assign an architectural weight $\beta^o$ to each operation $o \in \mathcal{O}$ inside the choice block. The output of the choice block $j$ is computed by adding element-wise the latent representations coming from the operations outputs: 
\begin{equation} \label{eq:block_output}
\begin{split}
x^j = \sum_{o\in\mathcal{O}}\frac{\exp(\beta^{o})}{\sum_{o^{\prime}\in\mathcal{O}}\exp(\beta^{o^{\prime}})} o(I^j),
\end{split}
\end{equation}
which is basically the so-called \textit{MixedOp} in DARTS.
NASBench cells contain 1x1 projections in front every operation (demonstrated in Figure 1 in \citep{ying19a}). The number of output channels of each projection is chosen such that the output has the same number of channels as the input. This adaptive choice for the number of channels is incompatible with the one-shot model due to the different tensor dimensionality coming from previous choice blocks. We used 1x1 projections with a fixed number of channels instead.

\subsection{Evaluation procedure}
\label{subsec: evaluation_procedure}

By means of these additional weights we do not restrict the possible architectures in the search space to contain only a fixed number of edges per cell, as done for example in \citet{zoph-arXiv18}, \citet{Pham18}, \citet{darts}, etc.
This requirement would have restricted our architectural decisions heavily, leading to only small search spaces. 

\begin{wraptable}[13]{r}{.45\textwidth}
\centering
\vskip -0.19in
\caption{Characteristic information of the search spaces.} 
\resizebox{.45\textwidth}{!}{%
\begin{tabular}{@{}llccc@{}}
\toprule
\multicolumn{1}{l}{}                                   & \multicolumn{1}{c}{\textbf{}} & \multicolumn{3}{c}{\textbf{Search space}} \\ \midrule
\multicolumn{1}{l}{}                                   &                  & \textbf{1}   & \textbf{2}   & \textbf{3} \\ \cmidrule(l){2-5} 
\multirow{6}{*}{No. parents} & Node 1                        & 1            & 1            & 1          \\
                                                       & Node 2                        & 2            & 1            & 1          \\
                                                       & Node 3                        & 2            & 2            & 1          \\
                                                       & Node 4                        & 2            & 2            & 2          \\
                                                       & Node 5                        & -            & -            & 2          \\
                                                       & Output                        & 2            & 3            & 2          \\ \midrule
\multicolumn{1}{l}{\multirow{3}{*}{No. archs.}} & w/ loose ends                        & 6240         & 29160        & 363648    \\
\multicolumn{1}{l}{}                                   & w/o loose ends                & 3702         & 12510        & 137406    \\  
\multicolumn{1}{l}{}                                   & w/o isomorphism               & 2685         & 7773         & 55854    \\ \bottomrule
\end{tabular}
}
\label{table:search_spaces}
\end{wraptable}

Table \ref{table:search_spaces} shows the characteristics of each search space. We propose three different search spaces by making different decisions on the number of parents each choice block has. The decisions affect the quality and quantity of the architectures contained in each search space. For all search spaces note that the sum of the number of parents of all nodes in the search space is chosen to be 9, to match the NAS-Bench-101 requirement.
Search space 1, 2 and 3 have 6240, 29160 and 363648 architectures with loose ends respectively, making search space 3 the largest investigated search space. To the best of our knowledge search space 3 is currently the largest and only available tabular benchmark for one-shot NAS. For details on each search space see \mbox{Appendix \ref{sec: search_spaces_details}}.

Given the architectural weights of the cell shown in Figure \ref{fig:sec2:overview_graphic} we query the test and validation error of the discrete architecture from NAS-Bench-101 as follows. 
\begin{enumerate}[leftmargin=*]
    \item We determine the operation chosen in each choice block by choosing the operation with the highest architectural weight.
    \item We determine the parents of each choice block and the output by choosing the top-$k$ edges according to Table \ref{table:search_spaces} (e.g. for choice block 4 in search space 3 we would choose the top-2 edges as parents).
    \item From 1. we construct the operation list and from 2. the adjacency matrix of the cell which we use to query NAS-Bench-101 for the test and validation error.
\end{enumerate}

Each node in the graph chooses its parents during evaluation following e.g. DARTS~\citep{darts}. However, because edges model information flow and the output edges are also architectural decisions there is possibility of a node being a loose end. These are nodes whose output does not contribute to the output of the discrete cell, as seen in the upper cell under evaluation of Figure~\ref{fig:sec2:overview_graphic}. As a result, we can count the number of architectures with or without loose ends. Note, that had we chosen the children of each node we could have invalid architectures where a node has an output but no input.

\section{A General Framework for One-shot NAS Methods}
\label{sec:framework}


Most of the follow-up works of DARTS (Algorithm \ref{alg:darts}), which focus on making the search even more efficient and effective, started from the original DARTS codebase\footnote{\url{https://github.com/quark0/darts}},
and each of them only change very few components compared to DARTS. 

\begin{minipage}{0.24\textwidth}
\begin{algorithm}[H]
    \centering
    \caption{DARTS}\label{alg:darts}
    \footnotesize
    \begin{algorithmic}[1]
        \State \tiny{$I^j = \sum_{i<j} S(\alpha^{i,j}) x^i$}
        \State \tiny{$O^k = \oplus_{j<k} S(\gamma^{j,k}) x^j$}
        \State \tiny{$x^j = \sum_{o\in\mathcal{O}}S(\beta^{o}) o(I^j)$}
        \State \text{m $\gets$ DAG($I^j, O^k, x^j$)}
        \While{not converged}
        \State \text{m.update($\Lambda$, $\nabla_{\Lambda} \mathcal{L}_{valid}$)}
        \State \text{m.update($w$, $\nabla_w \mathcal{L}_{train}$)}
        \EndWhile
        \State \textbf{Return}  \text{$\Lambda$}
    \end{algorithmic}
\end{algorithm}
\end{minipage}
\hfill
\begin{minipage}{0.4\textwidth}
\begin{algorithm}[H]
    \centering
    \caption{PC-DARTS}\label{alg:pcdarts}
    \footnotesize
    \begin{algorithmic}[1]
        \State \tiny{$I^j = \sum_{i<j} S(\alpha^{i,j}) x^i$}
        \State \tiny{$O^k = \oplus_{j<k} S(\gamma^{j,k}) x^j$}
        \State \tiny{$x^j = \sum_{o\in\mathcal{O}}S(\beta^{o}) o($\hl{$M^o \ast$}$ I^j) $\hl{$+ (1- M^o \ast I^j)$}}
        \State \text{m $\gets$ DAG($I^j, O^k, x^j$)}
        \While{not converged}
        \State \text{m.update($\Lambda$, $\nabla_{\Lambda} \mathcal{L}_{valid}$)}
        \State \text{m.update($w$, $\nabla_w \mathcal{L}_{train}$)}
        \EndWhile
        \State \textbf{Return}  \text{$\Lambda$}
    \end{algorithmic}
\end{algorithm}
\end{minipage}
\hfill \sethlcolor{green}
\begin{minipage}{0.32\textwidth}
\begin{algorithm}[H]
    \centering
    \caption{GDAS}\label{alg:gdas}
    \footnotesize
    \begin{algorithmic}[1]
        \State \tiny{$I^j = \sum_{i<j}$\hl{GS}$(\alpha^{i,j}) x^i$}
        \State \tiny{$O^k = \oplus_{j<k}$\hl{GS}$(\gamma^{j,k}) x^j$}
        \State \tiny{$x^j = \sum_{o\in\mathcal{O}}$\hl{GS}$(\beta^{o}) o(I^j)$} 
        \State \text{m $\gets$ DAG($I^j, O^k, x^j$)}
        \While{not converged}
        \State \text{m.update($\Lambda$, $\nabla_{\Lambda} \mathcal{L}_{valid}$)}
        \State \text{\hl{m.update\_single\_path($w$, $\nabla_w \mathcal{L}_{train}$)}}
        \EndWhile
        \State \textbf{Return}  \text{$\Lambda$}
    \end{algorithmic}
\end{algorithm}
\end{minipage}

\sethlcolor{pink}
\begin{minipage}{0.49\textwidth}
\begin{algorithm}[H]
    \centering
    \caption{Random NAS with Weight-sharing}\label{alg:RandomNAS}
    \footnotesize
    \begin{algorithmic}[1]
        \State \tiny{$I^j = \sum_{i<j}$ \hl{$x^i$}}
        \State \tiny{$O^k = \oplus_{j<k}$ \hl{$x^j$}}
        \State \tiny{$x^j = \sum_{o\in\mathcal{O}}$ \hl{$o(I^j)$}}
        \State \text{m $\gets$ DAG($I^j, O^k, x^j$)}
        \While{not converged}
        \State \text{\hl{arch $\gets$ sample\_uniformly\_at\_random(m)}}
        \State \text{\hl{m.update\_weights\_of\_single\_architecture(arch, $w$, $\nabla_w \mathcal{L}_{train}$)}}
        \State 
        \EndWhile
        \For{\hl{i $\in$ 1..1000}}
        \State \text{\hl{arch\_samples $\gets$ sample\_uniformly\_at\_random(m)}}
        \EndFor
        \State \textbf{Return} \hl{arch $\in$ arch\_samples with lowest validation error}
    \end{algorithmic}
\end{algorithm}
\end{minipage}
\hfill
\begin{minipage}{0.49\textwidth}
\begin{algorithm}[H]
    \centering
    \caption{ENAS}\label{alg:enas}
    \footnotesize
    \begin{algorithmic}[1]
        \State \tiny{$I^j = \sum_{i<j}$ \hl{$x^i$}}
        \State \tiny{$O^k = \oplus_{j<k}$ \hl{$x^j$}}
        \State \tiny{$x^j = \sum_{o\in\mathcal{O}}$ \hl{$o(I^j)$}}
        \State \text{m $\gets$ DAG($I^j, O^k, x^j$)}
        \While{not converged}
        \sethlcolor{cyan}
        \State \text{\hl{arch $\gets$ sample\_using\_controller(m)}}
        \sethlcolor{pink}
        \State \text{\hl{m.update\_weights\_of\_single\_architecture(arch, $w$, $\nabla_w \mathcal{L}_{train}$)}}
        \sethlcolor{cyan}
        \State \hl{Update RNN controller}
        \EndWhile
        \For{\hl{i $\in$ 1..100}}
        \State \text{\hl{arch\_samples $\gets$ sample\_using\_controller(m)}}
        \EndFor
        \sethlcolor{pink}
        \State \textbf{Return} \hl{arch $\in$ arch\_samples with lowest validation error}
    \end{algorithmic}
\end{algorithm}
\end{minipage}

Algorithm \ref{alg:pcdarts} and Algorithm \ref{alg:gdas} highlight these components (relative to DARTS) for PC-DARTS~\citep{xu2019pcdarts} and GDAS~\citep{dong2019search}, respectively. 
For example, when comparing PC-DARTS and DARTS, the only difference in our benchmark is the partial channel connections (line 3 of Algorithm \ref{alg:pcdarts}) in the choice blocks, which consists of a channel sampling mask $M^o$ that drops feature maps coming from $I^j$. 
GDAS, on the other hand, replaces the Softmax (\emph{S)} function in DARTS by a Gumbel-Softmax (\emph{GS}), which applies for every architectural weight in $\Lambda =\{ \alpha, \beta, \gamma \}$ (lines 1-3 in Algorithm \ref{alg:gdas}), and uses this concrete distribution to sample single paths through the cell during search (line 7 in Algorithm \ref{alg:gdas}).
Random Search with Weight Sharing (Random WS)~\citep{li2019random} (Algorithm \ref{alg:RandomNAS}) and ENAS~\citep{Pham18} (Algorithm \ref{alg:enas}) do not need the continuous relaxation in order to conduct the architecture search, instead they sample randomly in RandomWS or from the recurrent neural network controller (line 6 in Algorithm \ref{alg:darts}) in ENAS, in order to select the sub-network in the one-shot model to train.

\begin{figure}[b]
\centering
\begin{subfigure}{.33\textwidth}
  \centering
\includegraphics[width=0.99999\textwidth]{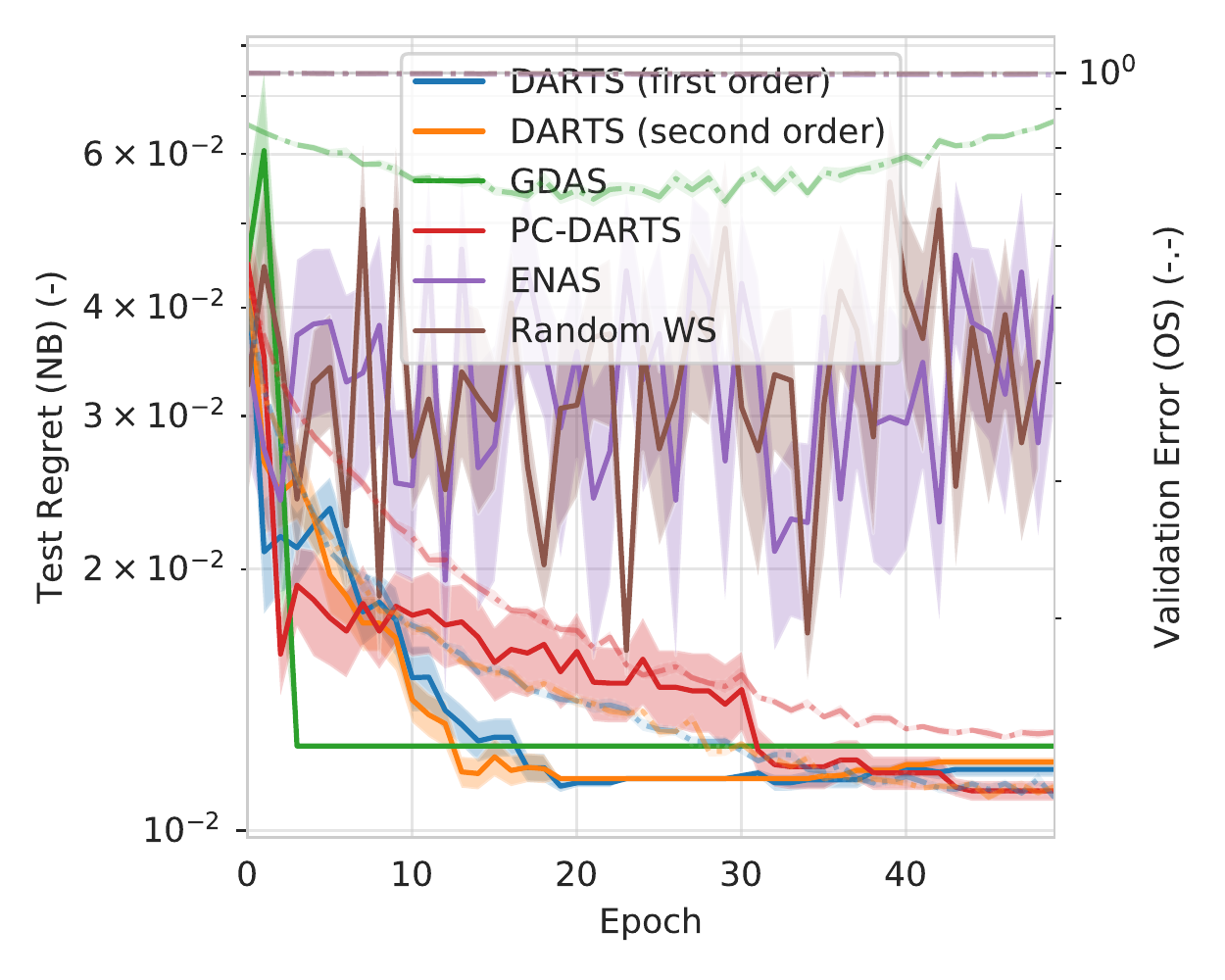}
  \caption{Search space 1}
  \label{fig:sec3:ss1_opt}
\end{subfigure}%
\begin{subfigure}{.33\textwidth}
  \centering
\includegraphics[width=0.99999\textwidth]{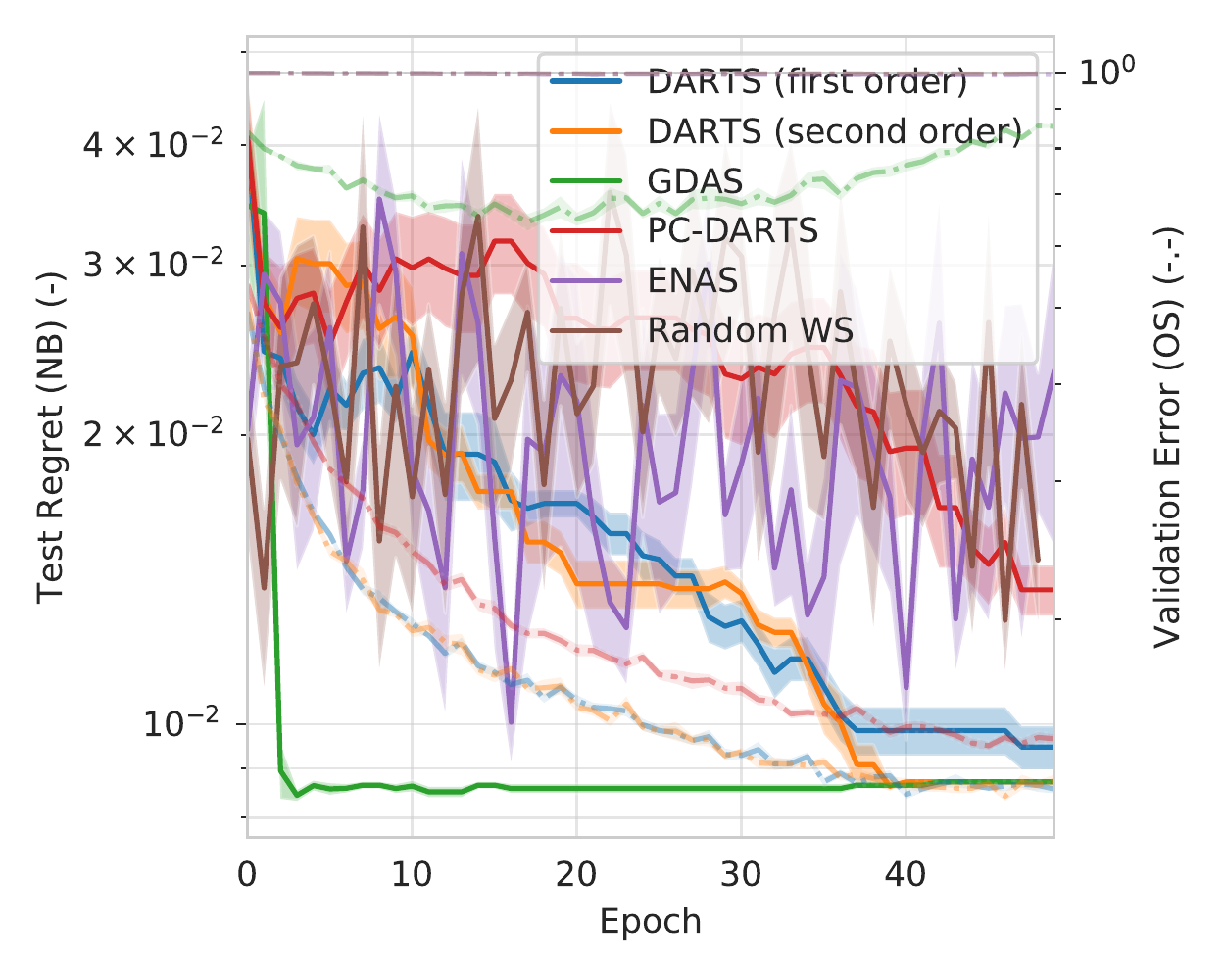}
  \caption{Search space 2}
  \label{fig:sec3:ss2_opt}
\end{subfigure}
\begin{subfigure}{.33\textwidth}
  \centering
\includegraphics[width=0.99999\textwidth]{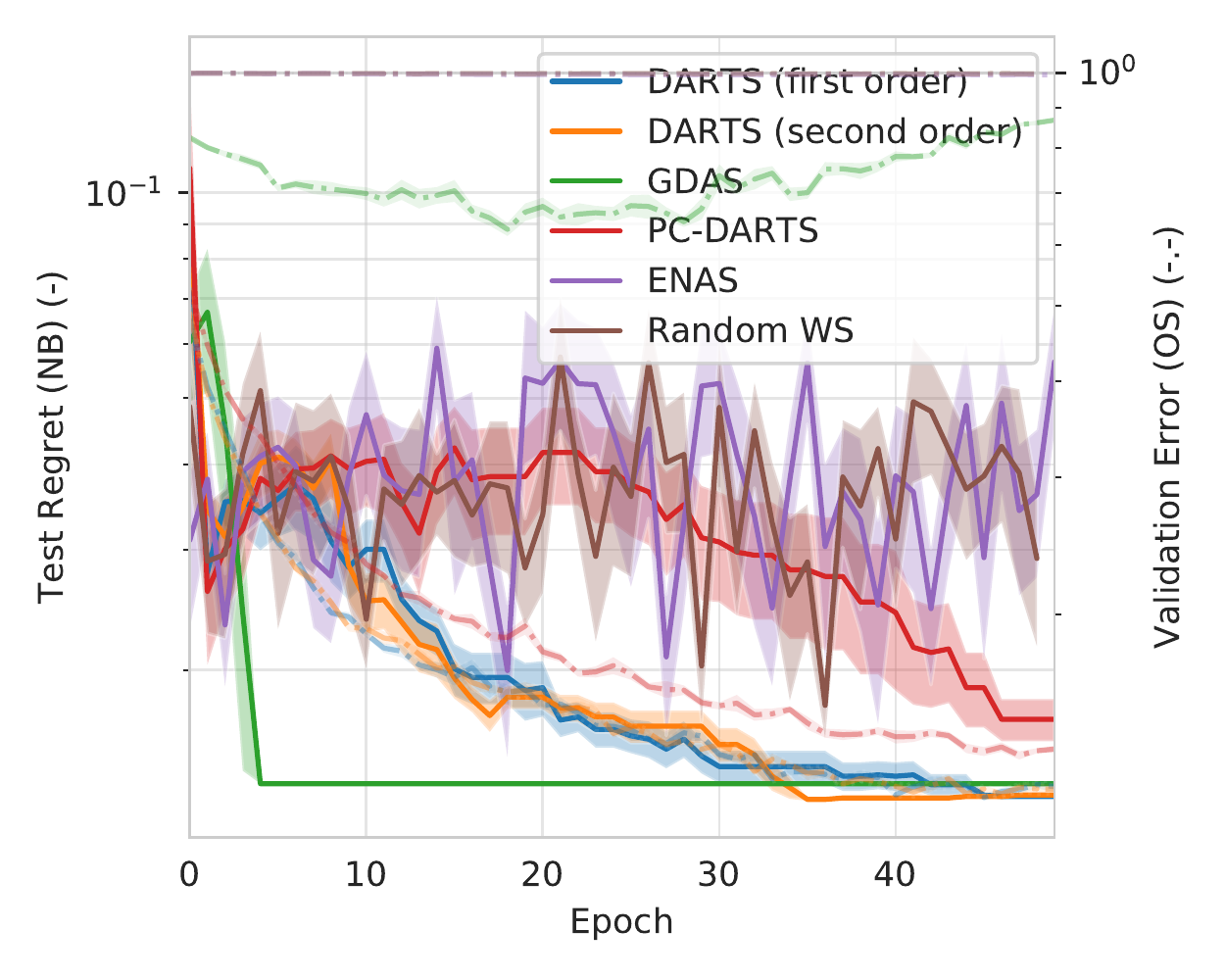}
  \caption{Search space 3}
  \label{fig:sec3:ss3_opt}
\end{subfigure}
\caption{Comparison of different one-shot NAS optimizers on the three different search spaces defined on NASBench. The solid lines show the anytime test regret (mean $\pm$ std), while the dashed blurred lines the one-shot validation error (Best viewed in color).}
\label{fig:sec3:optimizers:comparison}
\end{figure}

These close correspondences between current one-shot NAS variants provide an opportunity to implement all of these variants in the same general code basis.
This allows us to (a) automatically guard against any confounding factors when evaluating the strengths and weaknesses of different approaches, and (b) allows us to mix and match the components of different algorithms. 
%
We implemented all variants in a single code basis, which we are committed to grow into a flexible library of primitives for one-shot NAS methods, and for which we will gladly accept any help the community wants to provide.

One-shot NAS methods in this code basis inherit all the methods and attributes necessary for building the one-shot computational graph from a base parent class. 
This encapsulation and modularity ensures that all differences in their performance come from a few lines of code, and that all other confounding factors cannot affect these results. This will also facilitate the incorporation of other one-shot NAS methods and pinpoint the components that differ in them.

Furthermore, the primitives encoding the search spaces presented in Section \ref{sec:methodology} are defined separately from the NAS optimizers. This encapsulation will 
allow researchers to study each of these components in isolation, experimenting with one of them while being sure that the other one does not change.

\section{NAS-Bench-1Shot1 as a benchmark and analysis framework}
\label{sec:experiments}

We now demonstrate the use of NAS-Bench-1Shot1 as a benchmark for one-shot NAS. We first evaluate the anytime performance of five different one-shot NAS methods: DARTS~\citep{darts}, GDAS~\citep{dong2019search}, PC-DARTS~\citep{xu2019pcdarts}, ENAS \citep{Pham18} and Random Search with Weight Sharing (Random WS) \citep{li2019random}.\footnote{Details on these optimizers can be found in Appendix \ref{sec:app:optimizers}.} Afterwards, we investigate the robustness of these one-shot NAS optimizers towards their search  hyperparameters and show that if these hyperparameters are carefully tuned, the one-shot NAS optimizer can outperform a wide range of other discrete NAS optimizers in our search spaces.

\subsection{Comparison of different One-shot NAS optimizers}
\label{subsec: comparison}

We ran the NAS search for 50 epochs\footnote{This required different amounts of time: DARTS 1st order: 3h, DARTS 2nd order: 4h, GDAS: 1.5h, PC-DARTS: 2h, ENAS: 4h, Random-WS: 11h.} using their respective default hyperparameter settings (see Appendix \ref{sec: hyperparameters}).
If not stated otherwise, all the following results were generated by running each experiment with six random seeds (0 to 5). All plots show the mean and standard deviation of the test regret queried from NAS-Bench-101. Over the three independent trainings contained in NASBench-101 for each architecture on each epoch budget we average. The search was done on a single NVIDIA RTX2080Ti using the same python environment. In Figure \ref{fig:sec3:optimizers:comparison} we report the anytime test regret for each of these methods. Our findings can be summarized as follows:

\begin{itemize}[leftmargin=*]
    \item The test error (queried from NAS-Bench-101) for DARTS, GDAS and PC-DARTS decreases along with the validation error of the one-shot model. However, these two quantities are not necessarily correlated. One such example is the learning curve of PC-DARTS. While the validation error smoothly decreases, the test error of the found solutions slightly goes up in the beginning and then decreases again. This behavior becomes even more substantial when increasing the number of search epochs (see Appendix~\ref{sec: more_search_epochs}). Figure~\ref{fig:sec3:optimizers:comparison_200} shows that when running DARTS for 200 epochs, the test regret diverges. The same result was also previously observed by \citet{zela19} on subspaces of the standard DARTS space.
    \item The ranking of the NAS optimizers differs across search spaces. For instance, while PC-DARTS performs the best in search space 1, this is clearly not the case in the other search spaces. 
    \item GDAS has a better anytime performance than the other optimizers in all 3 benchmarks, however, due to the temperature annealing of the Gumbel Softmax, it manifests some premature convergence (in less than 5 search epochs) to a sub-optimal local minimum.
    \item Random WS and ENAS mainly explore poor architectures across all three search spaces. This behaviour is a result of the small correlation between the architectures evaluated with the one-shot weights and their true performance during architecture evaluation (as queried from NAS-Bench-101 (see Section \ref{sec:corr_analysis})). This correlation directly affects these methods since in the end of search they sample a certain number of architectures (1000 randomly sampled for Random WS and 100 using the learned controller policy for ENAS) and evaluate them using the one-shot model weights in order to select the architecture which is going to be trained from scratch in the final evaluation phase. When running ENAS for 100 epochs in search space 2 (Figure \ref{fig:sec3:optimizers:comparison_100_epochs} in the appendix) we see that the it performs better than Random WS. ENAS also has a stronger correlation between the sampled architectures and the NAS-Bench-101 architectures for search space 2 (see Section \ref{sec:corr_analysis}).
\end{itemize}

\subsection{Correlation analysis} \label{sec:corr_analysis}
Many one-shot NAS methods, such as ENAS~\citep{Pham18}, NAONet~\citep{Luo2018NeuralAO} or Random WS~\citep{li2019random} select the final architecture by means of the one-shot parameters.
In order to assess if this is optimal we computed the correlation between a given architecture's one-shot test error and its respective NAS-Bench-101 test error, for all 4 available budgets in NAS-Bench-101 on every 10th search epoch. This analysis was performed for all architectures without loose ends in each search space. The only exception is ENAS, for which we decided to evaluate the correlation by sampling 100 architectures (as done by the algorithm after the search has finished to select the one to retrain from scratch) from the controller instead of evaluating every architecture in the search space.

\begin{figure}[t]
\centering
\begin{subfigure}{.325\textwidth}
    \includegraphics[width=0.99\textwidth]{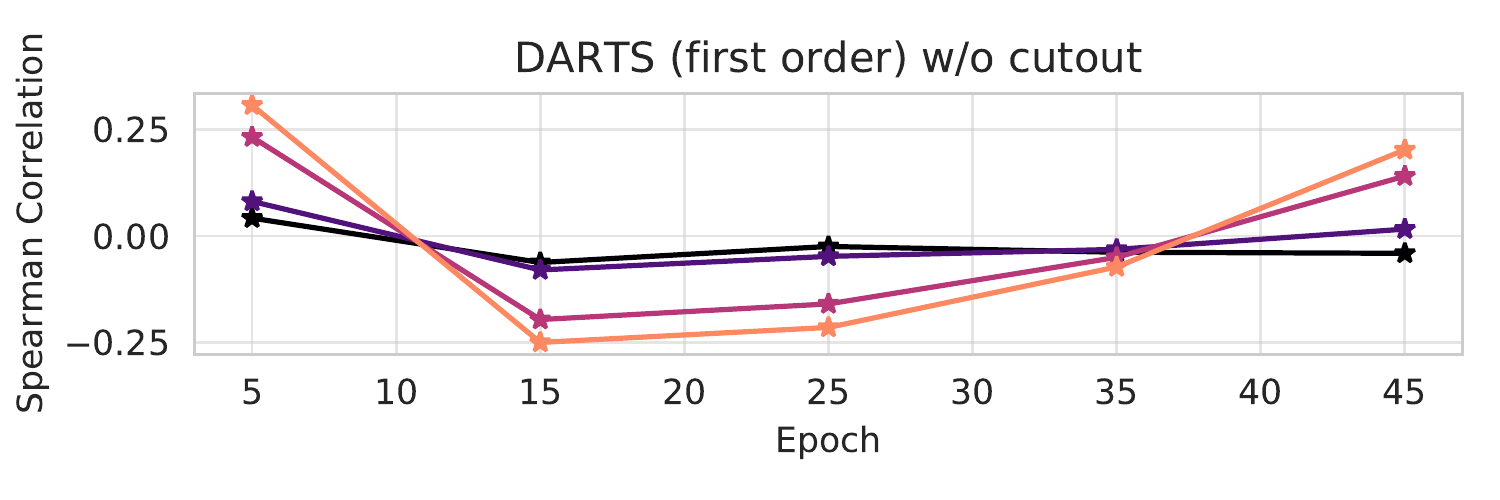}\\
    \includegraphics[width=0.99\textwidth]{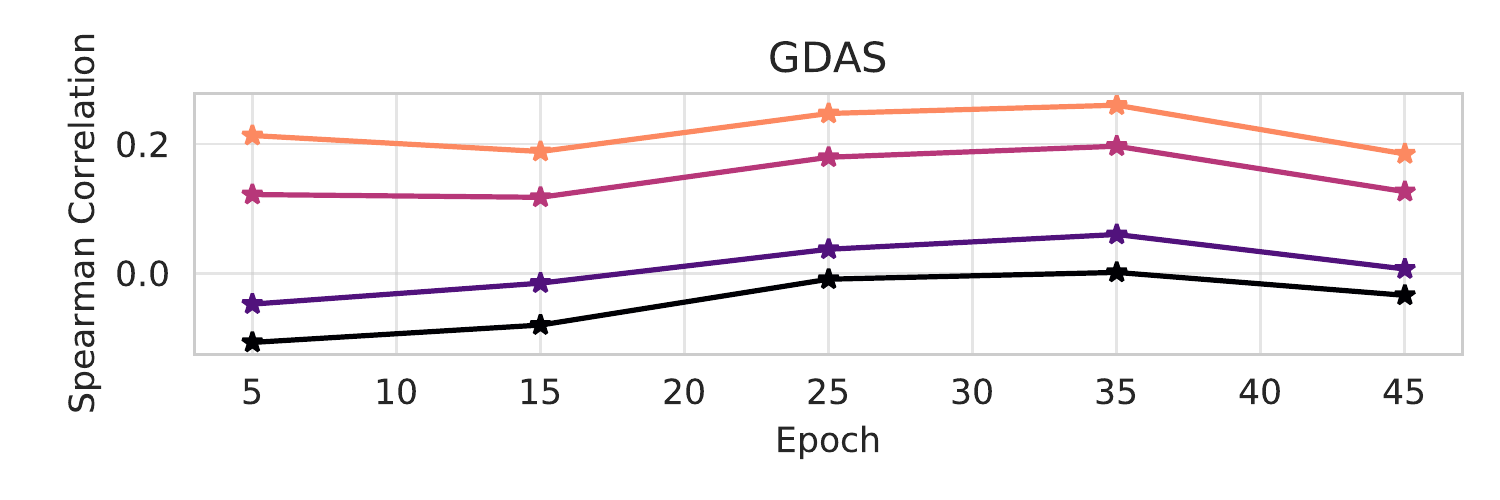}\\
    \includegraphics[width=0.99\textwidth]{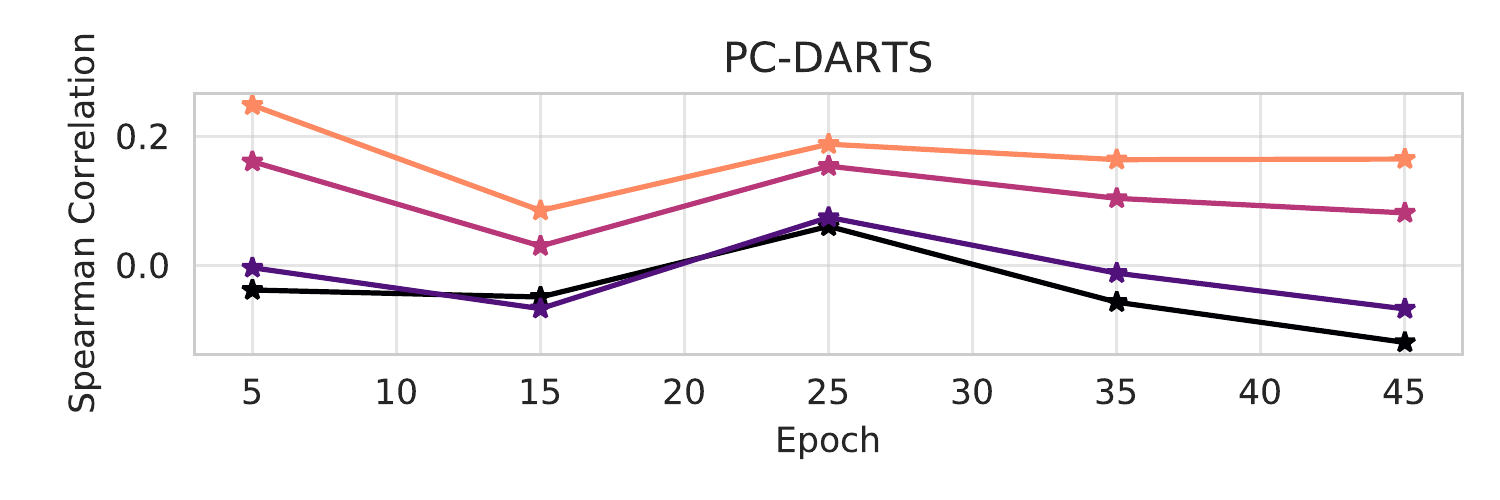}\\
    \includegraphics[width=0.99\textwidth]{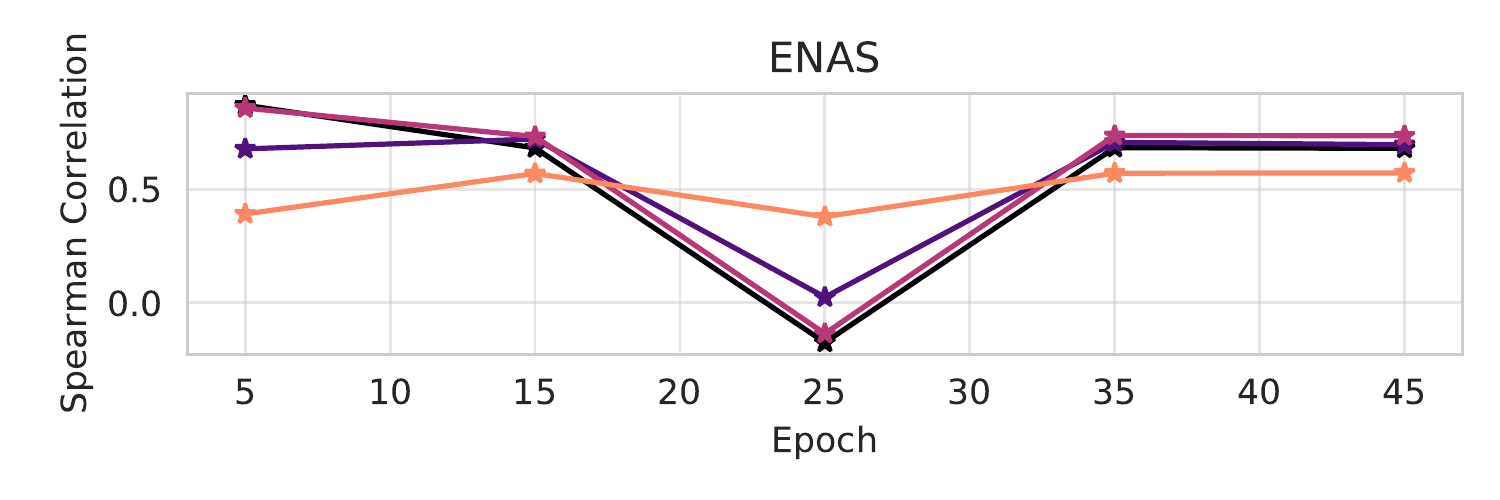}\\
    \includegraphics[width=0.99\textwidth]{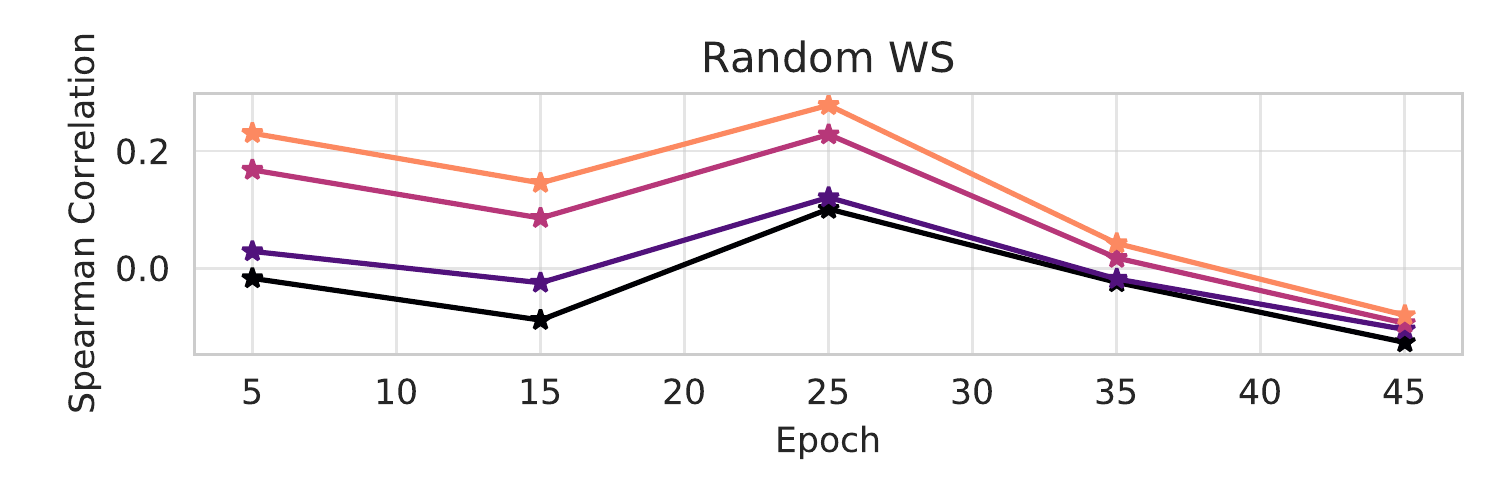} 
    \caption{Search space 1}
    \label{fig:sec3:ss1:corr}
\end{subfigure}
\centering
\begin{subfigure}{.33\textwidth}
    \includegraphics[width=0.99\textwidth]{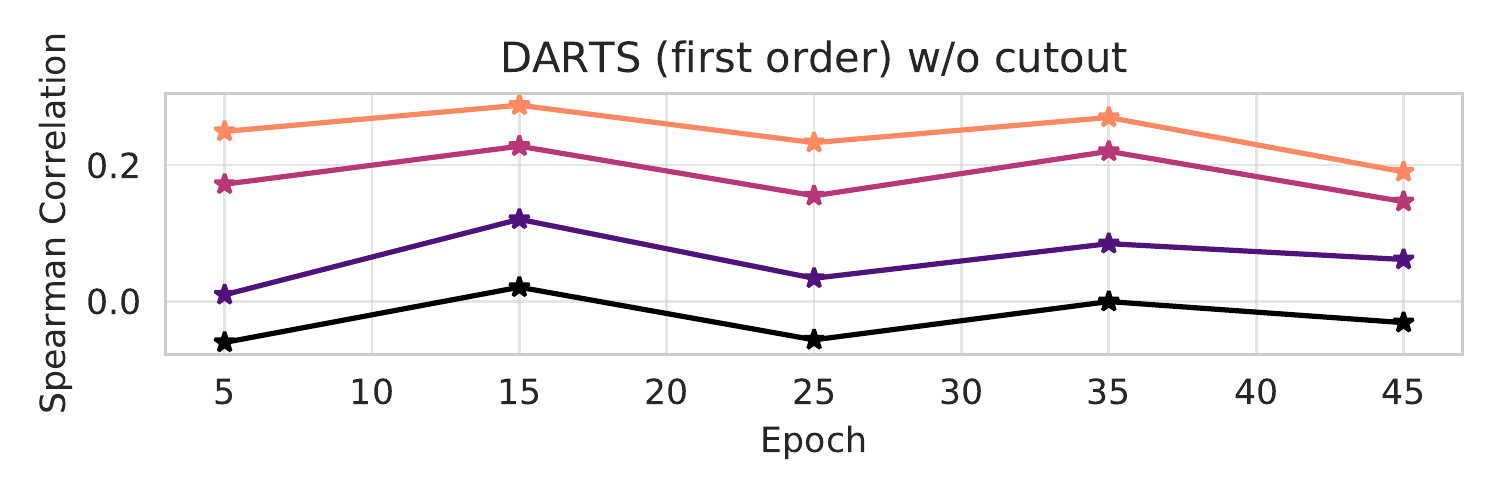}\\
    \includegraphics[width=0.99\textwidth]{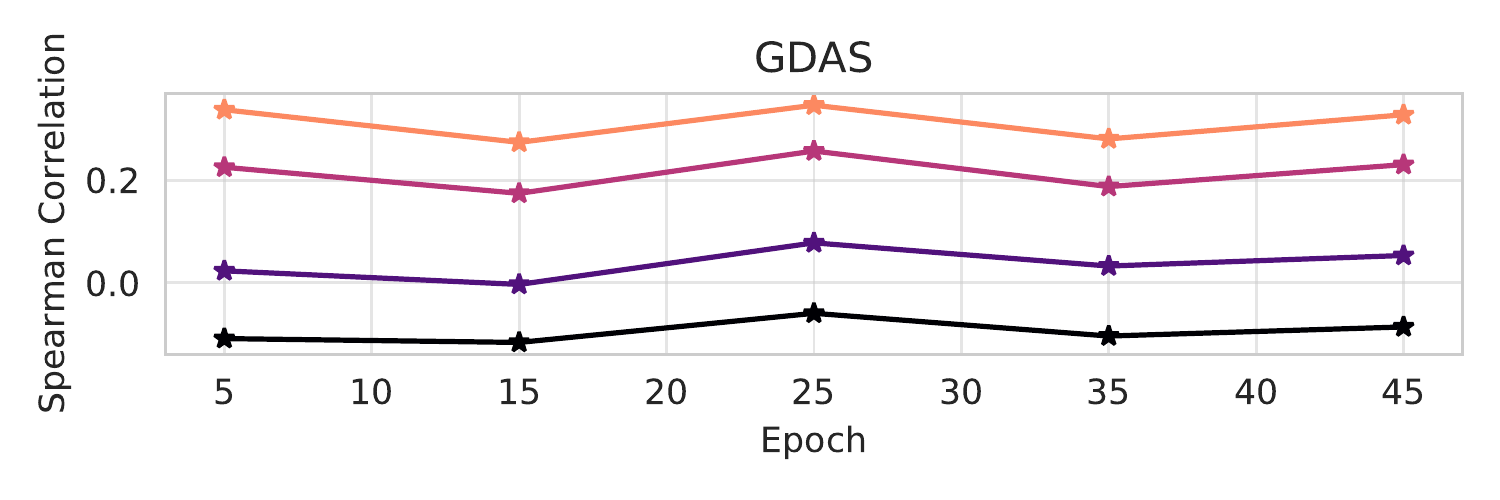}\\
    \includegraphics[width=0.99\textwidth]{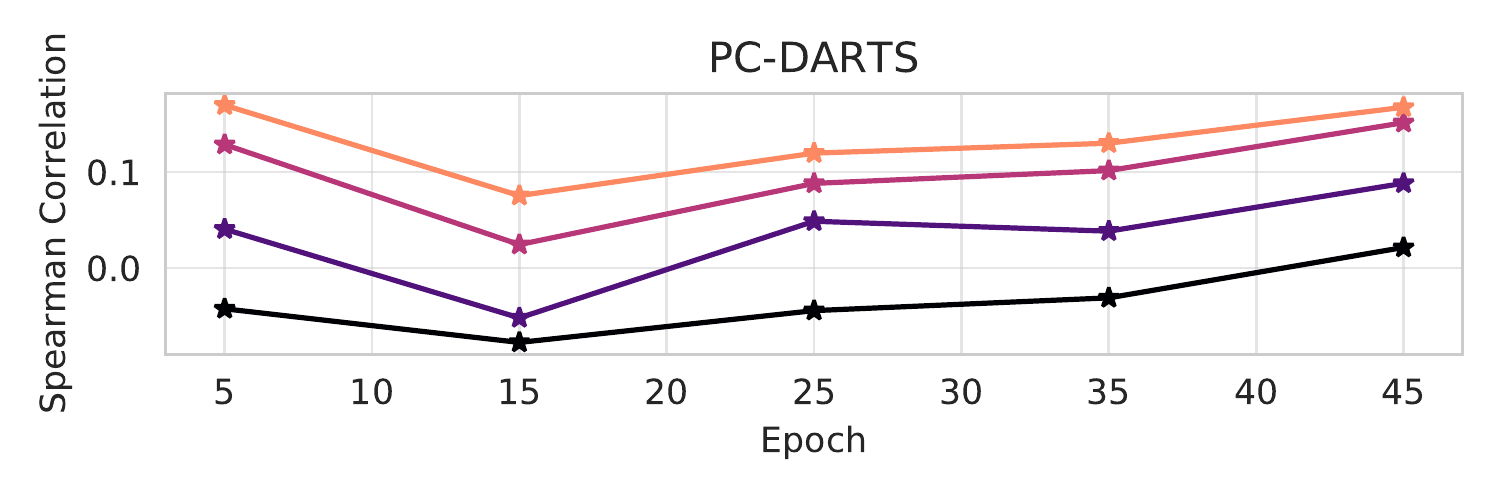}\\
    \includegraphics[width=0.99\textwidth]{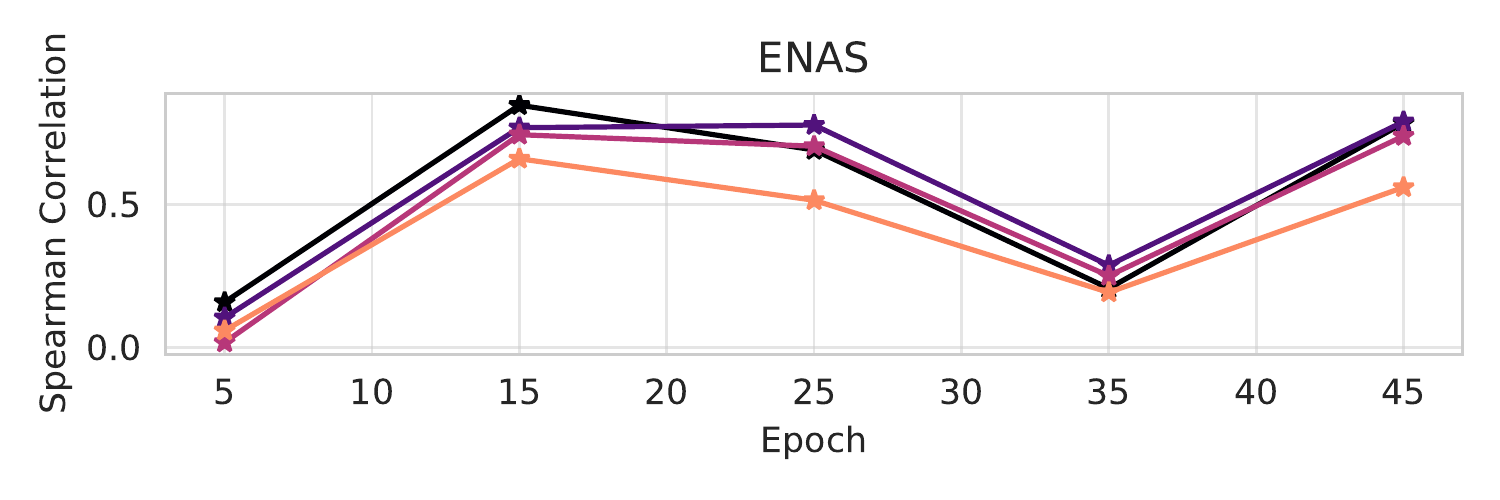}\\
    \includegraphics[width=0.99\textwidth]{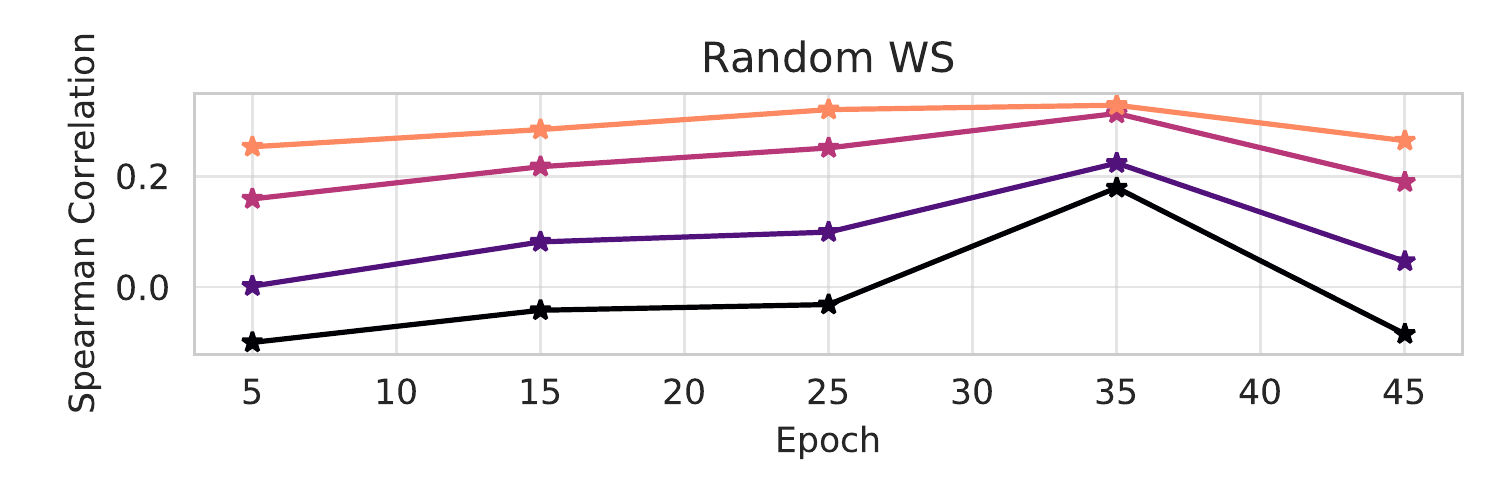}   
    \caption{Search space 2}
    \label{fig:sec3:ss2:corr}
\end{subfigure}
\begin{subfigure}{.33\textwidth}
    \includegraphics[width=0.99\textwidth]{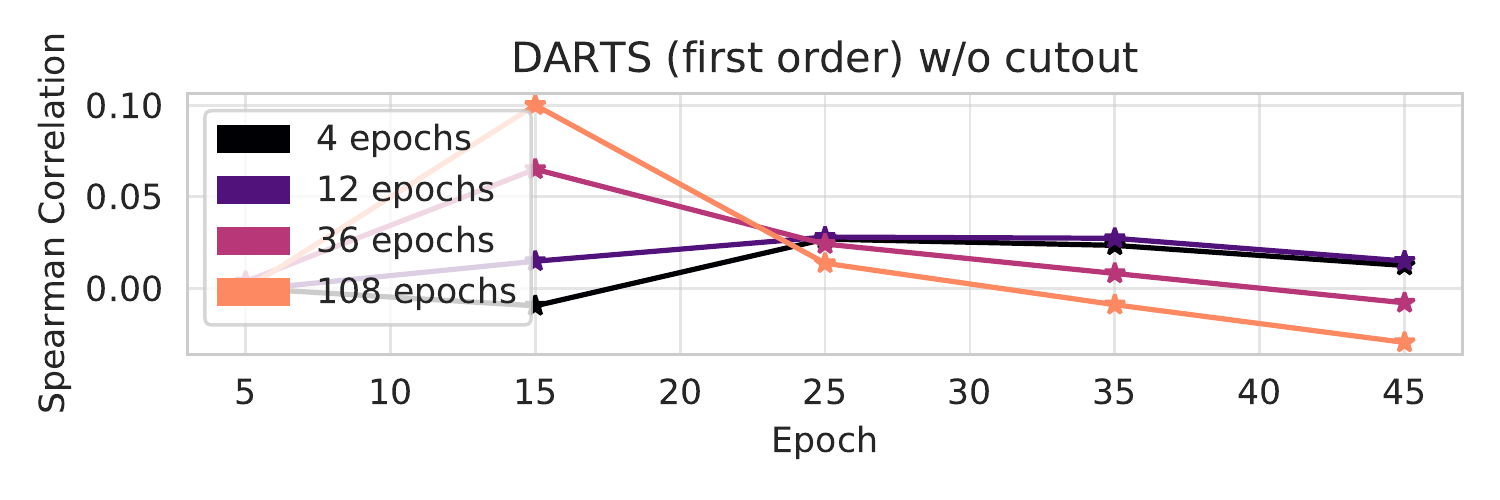}\\
    \includegraphics[width=0.99\textwidth]{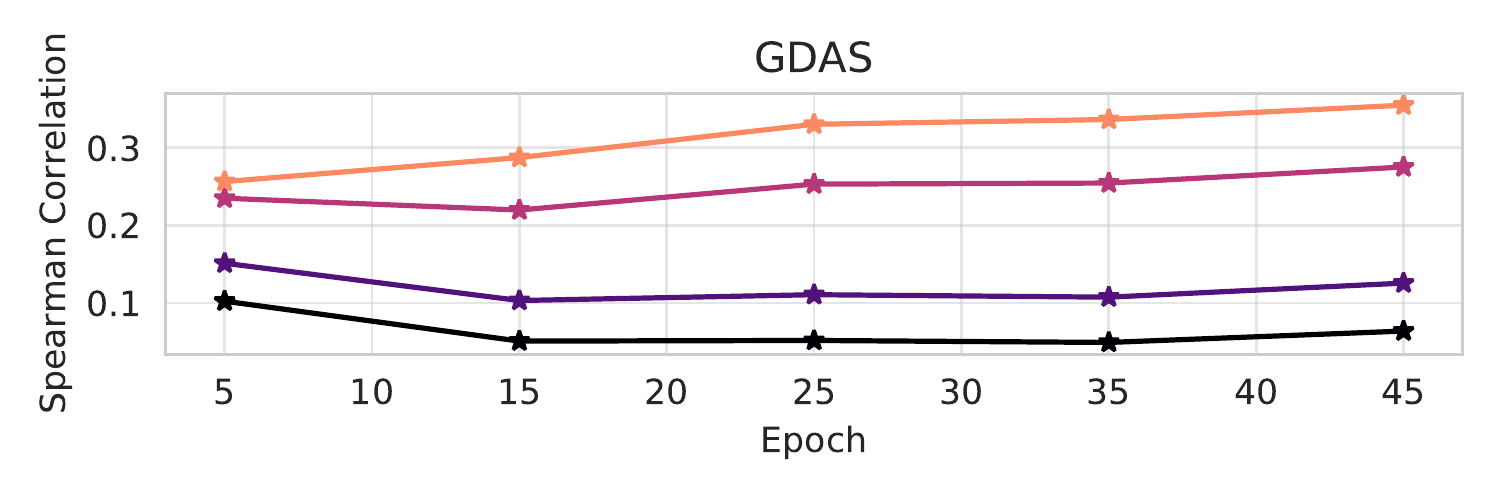}\\
    \includegraphics[width=0.99\textwidth]{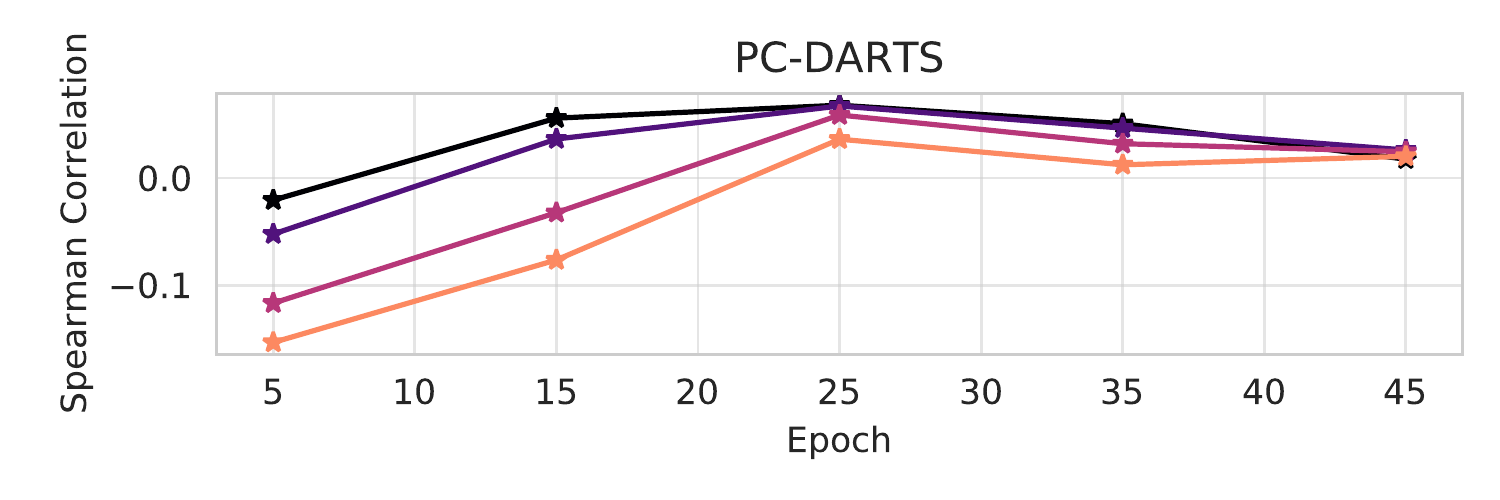}\\
    \includegraphics[width=0.99\textwidth]{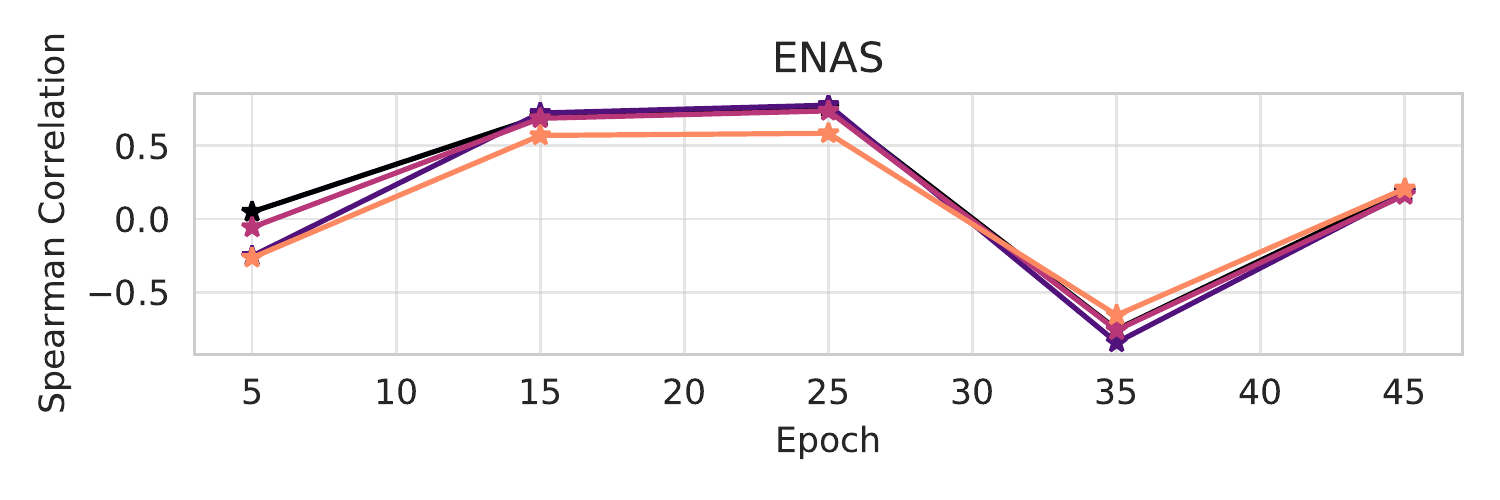}\\
    \includegraphics[width=0.99\textwidth]{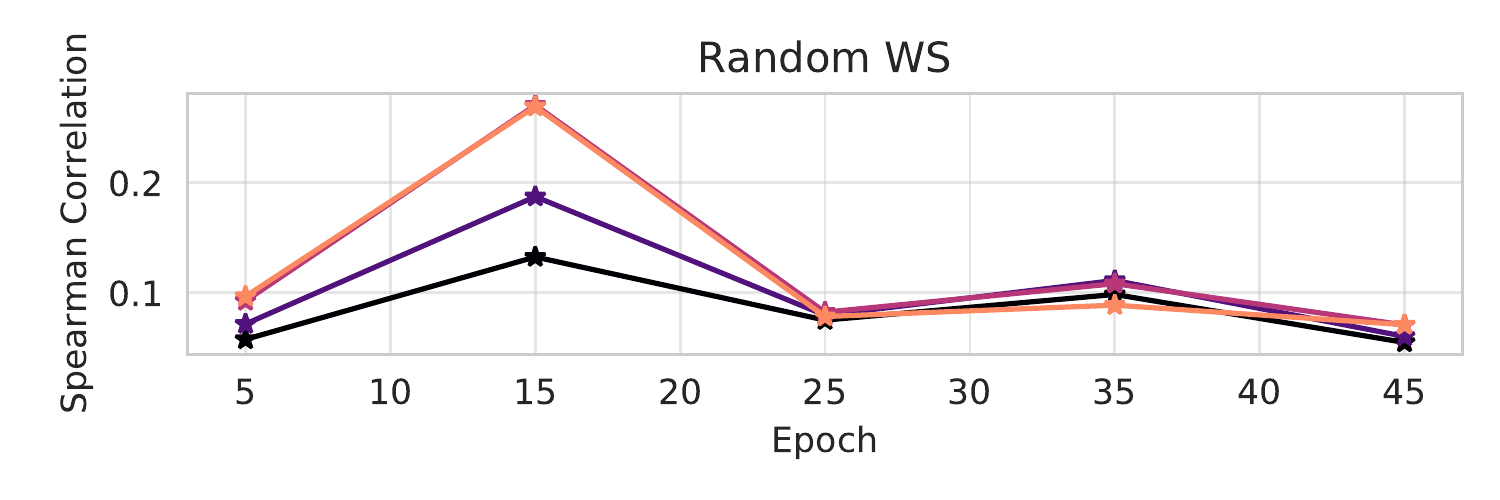}   
    \caption{Search space 3}
    \label{fig:sec3:ss3:corr}
\end{subfigure}
\caption{Correlation between the one-shot validation error and the corresponding NAS-Bench-101 test error for each search space. (Best viewed in color).}
\label{fig: correlation_s1_s2_s3}
\end{figure}

As shown in Figure \ref{fig: correlation_s1_s2_s3}, there is almost no correlation between the weight sharing ranking and the true one (Spearman correlation coeff. between -0.25 and 0.3) during search for DARTS, PC-DARTS, GDAS and Random WS. Only ENAS shows some correlation for search space 2 and some anticorrelation for search spaces 1 and 3. These results agree with the ones reported by \citet{sciuto19} (who could only do this evaluationx on a small search space) and explain the poor performance of Random WS and ENAS on our benchmarks, since the architectures sampled during evaluation and ranked according to their one-shot validation error are unlikely to perform well when evaluated independently. To the best of our knowledge this is the first time that an evaluation of this correlation is conducted utilizing such a large number of architectures in the search space, namely 137406 different architectures for search space 3. We added further experiments on the correlation between the lower fidelity proxy model used in architecture search and the final model from architecture evaluation in the Appendix~\ref{sec:correlation_proxy}.

\subsection{Robustness of One-Shot NAS optimizers}
\label{subsec:regulatization}

\citet{zela19} observed that the generalization performance of architectures found by DARTS heavily depends on the hyperparameters used during the search phase. More specifically, they investigate regularization hyperparameters that shape the inner objective landscape, such as $L_2$ regularization or CutOut~\citep{cutout} and ScheduledDropPath~\citep{zoph-arXiv18}.

Due to the fast queries from NAS-Bench-101, NAS-Bench-1Shot1 enables researchers to assert if their one-shot NAS optimizers with some specified hyperparameter settings overfit on the validation set during search~\citep{zela19}. In Figure~\ref{fig:sec3:regularization:l2_ss3} we plot the anytime performance of DARTS, GDAS and PC-DARTS on Search Space 3 when ran with different $L_2$ factors in the inner objective (see Appendix~\ref{sec: regularization} for more results on other search spaces and regularizers such as CutOut). Based on these results we observe that:

\begin{figure}[ht]
\centering
\begin{subfigure}{.33\textwidth}
  \centering
\includegraphics[width=0.99\textwidth]{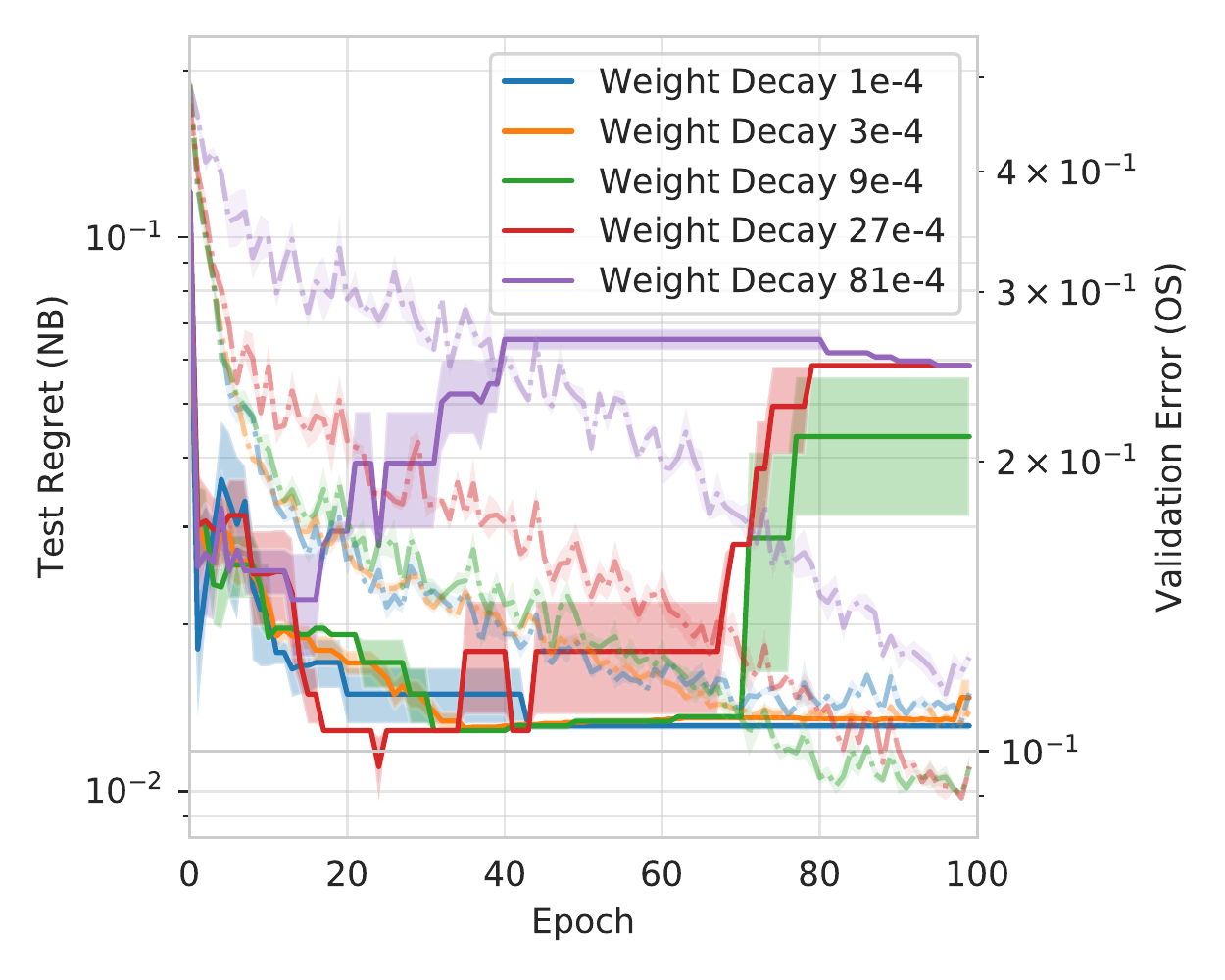}
  \caption{DARTS}
  \label{fig:sec3:ss3_l2}
\end{subfigure}
\begin{subfigure}{.325\textwidth}
  \centering
\includegraphics[width=0.99\textwidth]{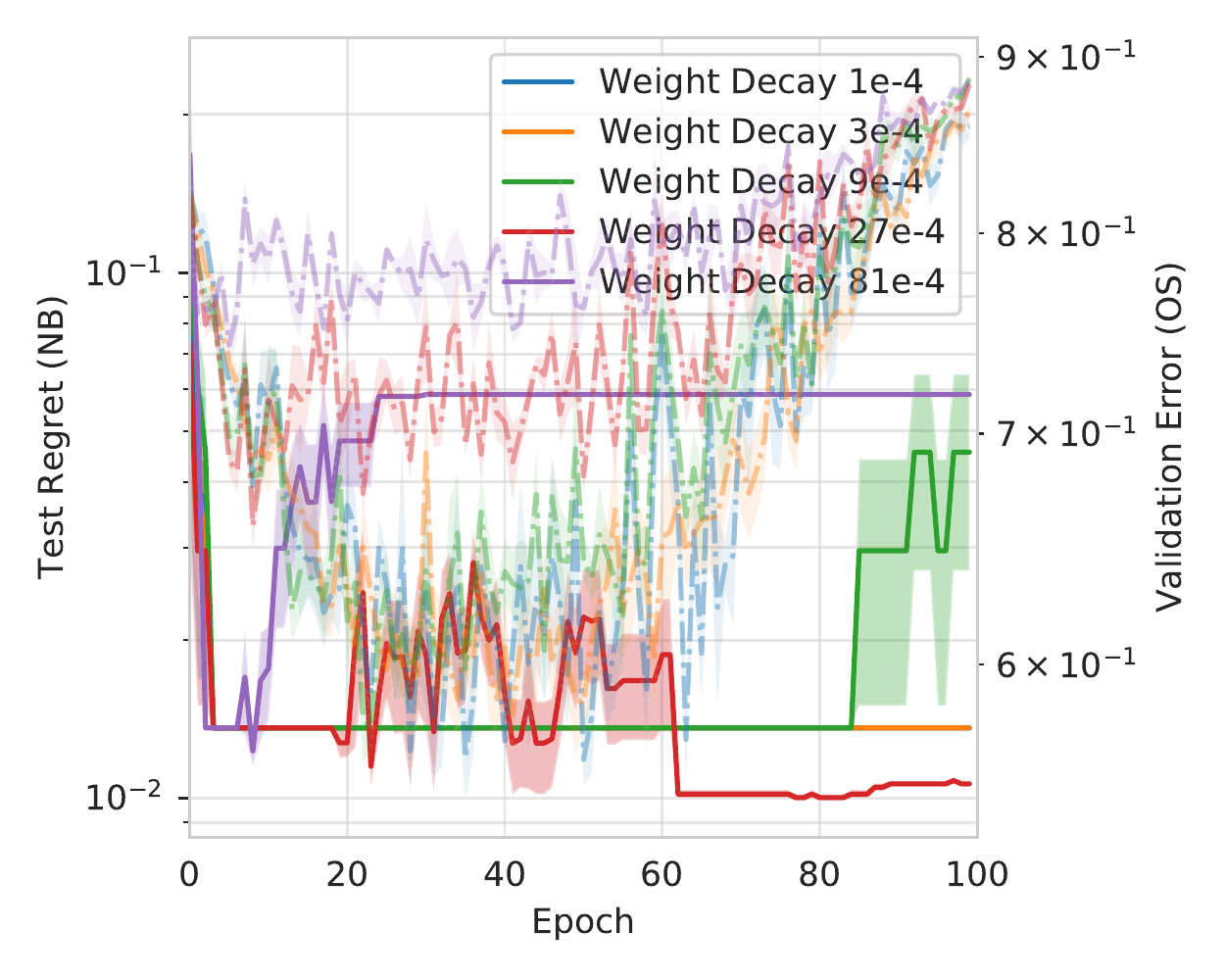}
  \caption{GDAS}
  \label{fig:sec3:reg:l2_gdas_3}
\end{subfigure}
\begin{subfigure}{.33\textwidth}
  \centering
\includegraphics[width=0.99\textwidth]{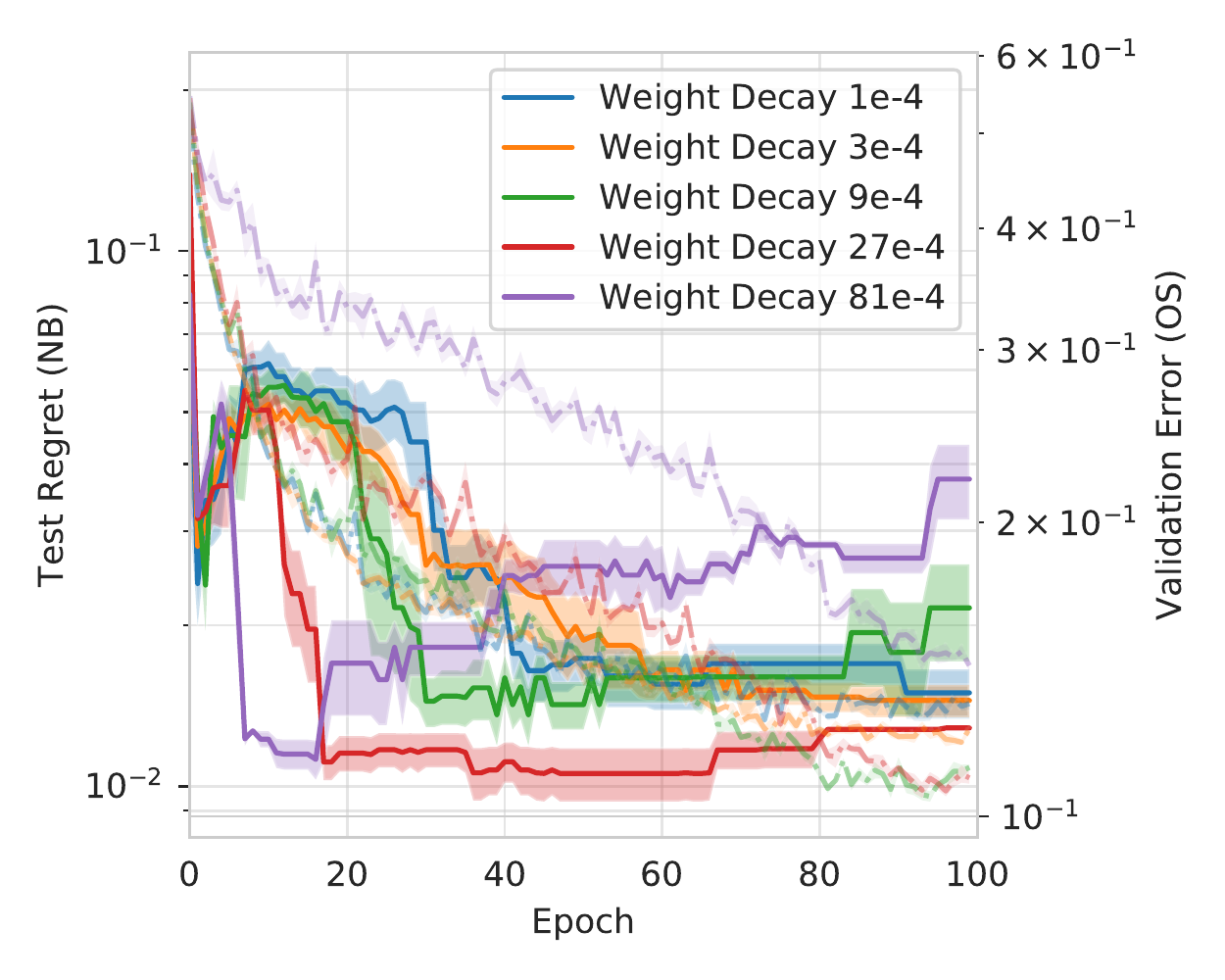}
  \caption{PC-DARTS}
  \label{fig:sec3:reg:l2_pc_darts_3}
\end{subfigure}
\caption{The impact that weight decay has on the test (one-shot validation) performance of architectures found by DARTS, GDAS and PC-DARTS on search space 3 (Best viewed in color).} 
\label{fig:sec3:regularization:l2_ss3}
\end{figure}

\begin{itemize}[leftmargin=*]
    \item Interestingly, while \citet{zela19} only showed overfitting behavior for DARTS, it may also occur for GDAS and PC-DARTS as shown Figure \ref{fig:sec3:reg:l2_gdas_3} and \ref{fig:sec3:reg:l2_pc_darts_3}. This indicates that this might be an intrinsic property of these methods due to the local updates in the architecture space.
    \item A good hyperparameter setting for one NAS optimizer is not necessarily good for others. In Figure~\ref{fig:sec3:regularization:l2_ss3} we can see that for $L_2 = 27\cdot 10^{-4}$ GDAS and PC-DARTS have the best anytime performance, whilst in DARTS we notice an overfitting behavior for the same value.
    \item Interestingly enough, it seems that these NAS optimizers show the same pattern across search spaces for the same $L_2$ values as shown in Figure~\ref{fig:sec3:regularization:l2_ss1} and Figure~\ref{fig:sec3:regularization:l2_ss2} in Appendix~\ref{sec: regularization}.
\end{itemize}

\subsection{Tunability of One-Shot NAS hyperparameters} \label{subsec: bohb_darts}

Next to the hyperparameters used in the evaluation pipeline, one-shot NAS methods have several hyperparameters of their own, such as the regularization hyperparameters studied in Section \ref{subsec:regulatization}, learning rates, and other hyperparameters of the search phase.

\begin{figure}[ht]
\centering
\begin{subfigure}{.33\textwidth}
  \centering
\includegraphics[width=0.99\textwidth]{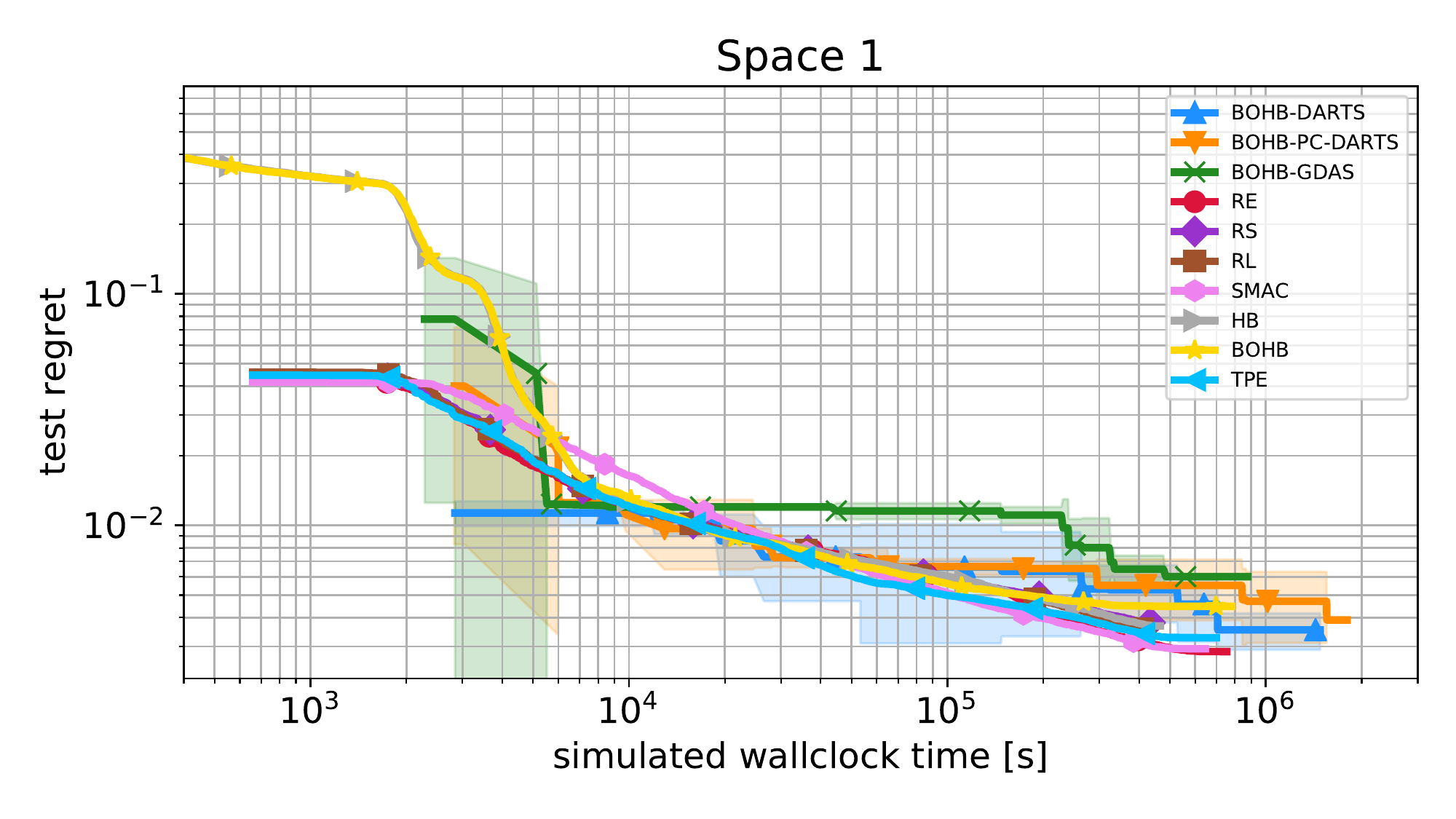}
  \caption{Search space 1}
  \label{fig:bohb-s1-cs2}
\end{subfigure}%
\begin{subfigure}{.33\textwidth}
  \centering
\includegraphics[width=0.99\textwidth]{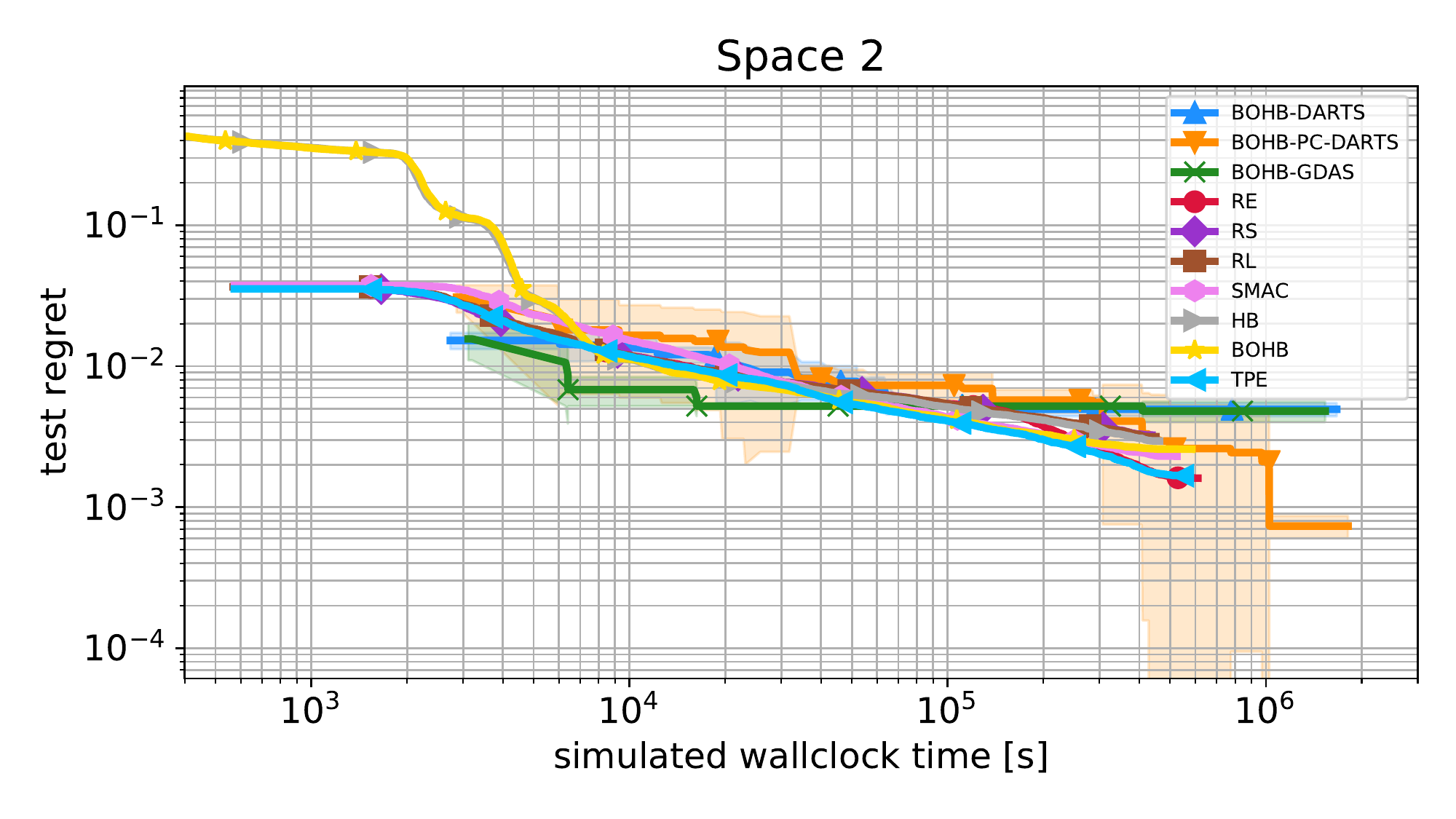}
  \caption{Search space 2}
  \label{fig:bohb-s2-cs2}
\end{subfigure}
\begin{subfigure}{.33\textwidth}
  \centering
\includegraphics[width=0.99\textwidth]{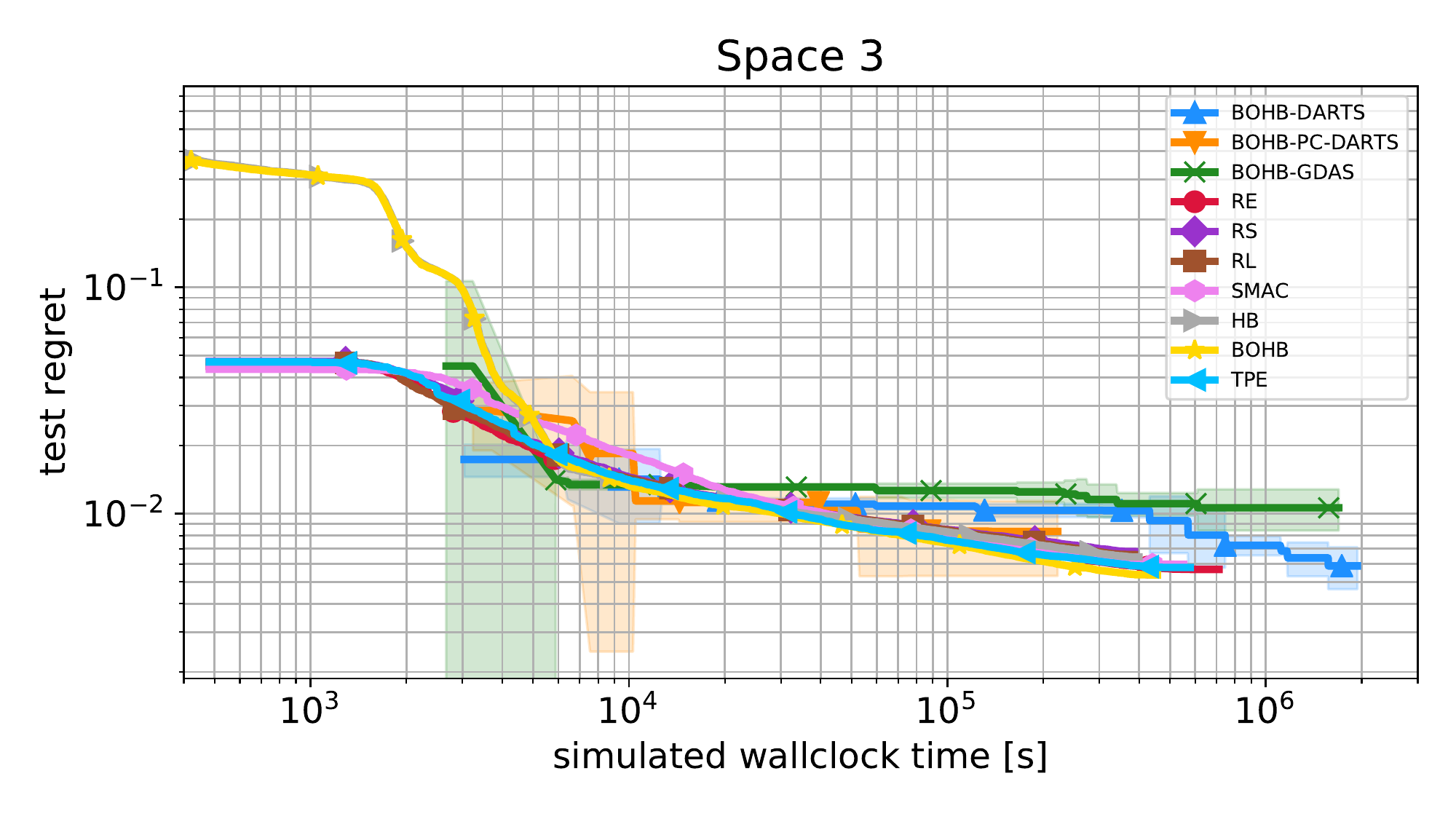}
  \caption{Search space 3}
  \label{fig:bohb-s3-cs2}
\end{subfigure}
\caption{Optimizing the hyperparameters of one-shot optimizers with BOHB on NAS-Bench-1Shot1. Every curve shows the mean$\pm$std of the incumbent trajectories (500 and 3 search repetitions for the black-box and one-shot optimizers, respectively).}
\label{fig:bohb-cs2}
\end{figure}


Naively tuning these hyperparameters with the one-shot validation error as the objective would lead to sub-optimal configurations, since, as we saw, this metric is not a good indicator of generalization.
In our proposed benchmarks, we can tune these hyperparameters of the NAS optimizer to minimize the validation error queried from NAS-Bench-101. By doing so, we aim to shed more light onto the influence these hyperparameters have during the one-shot search, and to study the sensitivity of the NAS method towards these hyperparameters.

To this end, we constructed 3 configuration spaces of increasing cardinality, CS1, CS2 and CS3 (see Appendix \ref{sec: bohb_details} for details), which only include hyperparameters controlling the NAS process. We chose BOHB (\cite{Falkner18}, see Appendix \ref{sec: bohb_details} for details) as the hyperparameter optimization method and DARTS (1st order), GDAS, PC-DARTS as our NAS methods to be tuned across all configuration spaces.

Starting from the respective default hyperparameter settings of each one-shot optimizer, we run BOHB with a total budget of 280 function evaluations, where in this case a function evaluation is a query from NAS-Bench-101.
Note that we never use the validation set split used to evaluate the individual architectures in NAS-Bench-101 during the architecture search. This subset with 10k examples is only used to compute the objective function value that BOHB optimizes. Therefore, the one-shot optimizers use 20k examples for training and 20k for the architectural parameter updates.
Figure \ref{fig:bohb-cs2} shows the anytime test regret of the architectures found by the respective NAS method's configurations tried by BOHB on CS2 (see Figure~\ref{fig:bohb-cs1} for results on CS1 and Figure~\ref{fig:bohb-cs3} for results on CS3). The x-axis in these figures shows the \textit{simulated wall-clock time}: $t_{sim}=t_{search}+t_{train}$, where $t_{search}$ is the time spent during search by each configuration and $t_{train}$ is the training time for 108 epochs (queried from NAS-Bench-101) of the architectures selected by the one-shot optimizer. As the figure shows, DARTS, PC-DARTS and GDAS can be tuned to find solutions closer to the global minimum compared to the default hyperparameter settings. 
We note that carrying out this optimization on NAS-Bench-1Shot1 reduced the time for tuning DARTS from a simulated 45 GPU days to 1 day on 16 GPUs.

Figure \ref{fig:bohb-cs2} also provides an evaluation of DARTS, GDAS and PC-DARTS compared to the state-of-art discrete NAS optimizers used by \citet{ying19a} (such as RL, RE, and HPO methods). Since these one-shot NAS methods are much faster than black-box methods, it is possible to tune them \emph{online} using BOHB and the resulting BOHB-$\{$DARTS, GDAS, PC-DARTS$\}$ typically still yields better performance over time. 
From this experiment, we make the following observations:

\begin{itemize}[leftmargin=*]
    \item Across all search spaces the architectures found using the best BOHB configurations outperformed the architectures found by the default DARTS configuration by up to a factor of 10.
    \item As in NAS-Bench-101, Regularized Evolution~\citep{real-arXiv18a} performed the best in all search spaces of NAS-Bench-1Shot1 compared to the other black-box optimizers.
    \item The found robust configurations did not only avoid overfitting in the architectural level, but they also typically outperformed the architectures found by state-of-art discrete NAS optimizers used by \citet{ying19a} (such as RL, RE, and HPO methods). 
    \item The multi-fidelity in BOHB does not only accelerate the hyperparameter optimization procedure, but in this case also allows to determine the sufficient number of epochs to run the NAS optimizer in order to get an optimal architecture. In fact, the best configuration on each of the incumbents comes usually from the lowest budget, i.e. 25 search epochs.
\end{itemize}

\section{Conclusion and Future Directions}
\label{sec: conclusion}

We proposed NAS-Bench-1Shot1, a set of 3 new benchmarks for one-shot neural architecture search which allows to track the trajectory and performance of the found architectures computationally cheaply. Using our analysis framework, we compared state-of-the-art one-shot NAS methods and inspected the robustness of the methods and how they are affected by different hyperparameters. Our framework allows a fair comparison of any one-shot NAS optimizer and discrete NAS optimizers without any confounding factors.
We hope that our proposed framework and benchmarks will facilitate the evaluation of existing and new one-shot NAS methods, improve reproducibility of work in the field, and lead to new insights on the underlying mechanisms of one-shot NAS.

\section*{Acknowledgments}
The authors acknowledge funding by the Robert Bosch GmbH, support by the European Research Council (ERC) under the European Unions Horizon 2020 research and innovation program through grant no. 716721, and by BMBF grant DeToL. We would like to thank Yuge Zhang for his work on reproducing our results, and making us aware of his results on counting the number architectures in our search space.

\clearpage
\bibliography{bib/strings,bib/local,bib/lib,bib/proc}

\begin{thebibliography}{33}
\providecommand{\natexlab}[1]{#1}
\providecommand{\url}[1]{\texttt{#1}}
\expandafter\ifx\csname urlstyle\endcsname\relax
  \providecommand{\doi}[1]{doi: #1}\else
  \providecommand{\doi}{doi: \begingroup \urlstyle{rm}\Url}\fi

\bibitem[Bender et~al.(2018)Bender, Kindermans, Zoph, Vasudevan, and
  Le]{bender_icml:2018}
Gabriel Bender, Pieter-Jan Kindermans, Barret Zoph, Vijay Vasudevan, and Quoc
  Le.
\newblock Understanding and simplifying one-shot architecture search.
\newblock In \emph{International Conference on Machine Learning}, 2018.

\bibitem[Brock et~al.(2018)Brock, Lim, Ritchie, and Weston]{brock2018smash}
Andrew Brock, Theo Lim, J.M. Ritchie, and Nick Weston.
\newblock {SMASH}: One-shot model architecture search through hypernetworks.
\newblock In \emph{International Conference on Learning Representations}, 2018.
\newblock URL \url{https://openreview.net/forum?id=rydeCEhs-}.

\bibitem[Cai et~al.(2019)Cai, Zhu, and Han]{Cai19}
Han Cai, Ligeng Zhu, and Song Han.
\newblock Proxylessnas: Direct neural architecture search on target task and
  hardware.
\newblock In \emph{International Conference on Learning Representations}, 2019.

\bibitem[Casale et~al.(2019)Casale, Gordon, and Fusi]{casale2019}
Francesco~Paolo Casale, Jonathan Gordon, and Nicolo Fusi.
\newblock Probabilistic neural architecture search.
\newblock \emph{arXiv preprint arXiv:1902.05116}, 2019.

\bibitem[DeVries \& Taylor(2017)DeVries and Taylor]{cutout}
Terrance DeVries and Graham~W Taylor.
\newblock Improved regularization of convolutional neural networks with cutout.
\newblock \emph{arXiv preprint arXiv:1708.04552}, 2017.

\bibitem[Dong \& Yang(2019)Dong and Yang]{dong2019search}
Xuanyi Dong and Yi~Yang.
\newblock Searching for a robust neural architecture in four gpu hours.
\newblock In \emph{Proceedings of the IEEE Conference on Computer Vision and
  Pattern Recognition (CVPR)}, pp.\  1761--1770, 2019.

\bibitem[Dong \& Yang(2020)Dong and Yang]{dong20}
Xuanyi Dong and Yi~Yang.
\newblock Nas-bench-102: Extending the scope of reproducible neural
  architecture search.
\newblock In \emph{International Conference on Learning Representations}, 2020.
\newblock URL \url{https://openreview.net/forum?id=HJxyZkBKDr}.

\bibitem[Elsken et~al.(2019)Elsken, Metzen, and Hutter]{Elsken19}
Thomas Elsken, Jan~Hendrik Metzen, and Frank Hutter.
\newblock Efficient multi-objective neural architecture search via lamarckian
  evolution.
\newblock In \emph{International Conference on Learning Representations}, 2019.

\bibitem[Eric~Jang \& Poole(2017)Eric~Jang and Poole]{gumbel}
Shixiang~Gu Eric~Jang and Ben Poole.
\newblock Categorical reparameterization with gumbel-softmax.
\newblock In \emph{International Conference on Learning Representations}, 2017.

\bibitem[Falkner et~al.(2018)Falkner, Klein, and Hutter]{Falkner18}
Stefan Falkner, Aaron Klein, and Frank Hutter.
\newblock {BOHB}: Robust and efficient hyperparameter optimization at scale.
\newblock In Jennifer Dy and Andreas Krause (eds.), \emph{Proceedings of the
  35th International Conference on Machine Learning}, volume~80 of
  \emph{Proceedings of Machine Learning Research}, pp.\  1437--1446,
  Stockholmsm{\"{a}}ssan, Stockholm Sweden, 10--15 Jul 2018. PMLR.
\newblock URL \url{http://proceedings.mlr.press/v80/falkner18a.html}.

\bibitem[Hutter et~al.(2014)Hutter, Hoos, and Leyton-Brown]{hutter-icml14a}
F.~Hutter, H.~Hoos, and K.~Leyton-Brown.
\newblock An efficient approach for assessing hyperparameter importance.
\newblock In E.~Xing and T.~Jebara (eds.), \emph{Proceedings of the 31th
  International Conference on Machine Learning, (ICML'14)}, pp.\  754--762.
  Omnipress, 2014.

\bibitem[Jamieson \& Talwalkar(2016)Jamieson and Talwalkar]{jamieson-aistats16}
K.~Jamieson and A.~Talwalkar.
\newblock Non-stochastic best arm identification and hyperparameter
  optimization.
\newblock In \emph{Proceedings of the Seventeenth International Conference on
  Artificial Intelligence and Statistics (AISTATS)}, 2016.

\bibitem[Krizhevsky(2009)]{krizhevsky-tech09a}
A.~Krizhevsky.
\newblock Learning multiple layers of features from tiny images.
\newblock Technical report, University of Toronto, 2009.

\bibitem[Li et~al.(2017)Li, Jamieson, DeSalvo, Rostamizadeh, and
  Talwalkar]{li_iclr17}
L.~Li, K.~Jamieson, G.~DeSalvo, A.~Rostamizadeh, and A.~Talwalkar.
\newblock Hyperband: Bandit-based configuration evaluation for hyperparameter
  optimization.
\newblock In \emph{International Conference on Learning Representations}, 2017.

\bibitem[Li \& Talwalkar(2019)Li and Talwalkar]{li2019random}
Liam Li and Ameet Talwalkar.
\newblock Random search and reproducibility for neural architecture search.
\newblock In \emph{Proceedings of the Thirty-Fifth Conference on Uncertainty in
  Artificial Intelligence, {UAI} 2019, Tel Aviv, Israel, July 22-25, 2019},
  pp.\  129, 2019.
\newblock URL \url{http://auai.org/uai2019/proceedings/papers/129.pdf}.

\bibitem[Liang et~al.(2017)Liang, Liaw, Nishihara, Moritz, Fox, Gonzalez,
  Goldberg, and Stoica]{rllib}
Eric Liang, Richard Liaw, Robert Nishihara, Philipp Moritz, Roy Fox, Joseph
  Gonzalez, Ken Goldberg, and Ion Stoica.
\newblock Ray rllib: {A} composable and scalable reinforcement learning
  library.
\newblock \emph{CoRR}, abs/1712.09381, 2017.
\newblock URL \url{http://arxiv.org/abs/1712.09381}.

\bibitem[Lindauer \& Hutter(2019)Lindauer and Hutter]{best_practices}
Marius Lindauer and Frank Hutter.
\newblock Best practices for scientific research on neural architecture search.
\newblock \emph{arXiv preprint arXiv:1909.02453}, 2019.

\bibitem[Liu et~al.(2019)Liu, Simonyan, and Yang]{darts}
Hanxiao Liu, Karen Simonyan, and Yiming Yang.
\newblock {DARTS}: Differentiable architecture search.
\newblock In \emph{International Conference on Learning Representations}, 2019.

\bibitem[Luo et~al.(2018)Luo, Tian, Qin, and Liu]{Luo2018NeuralAO}
Renqian Luo, Fei Tian, Tao Qin, and T.~M. Liu.
\newblock Neural architecture optimization.
\newblock In \emph{NeurIPS}, 2018.

\bibitem[Pham et~al.(2018)Pham, Guan, Zoph, Le, and Dean]{Pham18}
Hieu Pham, Melody~Y. Guan, Barret Zoph, Quoc~V. Le, and Jeff Dean.
\newblock Efficient neural architecture search via parameter sharing.
\newblock In \emph{International Conference on Machine Learning}, 2018.

\bibitem[Real et~al.(2017)Real, Moore, Selle, Saxena, Suematsu, Tan, Le, and
  Kurakin]{real17a}
Esteban Real, Sherry Moore, Andrew Selle, Saurabh Saxena, Yutaka~Leon Suematsu,
  Jie Tan, Quoc~V. Le, and Alexey Kurakin.
\newblock Large-scale evolution of image classifiers.
\newblock In Doina Precup and Yee~Whye Teh (eds.), \emph{Proceedings of the
  34th International Conference on Machine Learning}, volume~70 of
  \emph{Proceedings of Machine Learning Research}, pp.\  2902--2911,
  International Convention Centre, Sydney, Australia, 06--11 Aug 2017. PMLR.
\newblock URL \url{http://proceedings.mlr.press/v70/real17a.html}.

\bibitem[Real et~al.(2019)Real, Aggarwal, Huang, and Le]{real-arXiv18a}
Esteban Real, Alok Aggarwal, Yanping Huang, and Quoc~V. Le.
\newblock Aging {Evolution} for {Image} {Classifier} {Architecture} {Search}.
\newblock In \emph{AAAI}, 2019.

\bibitem[Saxena \& Verbeek(2016)Saxena and Verbeek]{SaxenaV16}
Shreyas Saxena and Jakob Verbeek.
\newblock Convolutional neural fabrics.
\newblock In D.~D. Lee, M.~Sugiyama, U.~V. Luxburg, I.~Guyon, and R.~Garnett
  (eds.), \emph{Advances in Neural Information Processing Systems 29}, pp.\
  4053--4061. Curran Associates, Inc., 2016.

\bibitem[Srivastava et~al.(2014)Srivastava, Hinton, Krizhevsky, Sutskever, and
  Salakhutdinov]{srivastava-jmlr14a}
N.~Srivastava, G.~Hinton, A.~Krizhevsky, I.~Sutskever, and R.~Salakhutdinov.
\newblock Dropout: A simple way to prevent neural networks from overfitting.
\newblock \emph{Journal of Machine Learning Research}, 15:\penalty0 1929--1958,
  2014.

\bibitem[Williams et~al.(2000)Williams, Santner, and Notz]{williams}
B.~Williams, T.~Santner, and W.~Notz.
\newblock {Sequential design of computer experiments to minimize integrated
  response functions}.
\newblock \emph{Statistica Sinica}, 2000.

\bibitem[Xie et~al.(2019)Xie, Zheng, Liu, and Lin]{Xie18}
Sirui Xie, Hehui Zheng, Chunxiao Liu, and Liang Lin.
\newblock {SNAS}: stochastic neural architecture search.
\newblock In \emph{International Conference on Learning Representations}, 2019.

\bibitem[Xu et~al.(2020)Xu, Xie, Zhang, Chen, Qi, Tian, and
  Xiong]{xu2019pcdarts}
Yuhui Xu, Lingxi Xie, Xiaopeng Zhang, Xin Chen, Guo-Jun Qi, Qi~Tian, and
  Hongkai Xiong.
\newblock Pc-darts: Partial channel connections for memory-efficient
  architecture search.
\newblock In \emph{International Conference on Learning Representations}, 2020.
\newblock URL \url{https://openreview.net/forum?id=BJlS634tPr}.

\bibitem[Yang et~al.(2020)Yang, Esperança, and Carlucci]{Yang2020NAS}
Antoine Yang, Pedro~M. Esperança, and Fabio~M. Carlucci.
\newblock Nas evaluation is frustratingly hard.
\newblock In \emph{International Conference on Learning Representations}, 2020.
\newblock URL \url{https://openreview.net/forum?id=HygrdpVKvr}.

\bibitem[Ying et~al.(2019)Ying, Klein, Christiansen, Real, Murphy, and
  Hutter]{ying19a}
Chris Ying, Aaron Klein, Eric Christiansen, Esteban Real, Kevin Murphy, and
  Frank Hutter.
\newblock {NAS}-bench-101: Towards reproducible neural architecture search.
\newblock In Kamalika Chaudhuri and Ruslan Salakhutdinov (eds.),
  \emph{Proceedings of the 36th International Conference on Machine Learning},
  volume~97 of \emph{Proceedings of Machine Learning Research}, pp.\
  7105--7114, Long Beach, California, USA, 09--15 Jun 2019. PMLR.
\newblock URL \url{http://proceedings.mlr.press/v97/ying19a.html}.

\bibitem[Yu et~al.(2020)Yu, Sciuto, Jaggi, Musat, and Salzmann]{sciuto19}
Kaicheng Yu, Christian Sciuto, Martin Jaggi, Claudiu Musat, and Mathieu
  Salzmann.
\newblock Evaluating the search phase of neural architecture search.
\newblock In \emph{International Conference on Learning Representations}, 2020.
\newblock URL \url{https://openreview.net/forum?id=H1loF2NFwr}.

\bibitem[Zela et~al.(2020)Zela, Elsken, Saikia, Marrakchi, Brox, and
  Hutter]{zela19}
Arber Zela, Thomas Elsken, Tonmoy Saikia, Yassine Marrakchi, Thomas Brox, and
  Frank Hutter.
\newblock Understanding and robustifying differentiable architecture search.
\newblock In \emph{International Conference on Learning Representations}, 2020.
\newblock URL \url{https://openreview.net/forum?id=H1gDNyrKDS}.

\bibitem[Zoph \& Le(2017)Zoph and Le]{zoph-iclr17}
Barret Zoph and Quoc~V. Le.
\newblock Neural architecture search with reinforcement learning.
\newblock In \emph{International Conference on Learning Representations (ICLR)
  2017 Conference Track}, 2017.

\bibitem[Zoph et~al.(2018)Zoph, Vasudevan, Shlens, and Le]{zoph-arXiv18}
Barret Zoph, Vijay Vasudevan, Jonathon Shlens, and Quoc~V. Le.
\newblock Learning transferable architectures for scalable image recognition.
\newblock In \emph{Conference on Computer Vision and Pattern Recognition},
  2018.

\end{thebibliography}
\bibliographystyle{iclr2020_conference}

\clearpage
\appendix

\section{Details on search spaces}\label{sec: search_spaces_details}
\begin{figure}[b]
\centering
\begin{subfigure}{.33\textwidth}
  \centering
  \includegraphics[width=1.0\linewidth]{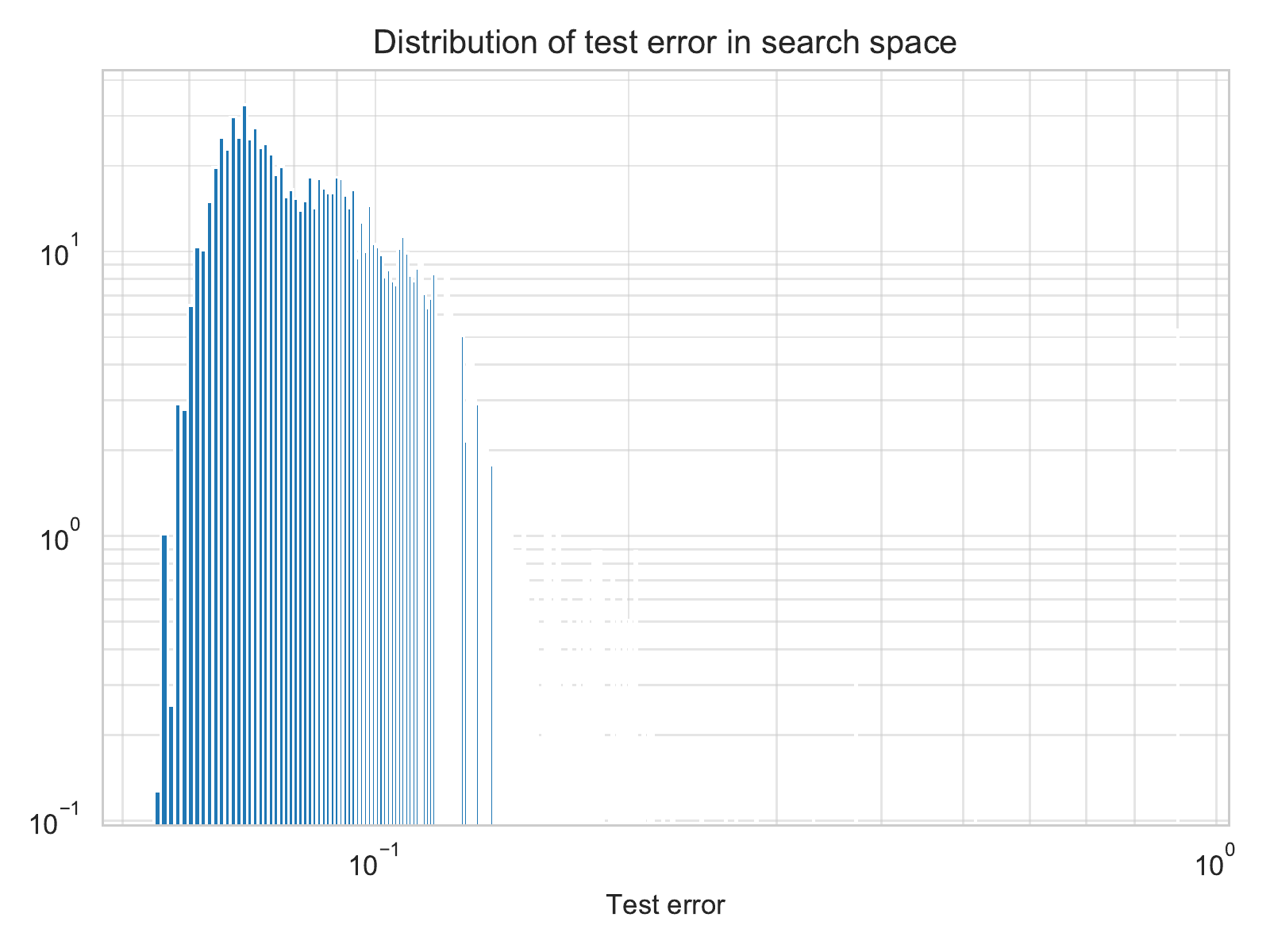}
  \caption{Search space 1}
  \label{fig:sec2:ss1_distr}
\end{subfigure}%
\begin{subfigure}{.33\textwidth}
  \centering
  \includegraphics[width=1.0\linewidth]{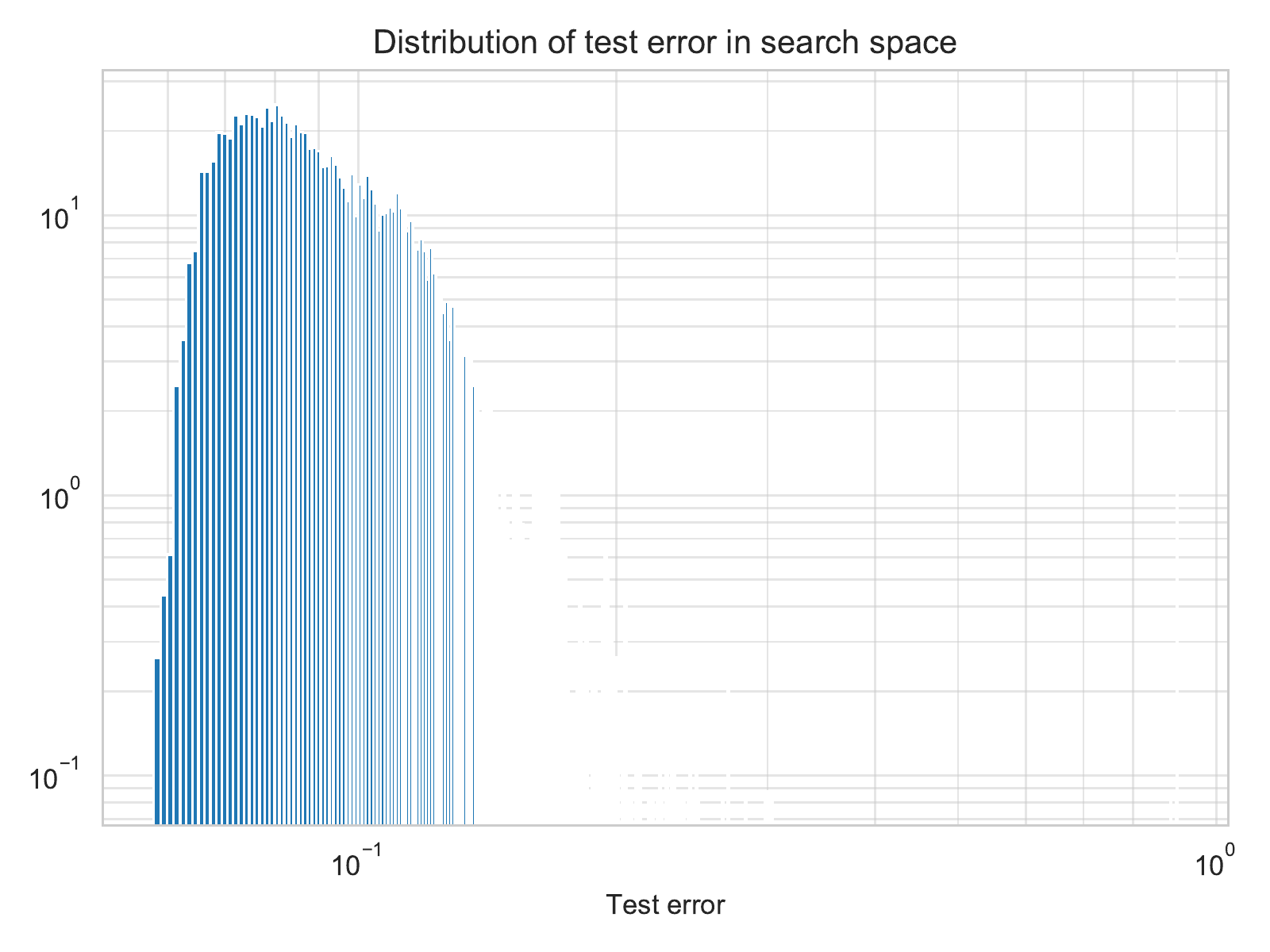}
  \caption{Search space 2}
  \label{fig:sec2:ss2_distr}
\end{subfigure}
\begin{subfigure}{.33\textwidth}
  \centering
  \includegraphics[width=1.0\linewidth]{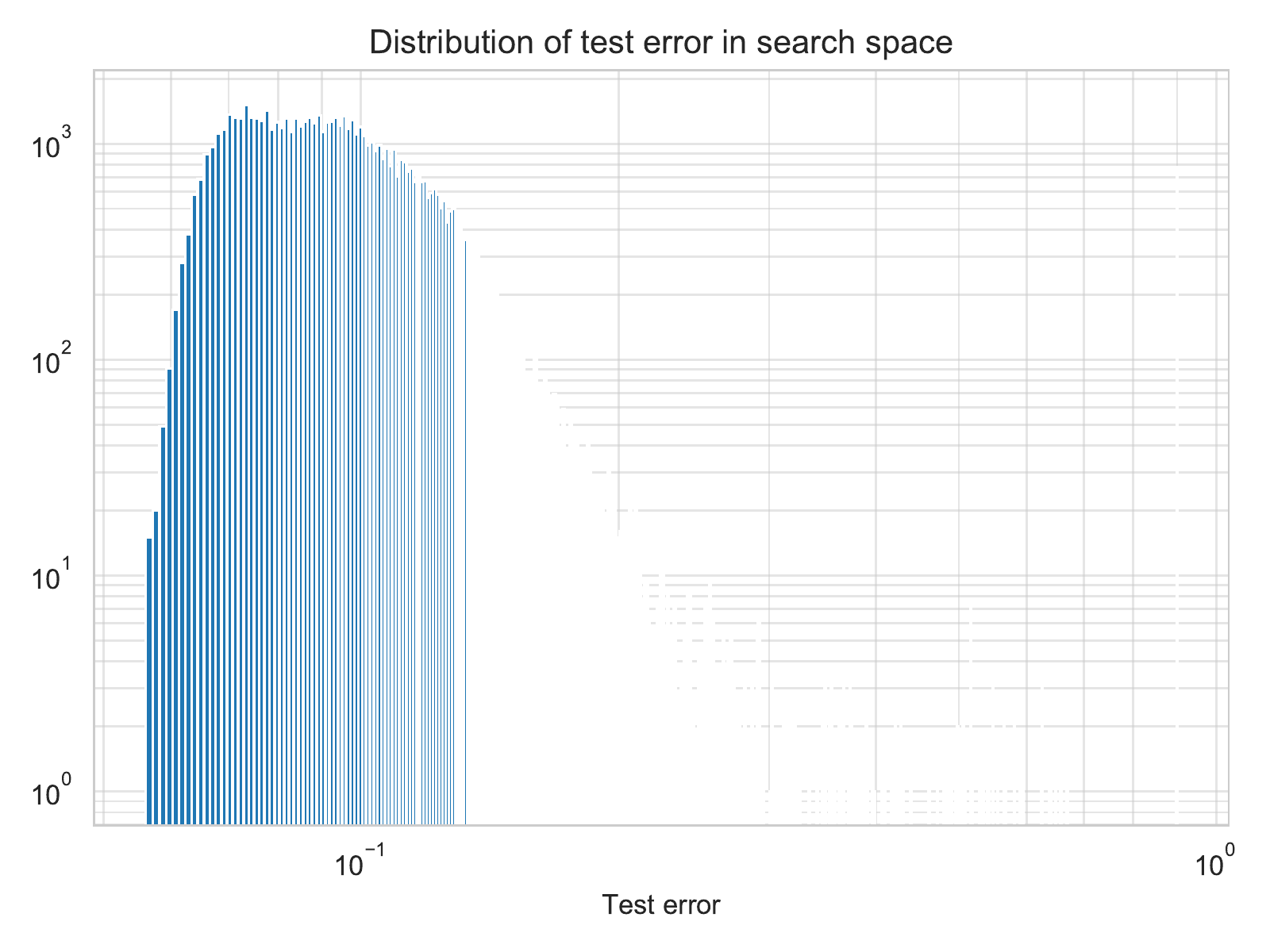}
  \caption{Search space 3}
  \label{fig:sec2:ss3_distr}
\end{subfigure}
\caption{Distribution of test error in the search spaces with loose ends.}
\label{fig:test}
\end{figure}

\paragraph{Search space 1} The main characteristic of this search space is that the number of parents for each choice block and output has to be exactly 2 (apart from choice block 1 which is only connected to the input). Because of this requirement one choice block had to be discarded as that would exceed the requirement to have at most 9 edges. The total distribution of test error in shown in Figure \ref{fig:sec2:ss1_distr}. It is the smallest search space discussed in this work.

\paragraph{Search space 2} This search space is related to search space 1 in that it only consists of 4 intermediate nodes, but in contrast the output has three parents and nodes 1 and 2 only one parent. This increases the number of architectures in this space compared to search space 1. The test error distribution is shown in Figure \ref{fig:sec2:ss2_distr}.

\paragraph{Search space 3} All available intermediate nodes (up to 5) are used in this search space, making this search space the largest, but also the search space where each node has on average the least number of parents. The test error distribution is shown in Figure \ref{fig:sec2:ss3_distr}.

\section{Optimizers}
\label{sec:app:optimizers}
\textbf{DARTS \citep{darts}} uses a weighted continuous relaxation over the operations to learn an architecture by solving a bilevel optimization problem. The training dataset is split in two parts, one used for updating the parameters of the operations in the one-shot model, and the other to update the weights appended to operations, that determine the importance of that operation.

For evaluation, we choose the parents of each choice block based on the highest architectural weights and the number of parents for that choice block given by Table \ref{table:search_spaces}. We pick the highest weighted operation from the choice block.

\textbf{GDAS \citep{dong2019search}} modifies DARTS, such that individual paths are sampled differentiably through each cell using Gumbel-Softmax \citep{gumbel} to adapt the architecture weights. This reduces the memory overhead created by DARTS as only the sampled paths have to be evaluated. GDAS uses the same search space and evaluation procedure as DARTS.

\textbf{PC-DARTS \citep{xu2019pcdarts}} reduces the memory overhead by only evaluating a random fraction of the channels with the mixed-ops. The authors argue that this also regularizes the search as it lowers the bias towards weight-free operations such as skip-connect and max-pooling, which are often preferred early on in DARTS search. In addition to partial channel connections, the authors propose edge normalization, which adds additional architectural parameters to the edges connecting to an intermediate node. This is done to compensate for the added fluctuations due to the partial channel connections. These additional weights are already part of the search spaces we proposed in this paper.

\textbf{Random Search with Weight Sharing (Random WS) \citep{li2019random}} randomly samples architectures from the one-shot model for each training mini-batch and trains only the selected subset of the one-shot model on that mini-batch. Differently from DARTS, PC-DARTS and GDAS, Random WS does not require a validation set, since there are no architectural weights that need to be updated. For evaluation Random WS samples 1000 architectures from the search space and evaluates each for only a small number of batches on the validation set using the optimized weights of the one-shot model corresponding to the sub-networks. Then the 5 architectures with the lowest one-shot validation error are chosen and fully evaluated on the validation set. The overall best architecture is returned.

\textbf{ENAS \citep{Pham18}} similarly to Random WS samples sub-networks of in the one-shot model, however by means of a recurrent neural network (RNN) controller rather than randomly. As the search progresses the parameters of the RNN controller are updated via REINFORCE~\citep{williams} using the validation error of the sampled architectures as a reward. This way the sampling procedure is handled in a more effective way.

\section{Hyperparameters}\label{sec: hyperparameters}
If not stated otherwise the following hyperparameters were used for all our experiments. We used a batch size of 96 throughout for DARTS, GDAS and PC-DARTS as the search spaces are small enough to allow it and as this reduces the randomness in the training, which makes the comparison between optimizers easier. Random WS was trained with a batch size of 64. All other hyperparameters were adapted from DARTS~\citep{darts}.

\section{Comparison of optimizers over different budgets}
\label{sec: more_search_epochs}

\begin{figure}[ht]
\centering
\begin{subfigure}{.33\textwidth}
  \centering
\includegraphics[width=0.99\textwidth]{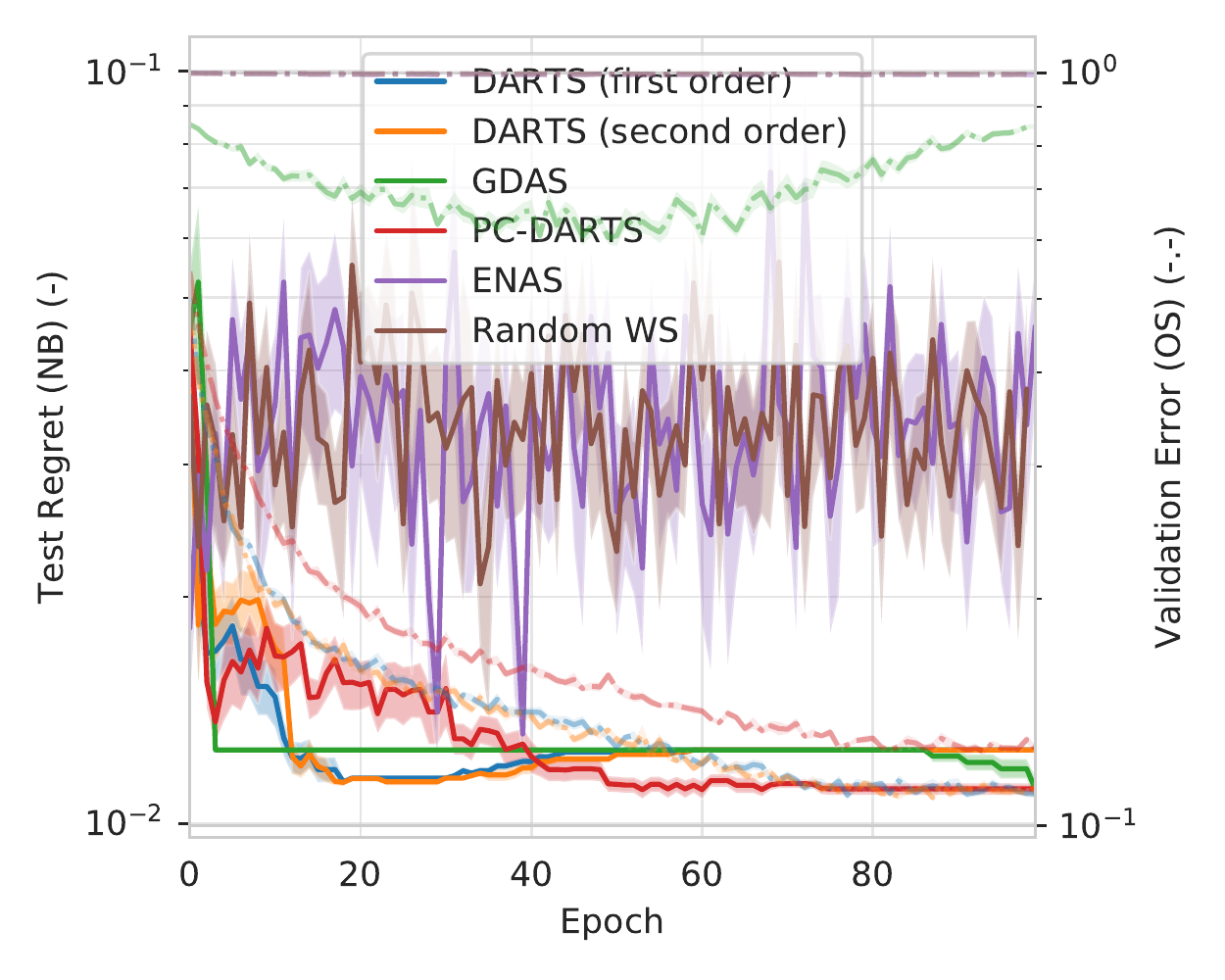}
  \caption{Search space 1}
  \label{fig:sec3:ss1_opt_100}
\end{subfigure}%
\begin{subfigure}{.33\textwidth}
  \centering
\includegraphics[width=0.99\textwidth]{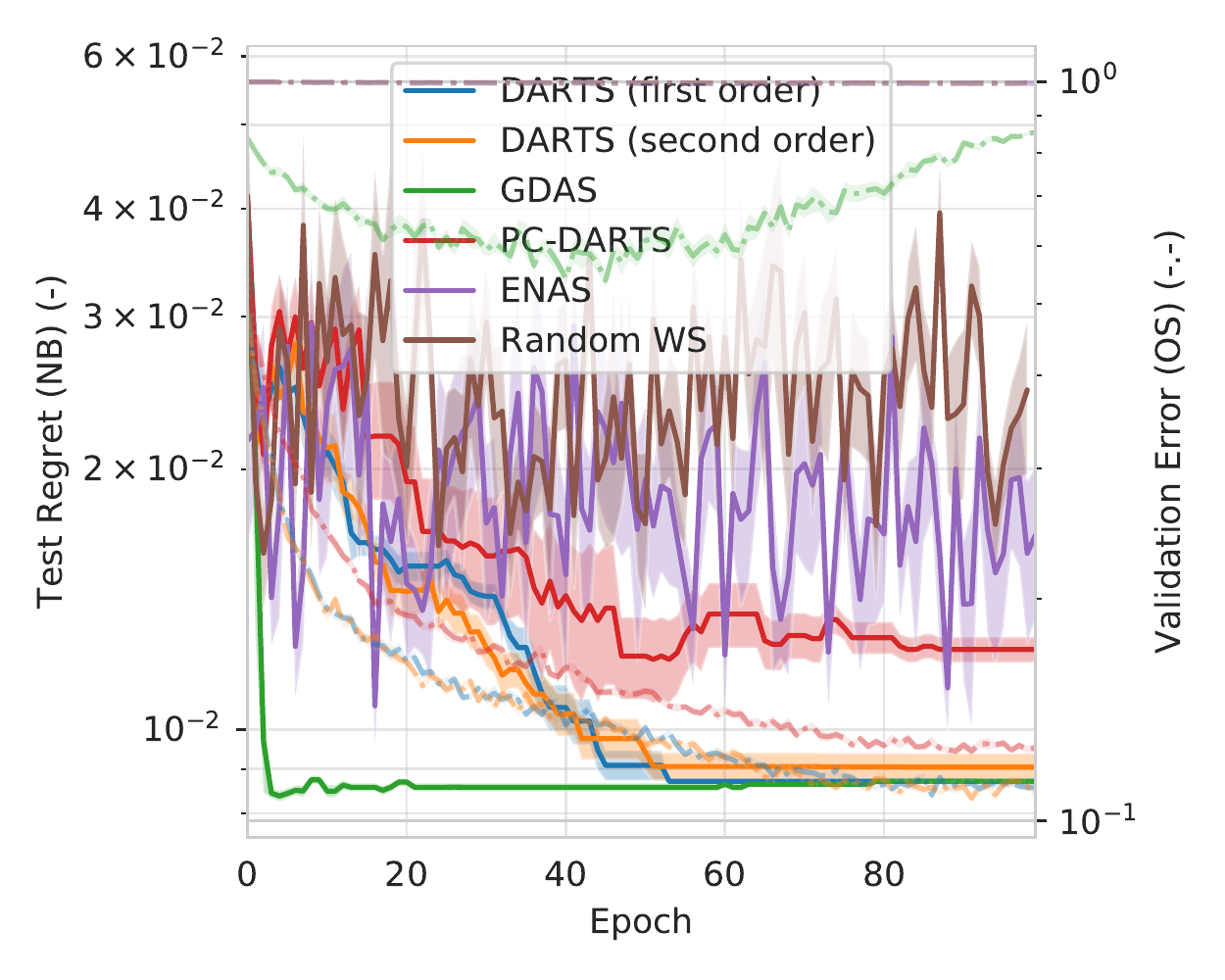}
  \caption{Search space 2}
  \label{fig:sec3:ss2_opt_100}
\end{subfigure}
\begin{subfigure}{.33\textwidth}
  \centering
\includegraphics[width=0.99\textwidth]{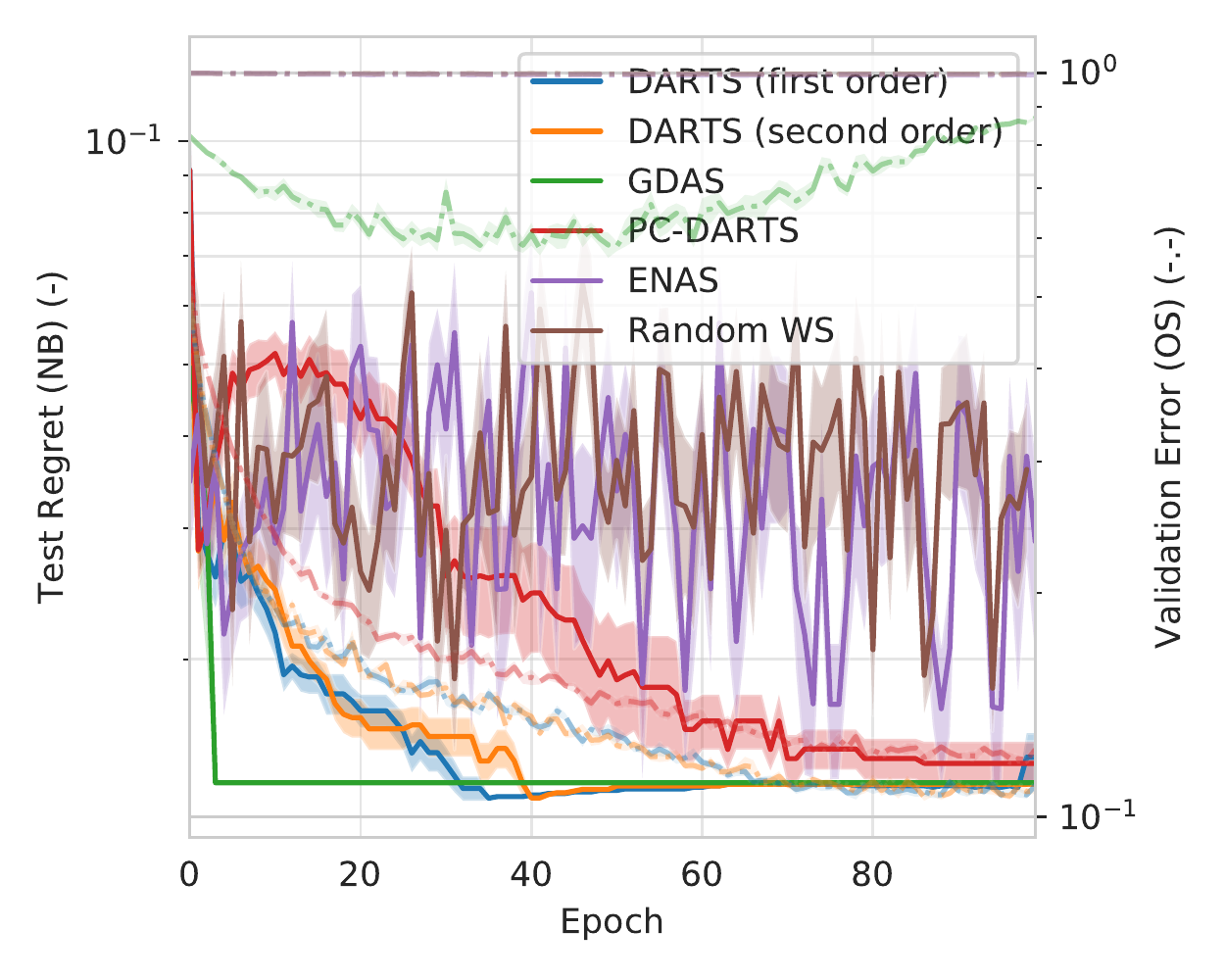}
  \caption{Search space 3}
  \label{fig:sec3:ss3_opt_100}
\end{subfigure}
\caption{Comparison of different One-Shot Neural Architecture optimizers on the three different search spaces defined on NAS-Bench-101 over 100 epochs.}
\label{fig:sec3:optimizers:comparison_100_epochs}
\end{figure}

\begin{figure}[ht]
\centering
\begin{subfigure}{.33\textwidth}
  \centering
\includegraphics[width=0.99\textwidth]{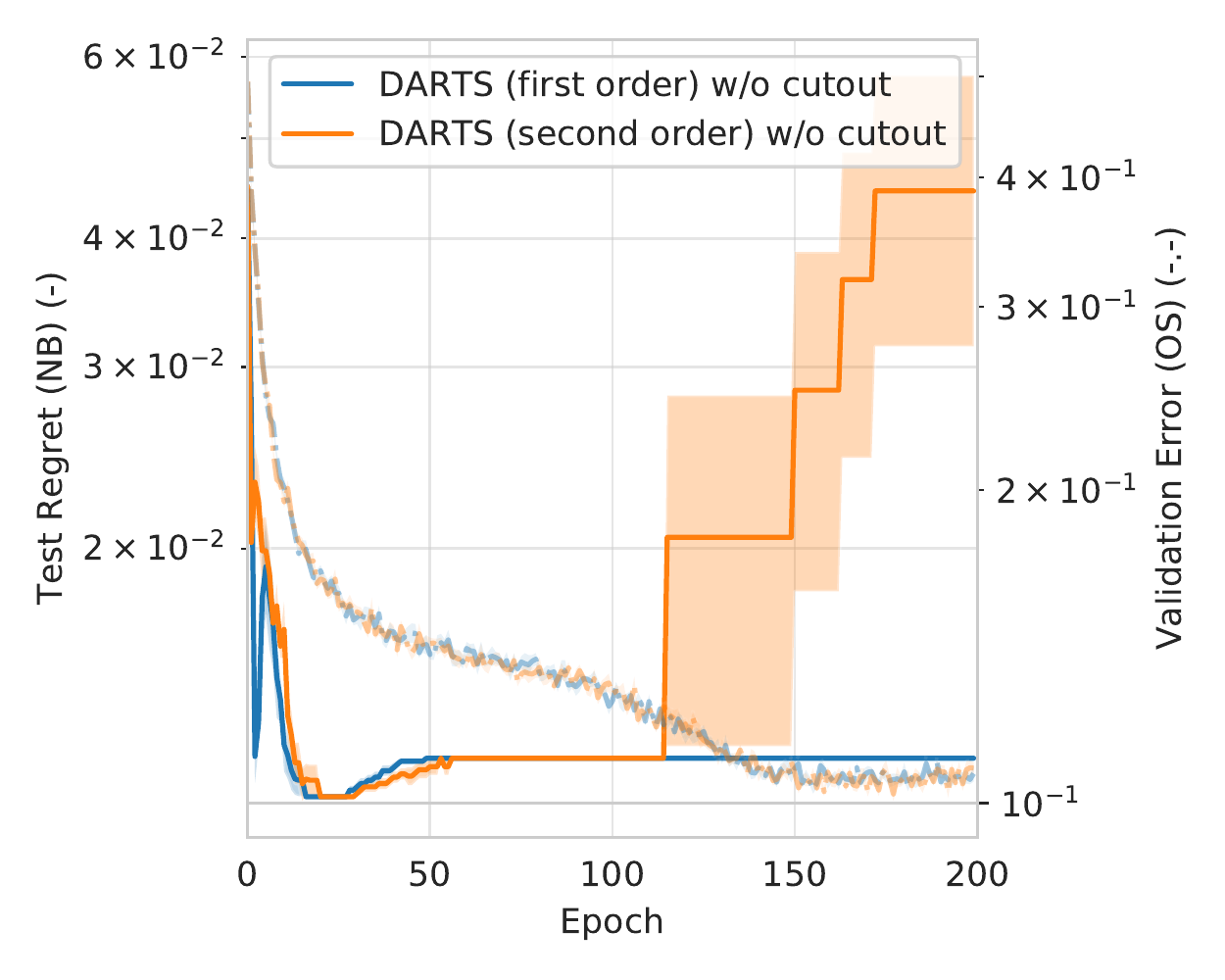}
  \caption{Search space 1}
  \label{fig:sec3:ss1_opt-200}
\end{subfigure}%
\begin{subfigure}{.33\textwidth}
  \centering
\includegraphics[width=0.99\textwidth]{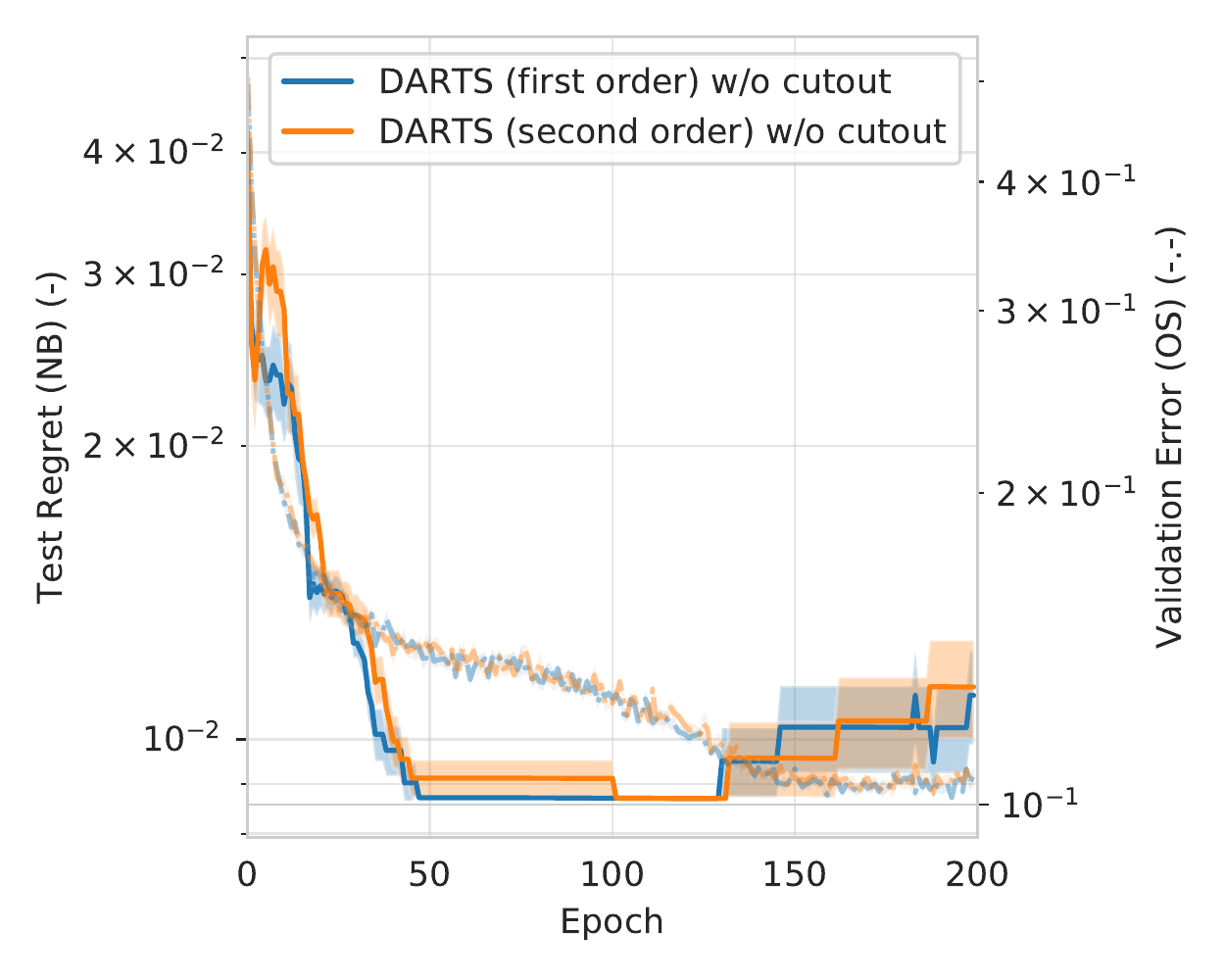}
  \caption{Search space 2}
  \label{fig:sec3:ss2_opt-200}
\end{subfigure}
\begin{subfigure}{.33\textwidth}
  \centering
\includegraphics[width=0.99\textwidth]{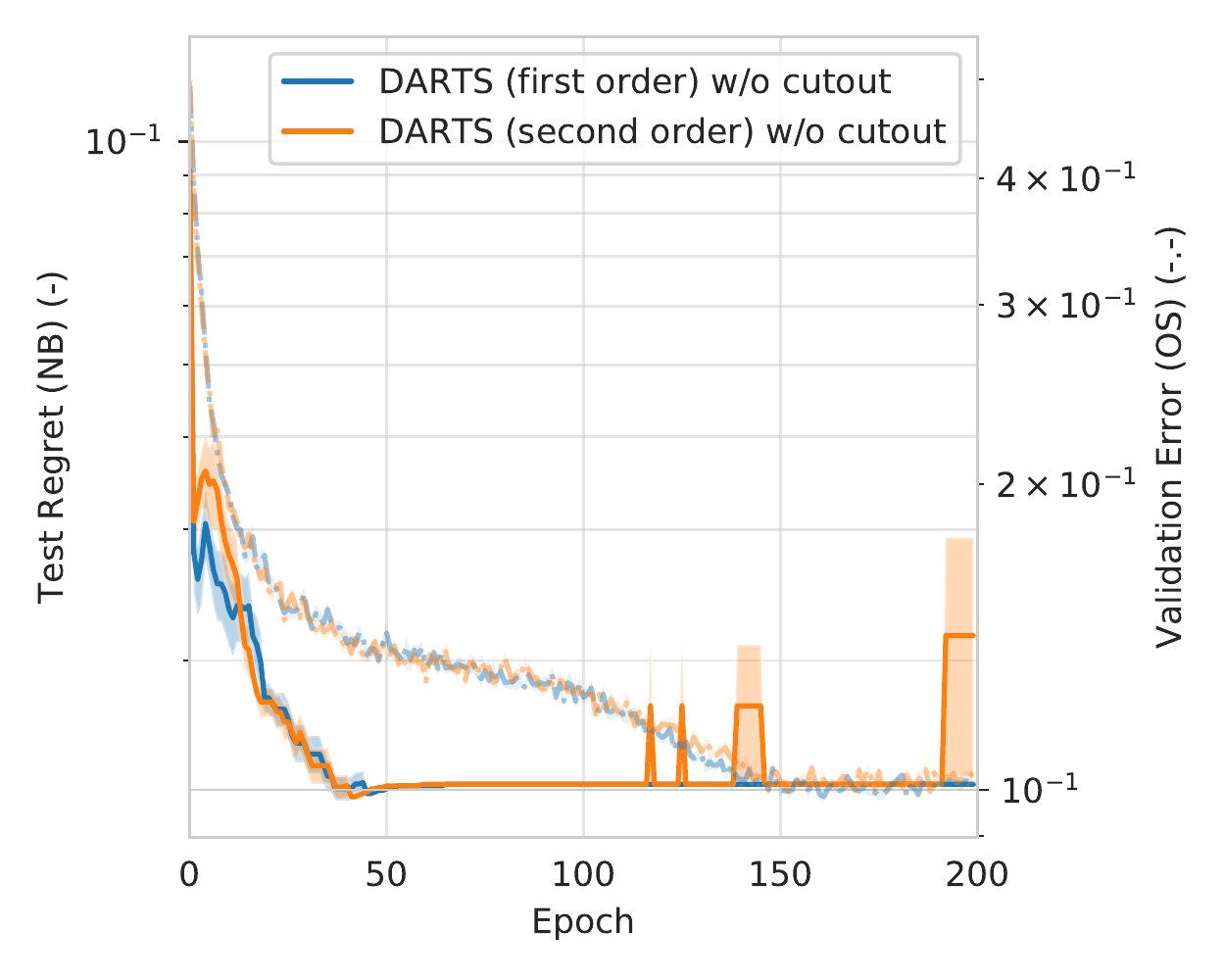}
  \caption{Search space 3}
  \label{fig:sec3:ss3_opt-200}
\end{subfigure}
\caption{Comparison of DARTS first and second order on the three different search spaces defined on NAS-Bench-101 for 200 epochs epochs.}
\label{fig:sec3:optimizers:comparison_200}
\end{figure}

\section{Regularization}
\label{sec: regularization}

\subsection{Cutout}

\begin{figure}[h]
\centering
\begin{subfigure}{.33\textwidth}
  \centering
\includegraphics[width=0.99999\textwidth]{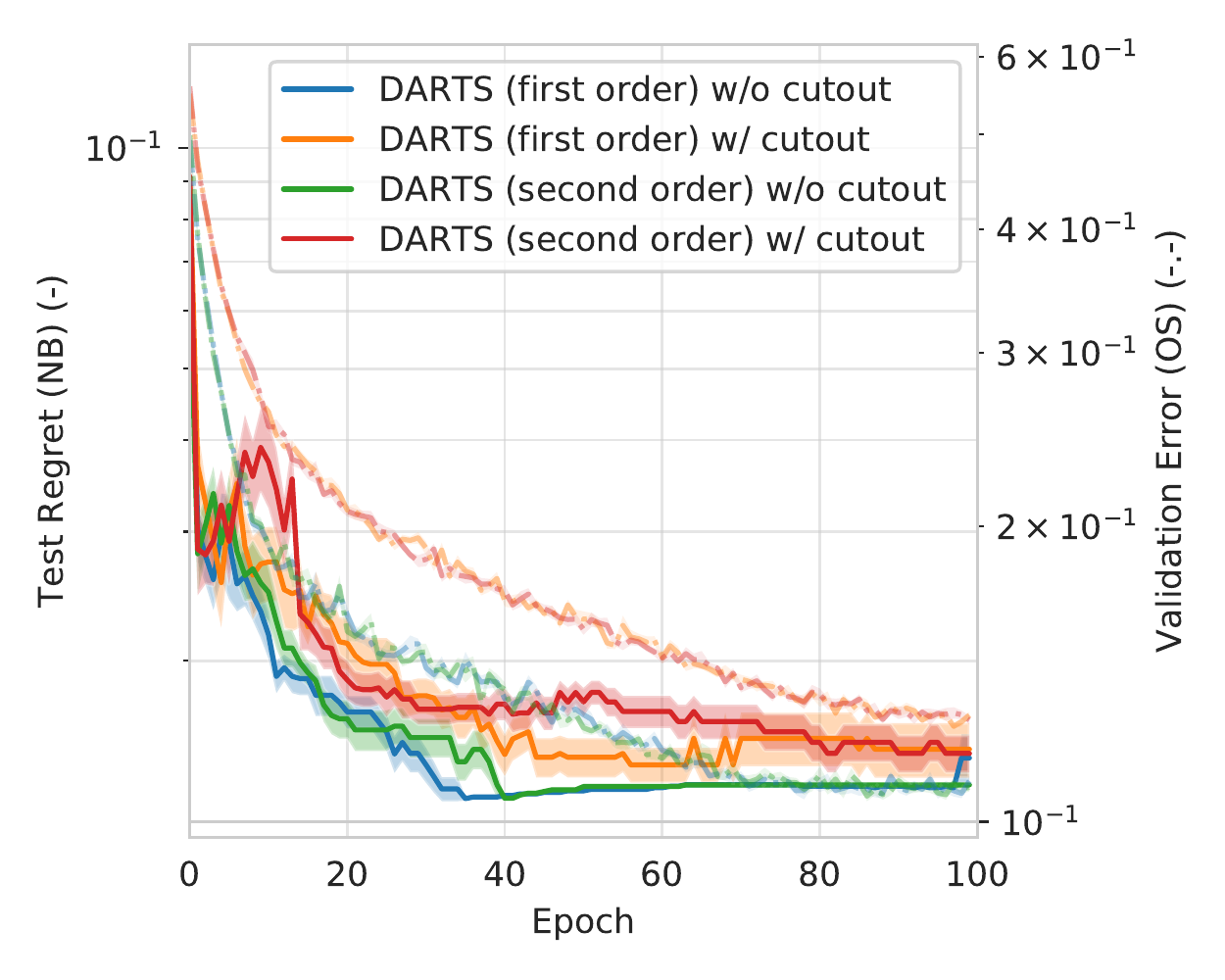}
  \caption{DARTS}
  \label{fig:sec3:ss3_cutout}
\end{subfigure}
\begin{subfigure}{.325\textwidth}
  \centering
\includegraphics[width=0.99999\textwidth]{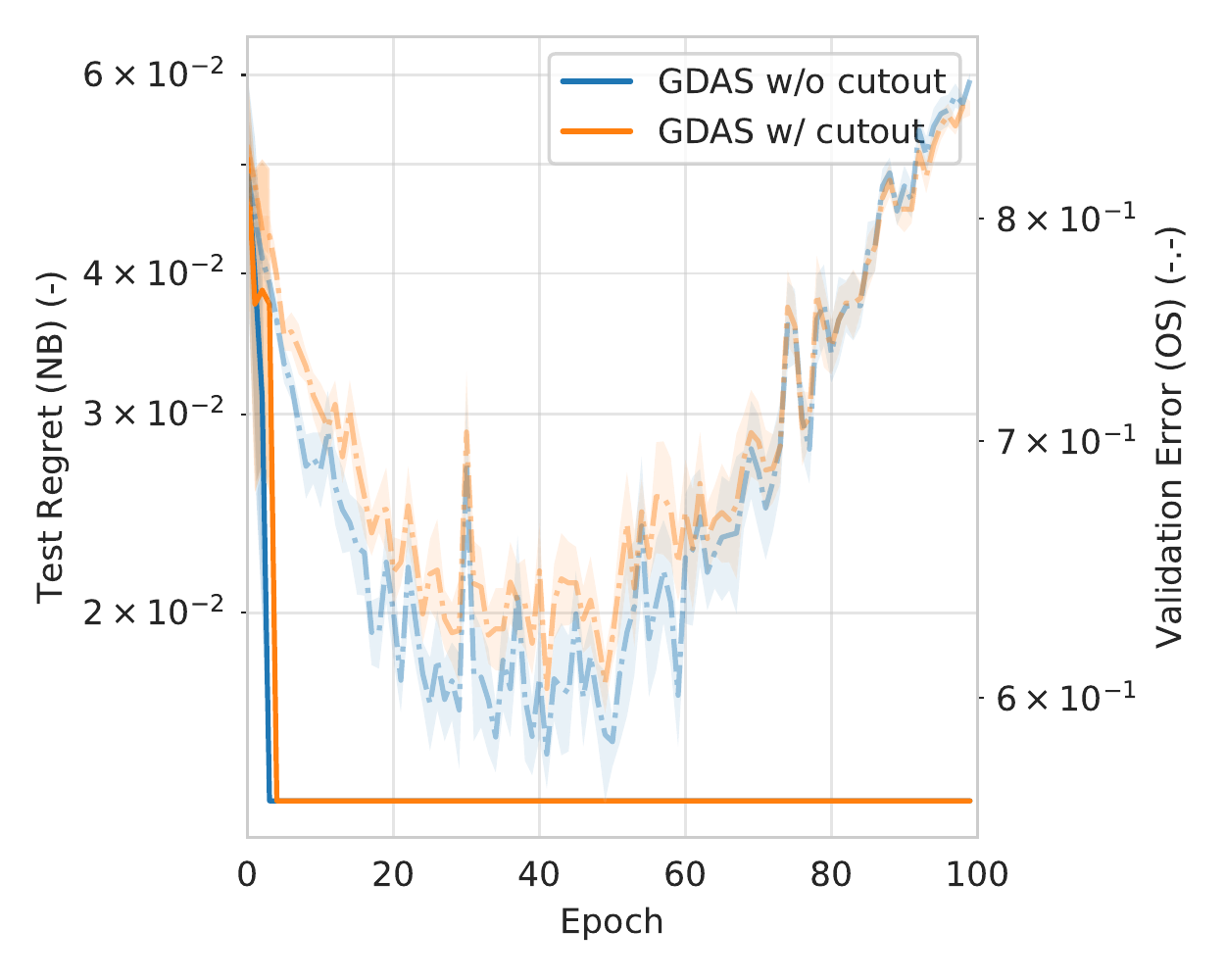}
  \caption{GDAS}
  \label{fig:sec3:reg:cutout_gdas_3}
\end{subfigure}
\begin{subfigure}{.33\textwidth}
  \centering
\includegraphics[width=0.99999\textwidth]{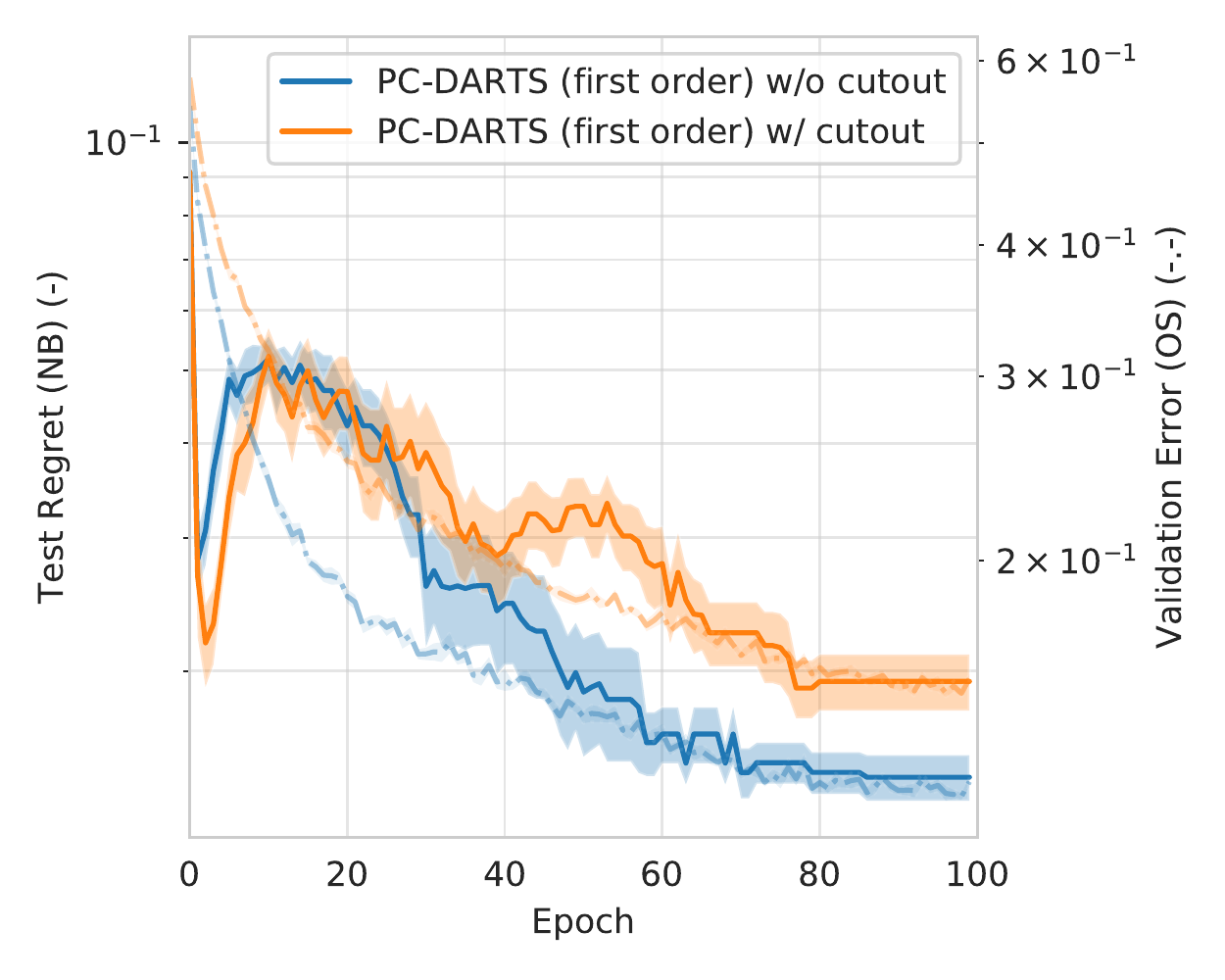}
  \caption{PC-DARTS}
  \label{fig:sec3:reg:cutout_pc_darts_3}
\end{subfigure}
\caption{Illustration of the impact that Cutout has on the test regret on NAS-Bench-101 and the validation error of the one-shot model using DARTS, GDAS and PC-DARTS on search space 3 (Best viewed in color).}
\label{fig:sec3:regularization:cutout}
\end{figure}

\begin{figure}[ht]
\centering
\begin{subfigure}{.33\textwidth}
  \centering
\includegraphics[width=0.99\textwidth]{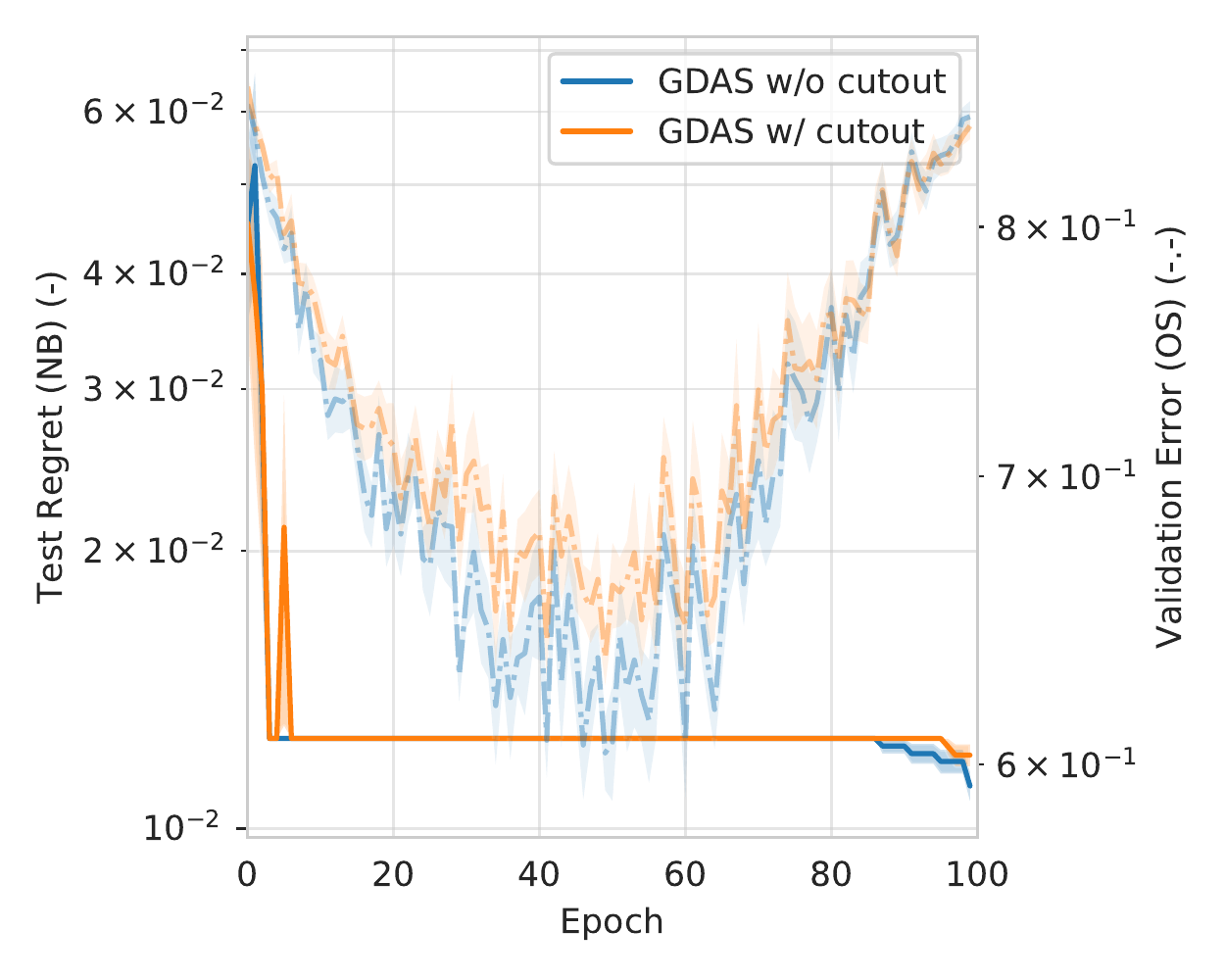}
  \caption{Search space 1}
  \label{fig:sec3:cutout_gdas_1}
\end{subfigure}%
\begin{subfigure}{.33\textwidth}
  \centering
\includegraphics[width=0.99\textwidth]{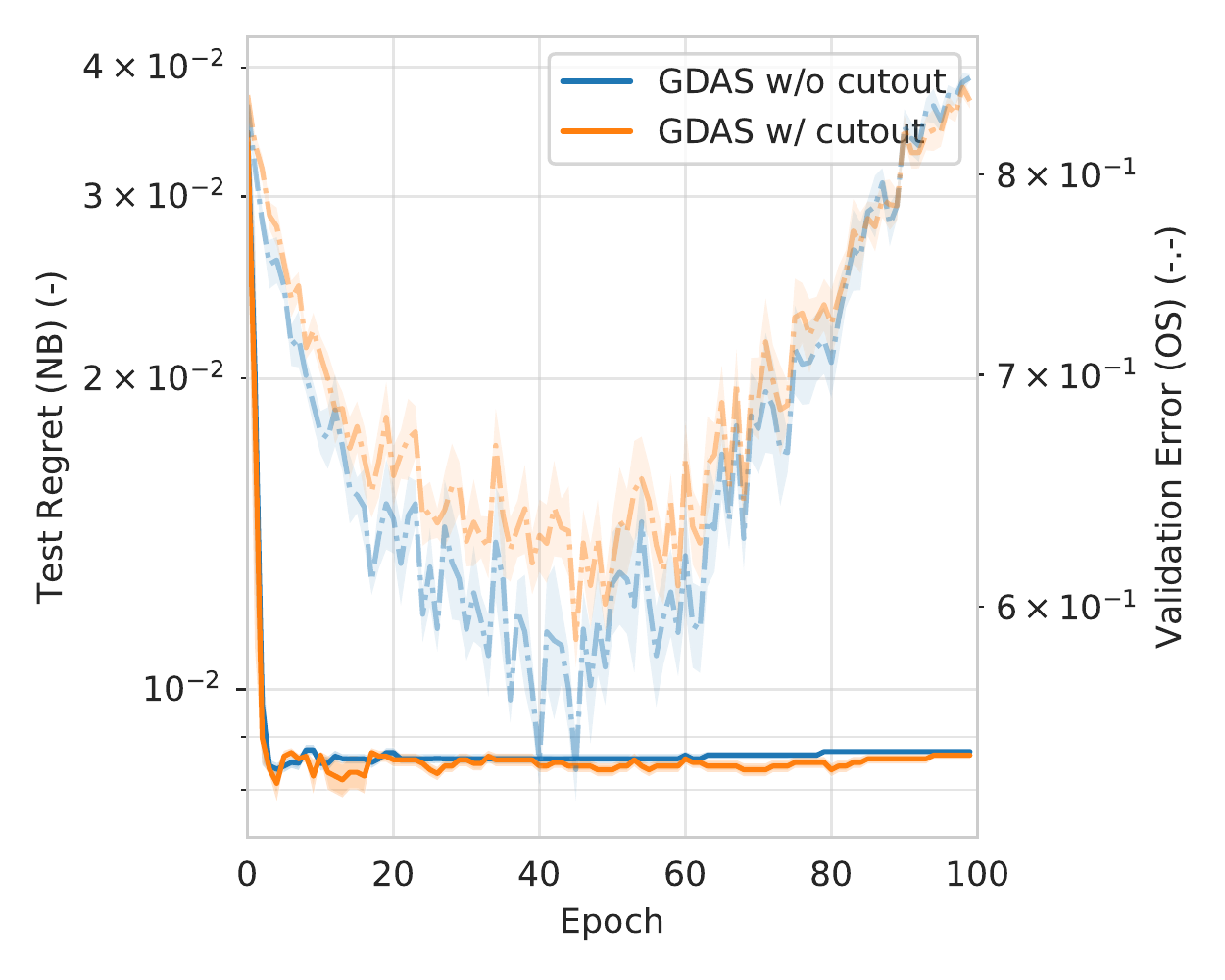}
  \caption{Search space 2}
  \label{fig:sec3:cutout_gdas_2}
\end{subfigure}
\caption{Comparison of the effect of using Cutout during architecture search on GDAS for search space 1 and 2.}
\label{fig:sec3:regularization:cutout_gdas}
\end{figure}

\begin{figure}[ht]
\centering
\begin{subfigure}{.33\textwidth}
  \centering
\includegraphics[width=0.99\textwidth]{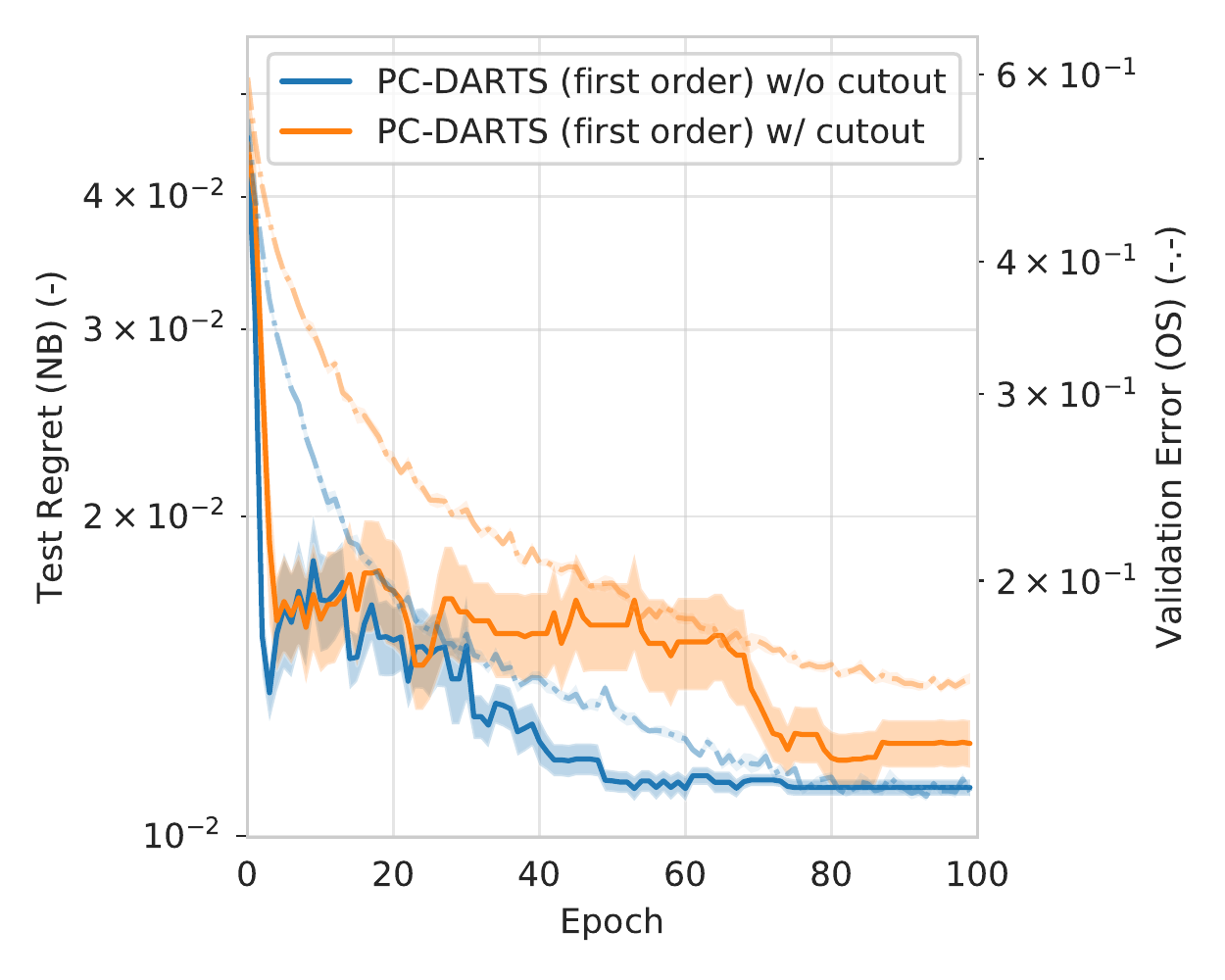}
  \caption{Search space 1}
  \label{fig:sec3:cutout_pc_darts_1}
\end{subfigure}%
\begin{subfigure}{.33\textwidth}
  \centering
\includegraphics[width=0.99\textwidth]{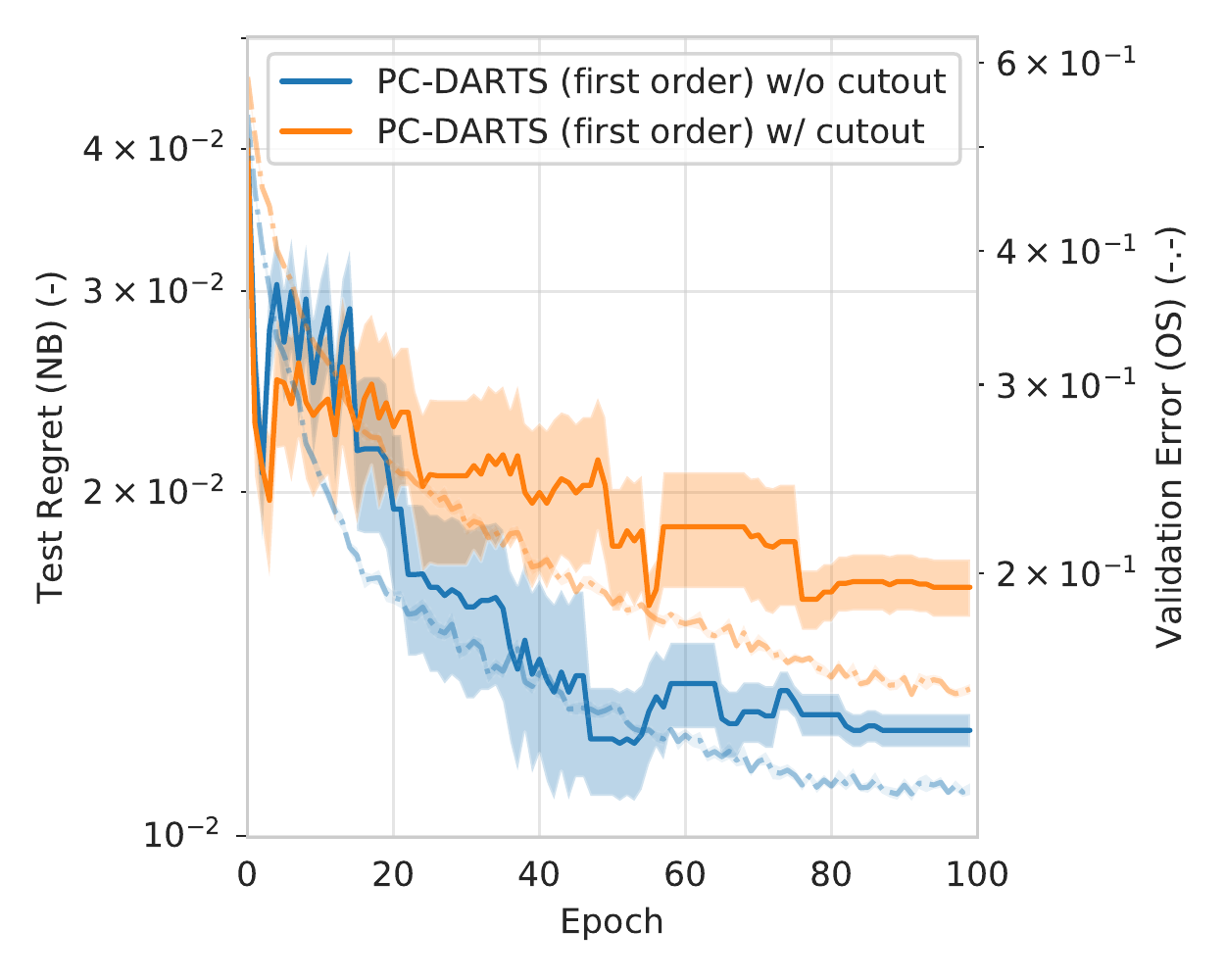}
  \caption{Search space 2}
  \label{fig:sec3:cutout_pc_darts_2}
\end{subfigure}
\caption{Comparison of the effect of using cutout during architecture search on PC-DARTS for search space 1 and 2.}
\label{fig:sec3:regularization:cutout_pc_darts}
\end{figure}

Enabling CutOut~\citep{cutout} during the architecture search has in general a negative on the quality of solutions found by one-shot NAS optimizers as shown in Figure~\ref{fig:sec3:regularization:cutout} for search space 3. The same observation also holds for search space 1 and 2 as shown in Figure~\ref{fig:sec3:regularization:cutout_gdas} and Figure~\ref{fig:sec3:regularization:cutout_pc_darts}.

\clearpage

\subsection{$L_2$ regularization}

\begin{figure}[ht]
\centering
\begin{subfigure}{.33\textwidth}
  \centering
\includegraphics[width=0.99\textwidth]{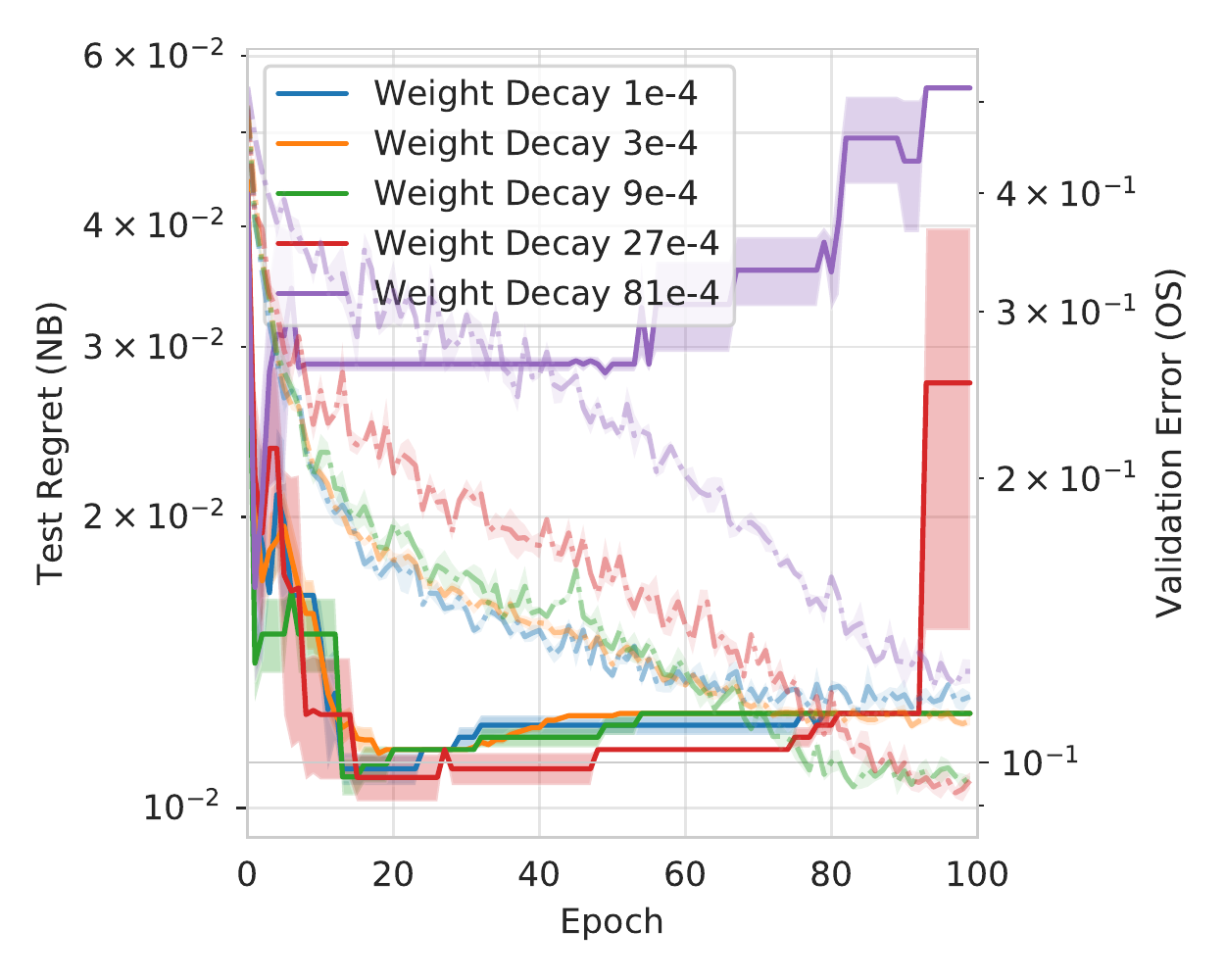}
  \caption{DARTS}
  \label{fig:sec3:ss1_l2_darts}
\end{subfigure}
\begin{subfigure}{.325\textwidth}
  \centering
\includegraphics[width=0.99\textwidth]{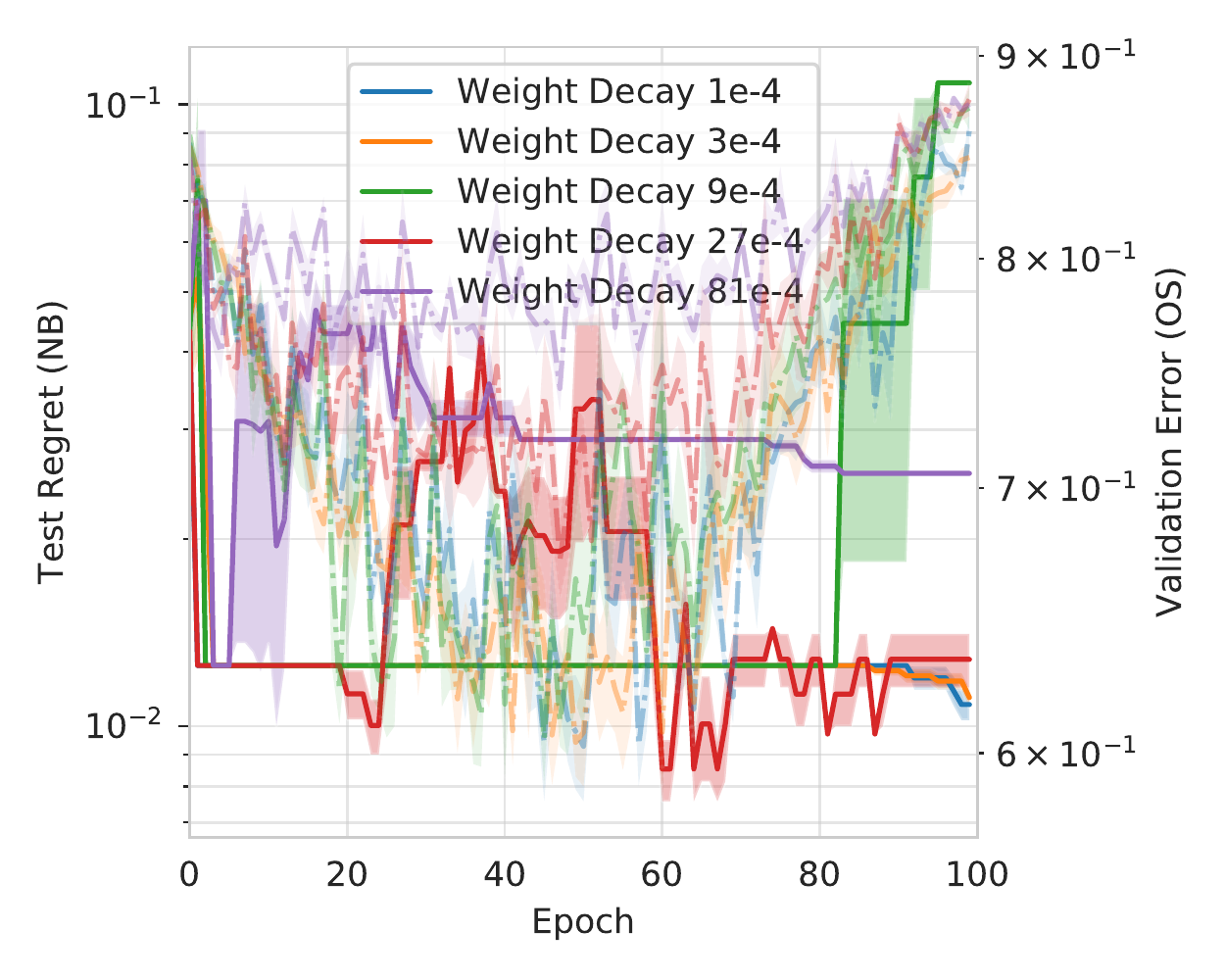}
  \caption{GDAS}
  \label{fig:sec3:ss1_l2_gdas}
\end{subfigure}
\begin{subfigure}{.33\textwidth}
  \centering
\includegraphics[width=0.99\textwidth]{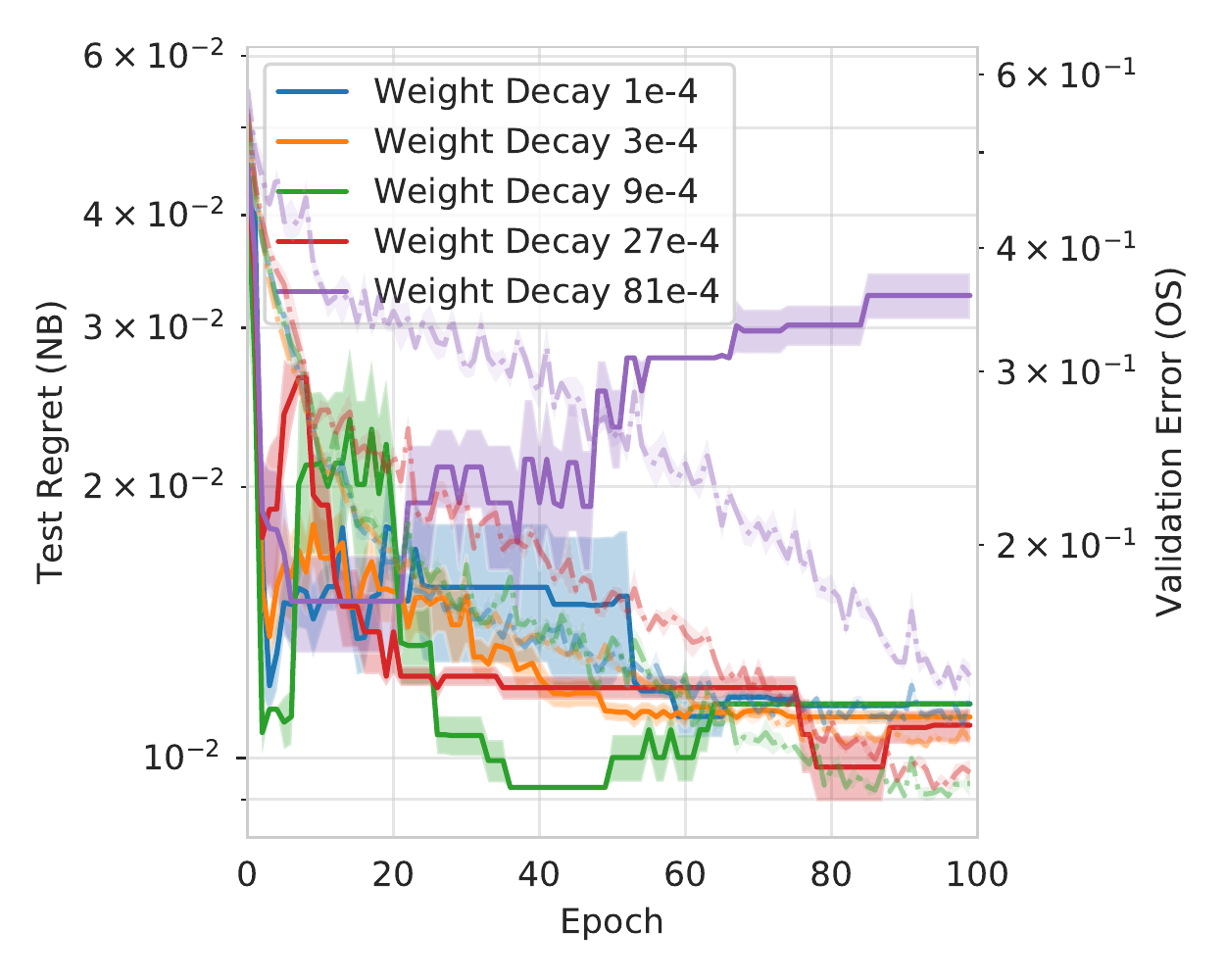}
  \caption{PC-DARTS}
  \label{fig:sec3:ss1_l2_pc_darts}
\end{subfigure}
\caption{The impact that weight decay has on the test (one-shot validation) performance of architectures found by DARTS, GDAS and PC-DARTS on search space 1 (Best viewed in color).} 
\label{fig:sec3:regularization:l2_ss1}
\end{figure}

\begin{figure}[ht]
\centering
\begin{subfigure}{.33\textwidth}
  \centering
\includegraphics[width=0.99\textwidth]{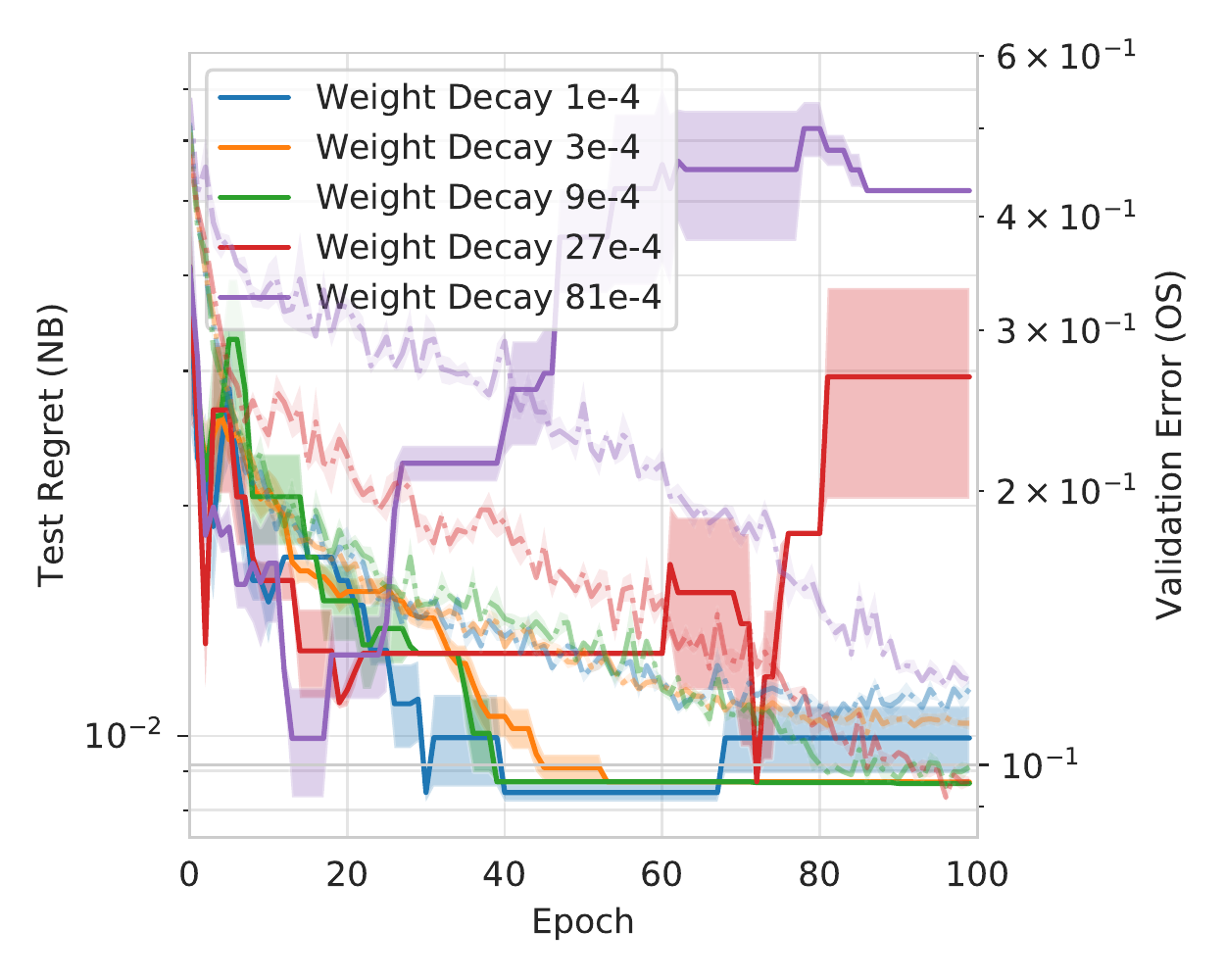}
  \caption{DARTS}
  \label{fig:sec3:ss2_l2_darts}
\end{subfigure}
\begin{subfigure}{.325\textwidth}
  \centering
\includegraphics[width=0.99\textwidth]{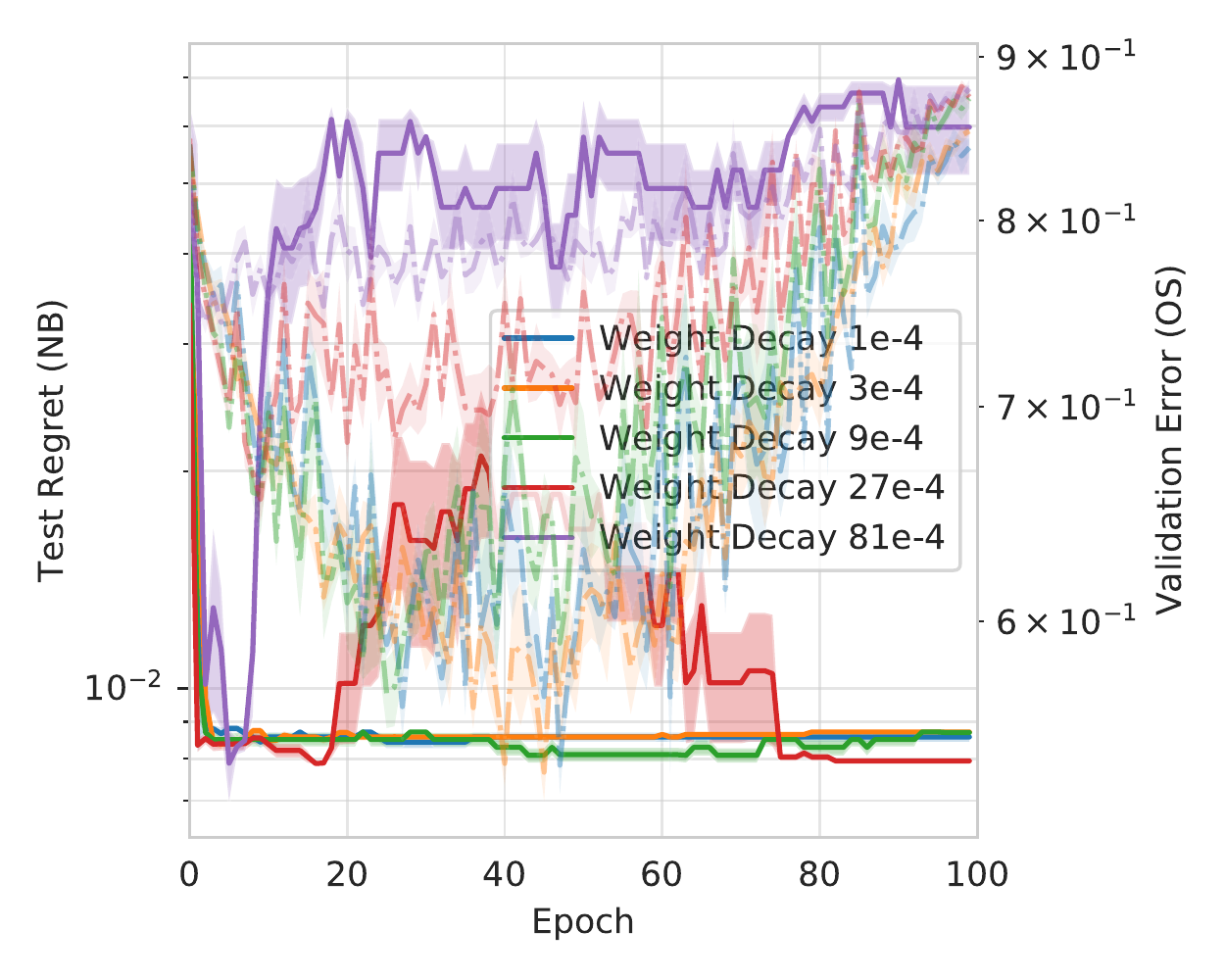}
  \caption{GDAS}
  \label{fig:sec3:ss2_l2_gdas}
\end{subfigure}
\begin{subfigure}{.33\textwidth}
  \centering
\includegraphics[width=0.99\textwidth]{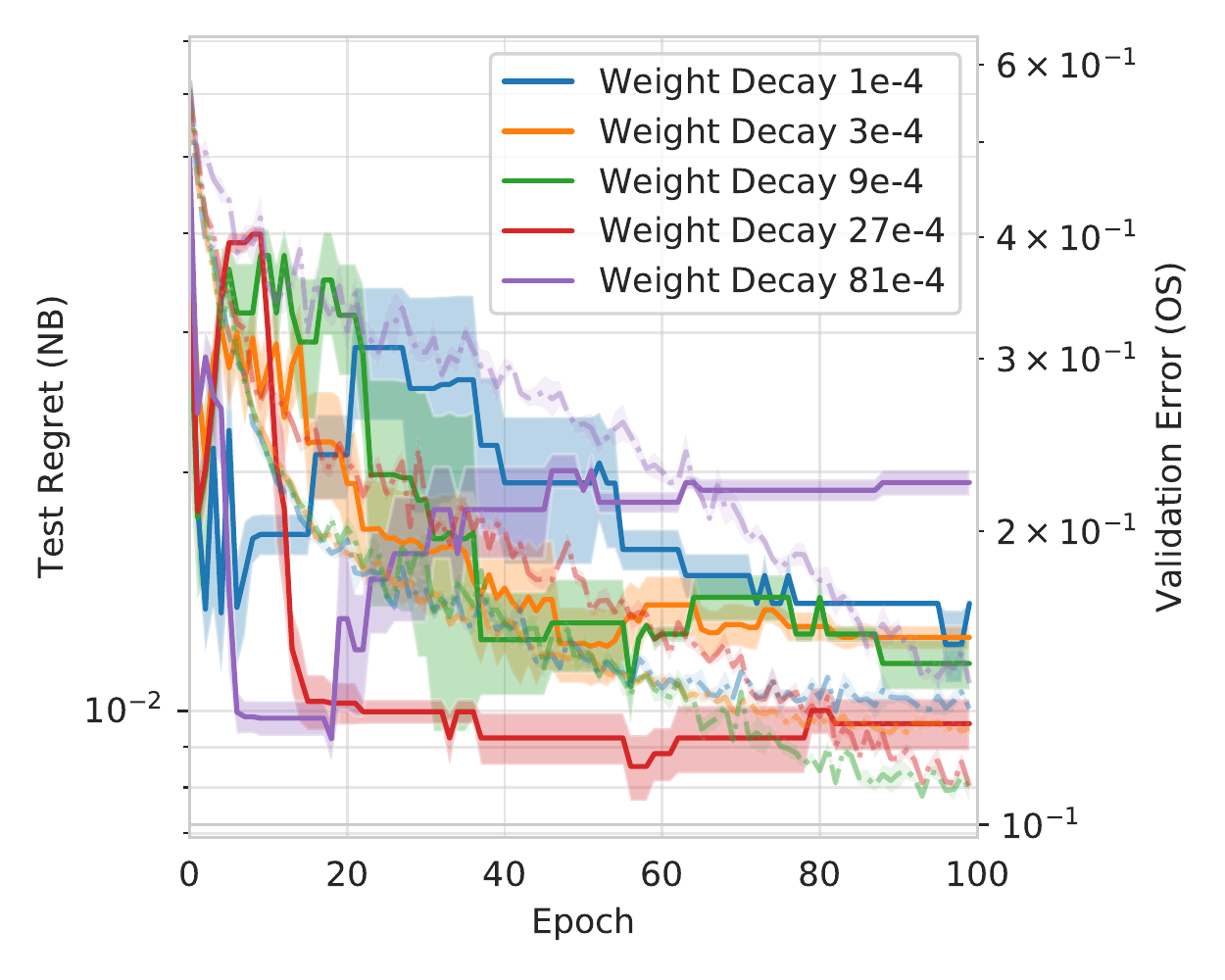}
  \caption{PC-DARTS}
  \label{fig:sec3:ss2_l2_pc_darts}
\end{subfigure}
\caption{The impact that weight decay has on the test (one-shot validation) performance of architectures found by DARTS, GDAS and PC-DARTS on search space 2 (Best viewed in color).} 
\label{fig:sec3:regularization:l2_ss2}
\end{figure}

In Figure \ref{fig:sec3:regularization:l2_ss1} and \ref{fig:sec3:regularization:l2_ss2} we can see that increasing the $L_2$ regularization has a positive effect on the found architectures for GDAS and PC-DARTS in search space 1 and 2, whilst in DARTS the quality of these architectures degenerates when increasing beyond the default in \citet{darts}.

\section{BOHB details}
\label{sec: bohb_details}

BOHB \citep{Falkner18} is a combination of Bayesian Optimization (BO) and Hyperband (HB) \citep{li_iclr17}. HB uses SuccesssiveHalving (SH) \citep{jamieson-aistats16} to stop poorly performing trainings early. SH starts trainings with an initial budget and advances the top fraction ($1/\eta$) of them to the next stage with $\eta$ higher budget. HB uses this a subroutine to evaluate many uniformly at random sampled configurations on small budgets. The budgets and scaling factors are chosen such that all SH evaluations take approximately the same time. BOHB combines HB with BO by using a probabilistic model to guide the search towards better configurations. As a result, BOHB performs as well as HB during early optimization, but samples better configurations once enough samples are available to build a model.

\subsection{Setup}

We ran BOHB for 64 iterations of \textit{SuccessiveHalving}~\citep{jamieson-aistats16} on 16 parallel workers, resulting in 280 full function evaluations.
In our experiments we use the number of epochs that the one-shot NAS optimizers run the search as the fidelity used by BOHB and optimize the validation error after 108 epochs of training queried from NAS-Bench-101.
Namely, we use $min\_budget=25$ epochs, $max\_budget=100$ epochs and $\eta=2$ in BOHB. Note that this is only the number of epochs used for the architecture search. Additionally, we never use the validation set split used to evaluate the individual architectures in NAS-Bench-101 during the architecture search. Therefore, each one-shot NAS optimizer will use 20k examples for training and 20k for search.
The x-axis in Figures \ref{fig:bohb-cs2}, \ref{fig:bohb-cs1}, \ref{fig:bohb-cs3} shows the \textit{simulated wall-clock time}: $t_{sim}=t_{search}+t_{train}$, where $t_{search}$ is the time spent during search by each NAS algorithm configuration and $t_{train}$ is the training time for 108 epochs (queried from NAS-Bench-101) of the architectures selected by the NAS optimizers.

We build 3 configuration spaces with different cardinality and which include hyperparameters affecting the architecture search process. The spaces are as follows:
\begin{enumerate}[leftmargin=*]
    \item $CS1 = \{L_2,\ CO\_prob \}$
    \item $CS2 = \{L_2,\ CO\_prob,\ lr \}$
    \item $CS3 = \{L_2,\ CO\_prob,\ lr,\ moment,\ CO\_len,\ batch\_size,\ grad\_clip,\ arch\_lr,\ arch\_L_2 \}$
\end{enumerate}

\subsection{Results} \label{bohb_results}

\begin{figure}[ht]
\centering
\begin{subfigure}{.33\textwidth}
  \centering
\includegraphics[width=0.99\textwidth]{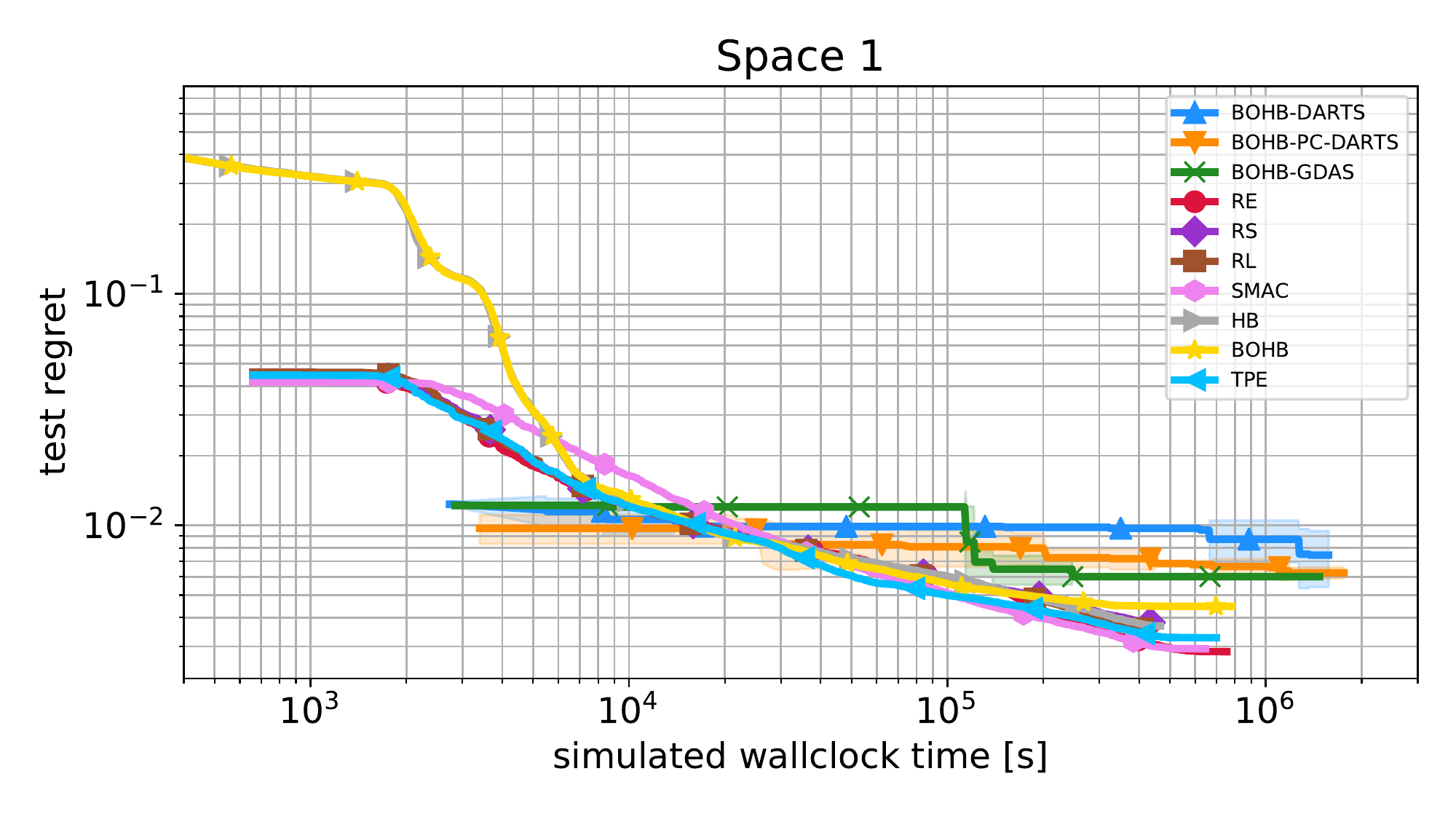}
  \caption{Search space 1}
  \label{fig:bohb-s1-cs1}
\end{subfigure}%
\begin{subfigure}{.33\textwidth}
  \centering
\includegraphics[width=0.99\textwidth]{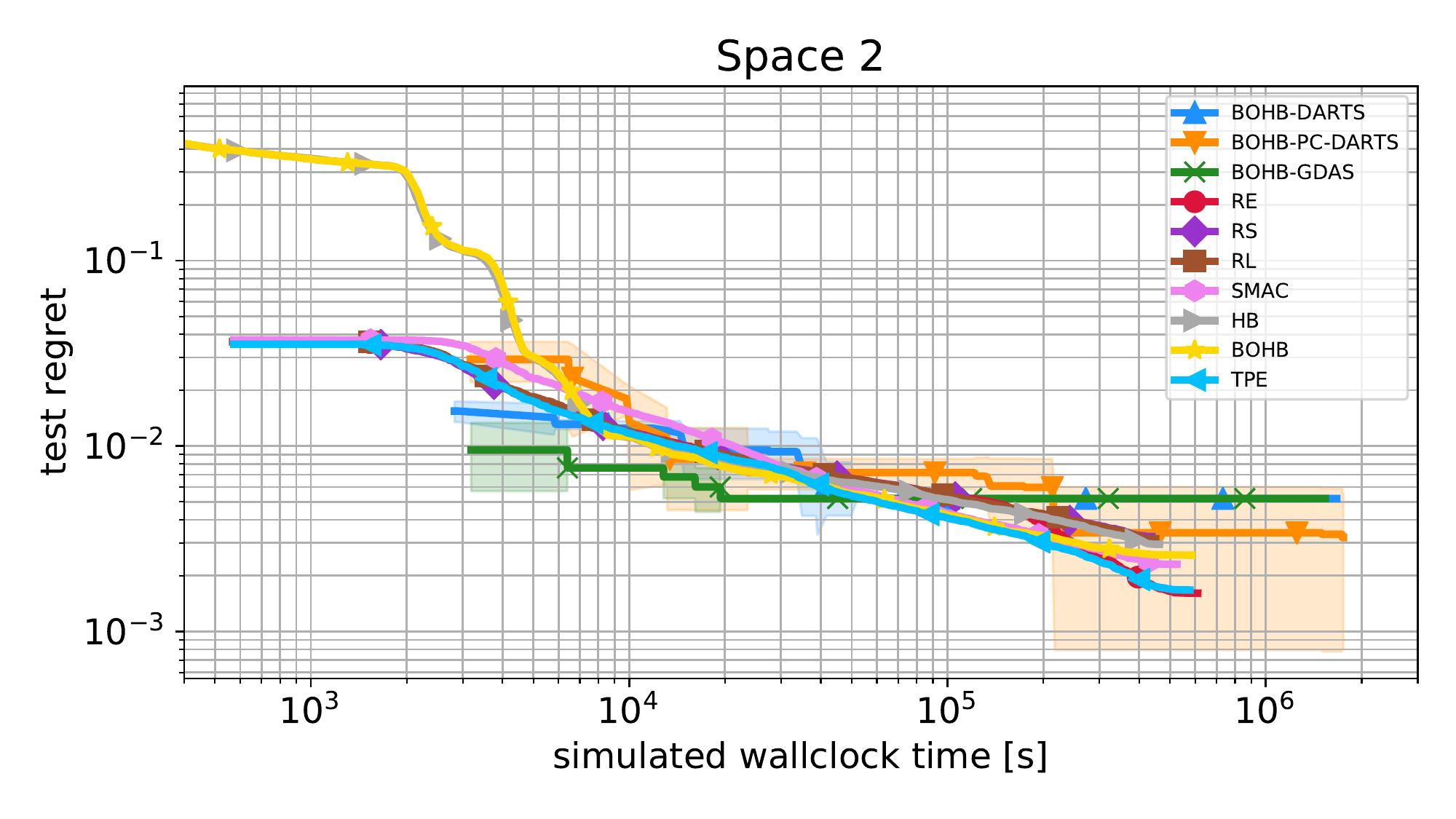}
  \caption{Search space 2}
  \label{fig:bohb-s2-cs1}
\end{subfigure}
\begin{subfigure}{.33\textwidth}
  \centering
\includegraphics[width=0.99\textwidth]{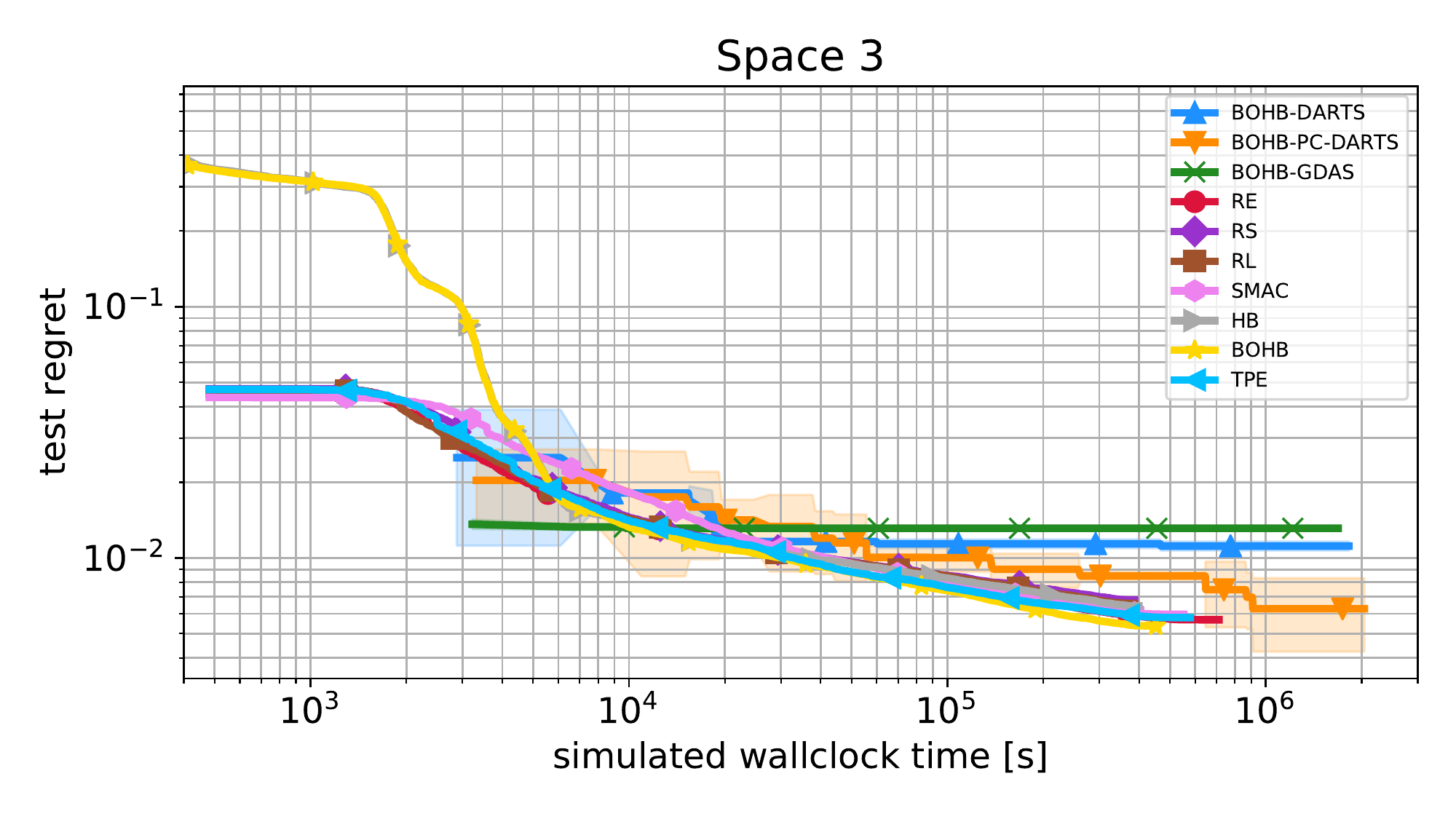}
  \caption{Search space 3}
  \label{fig:bohb-s3-cs1}
\end{subfigure}
\caption{Analogous to Figure \ref{fig:bohb-cs2}, with the only difference being that here we optimize on CS1.}
\label{fig:bohb-cs1}
\end{figure}

\begin{figure}[ht]
\centering
\begin{subfigure}{.33\textwidth}
  \centering
\includegraphics[width=0.99\textwidth]{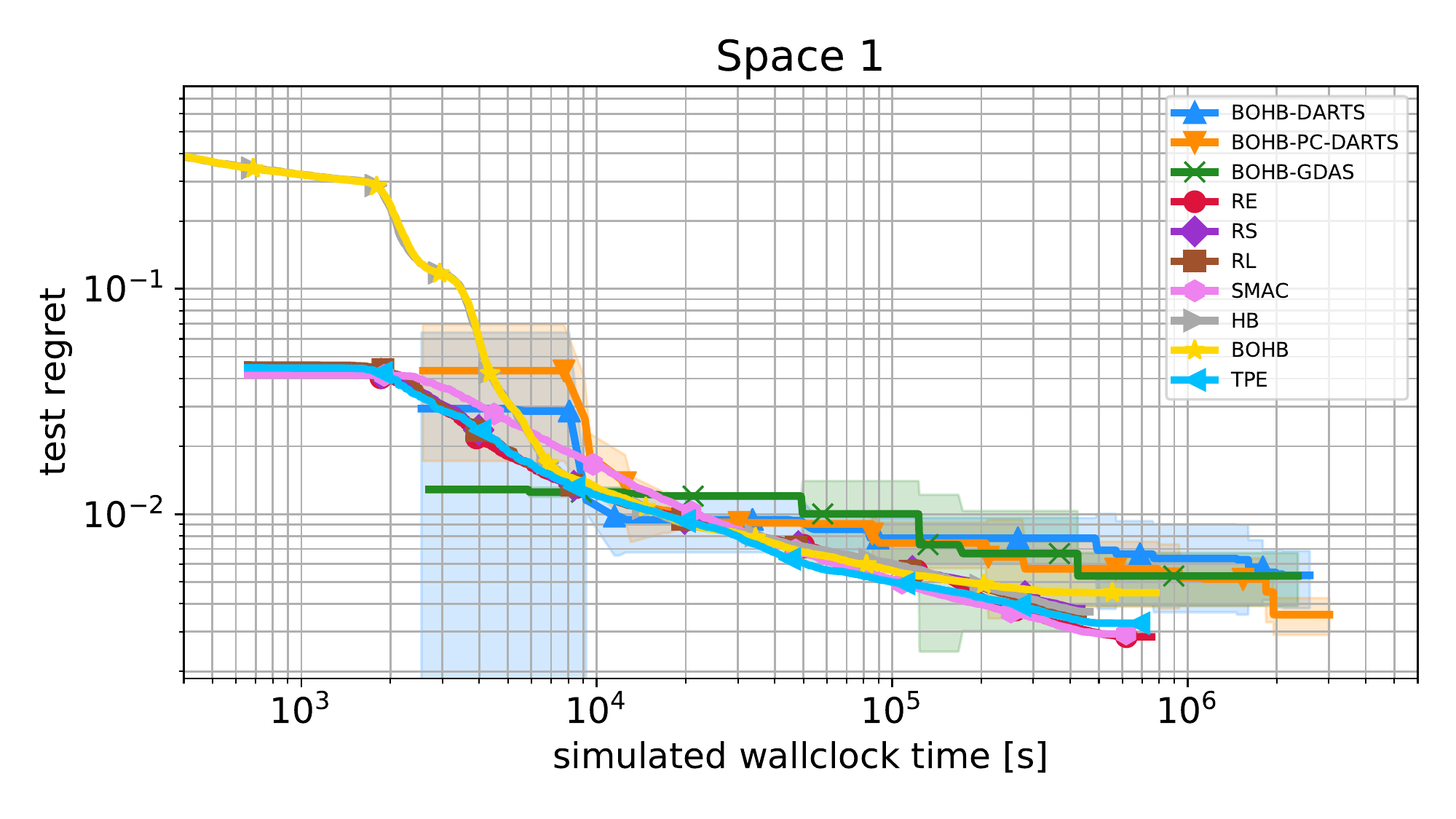}
  \caption{Search space 1}
  \label{fig:bohb-s1-cs3}
\end{subfigure}%
\begin{subfigure}{.33\textwidth}
  \centering
\includegraphics[width=0.99\textwidth]{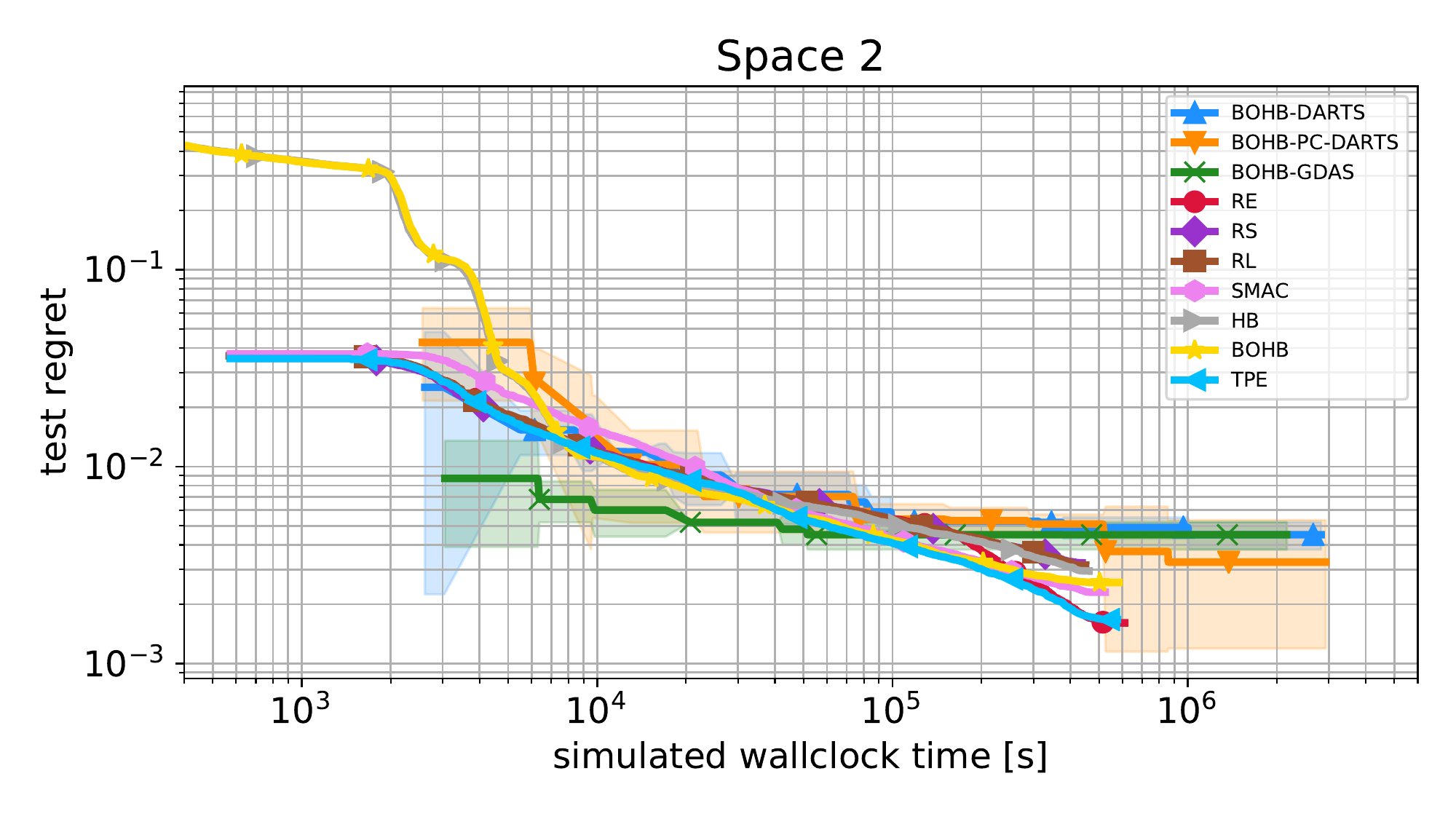}
  \caption{Search space 2}
  \label{fig:bohb-s2-cs3}
\end{subfigure}
\begin{subfigure}{.33\textwidth}
  \centering
\includegraphics[width=0.99\textwidth]{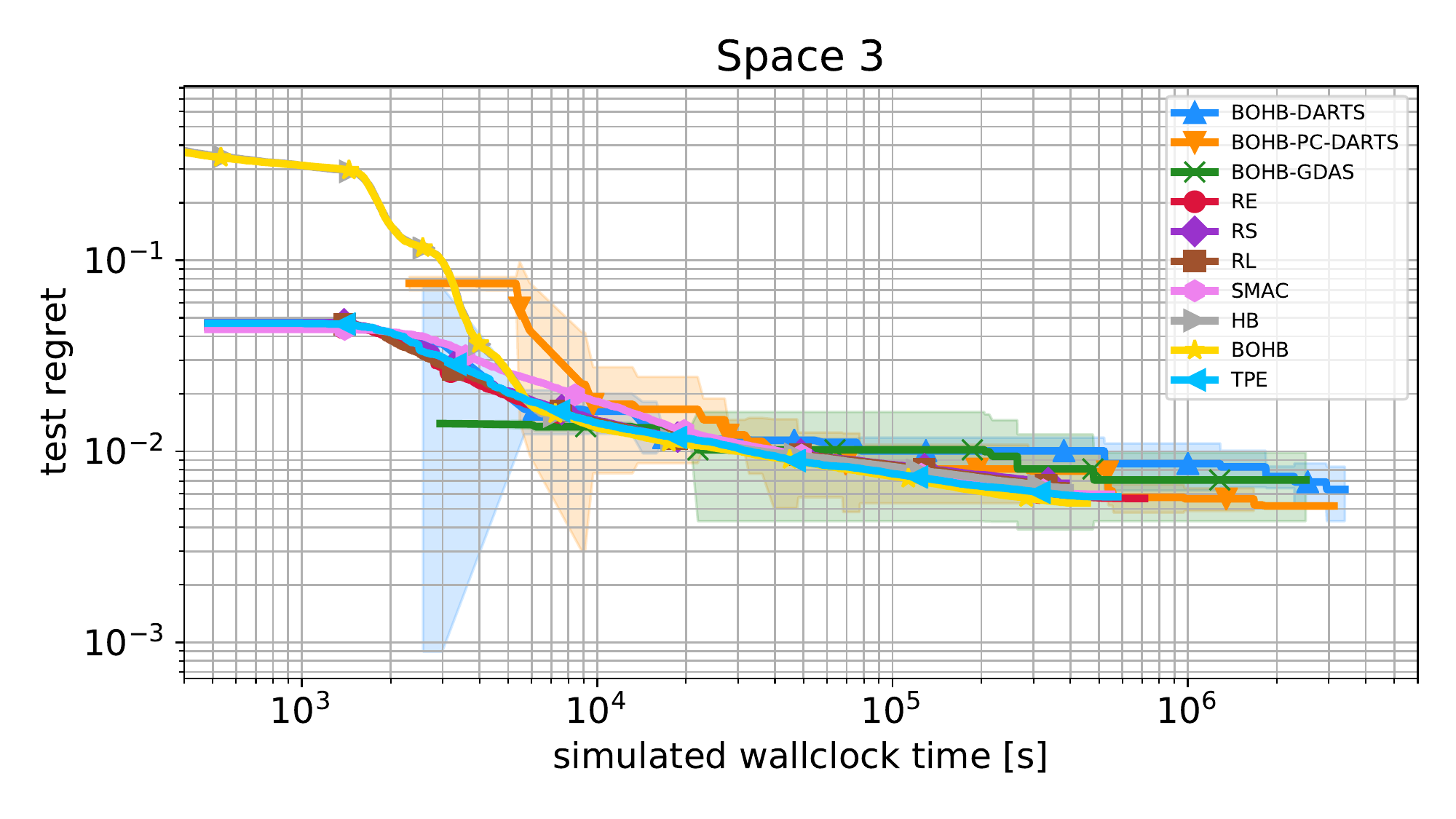}
  \caption{Search space 3}
  \label{fig:bohb-s3-cs3}
\end{subfigure}
\caption{Analogous to Figure \ref{fig:bohb-cs2}, with the only difference being that here we optimize on CS3.}
\label{fig:bohb-cs3}
\end{figure}

Interestingly, optimizing on CS2 (see Figure~\ref{fig:bohb-cs2}) led to not only a better configuration of the one-shot NAS optimizers, but also in outperforming a state-of-the-art black-box NAS optimizer such as Regularized Evolution (RE)~\citep{real-arXiv18a}. 
Including the learning rate in the configuration space was crucial to achieve such a performance. Figure \ref{fig:bohb-cs1} shows the results when running BOHB with the same settings on CS1. Note that none of the sampled configurations outperforms RE. On the other hand, increasing the cardinality of the configuration space requires many more samples to build a good model. Figure \ref{fig:bohb-cs3} shows the results when optimizing with BOHB on CS3. Even though the learning rate was inside this space, again none of the one-shot optimizers (except PC-DARTS on search space 3) with the sampled configurations is better than the discrete NAS optimizers.

\subsection{Transferability between spaces.} \label{subsec:transferability}
Table \ref{tab:transfer} shows the performance of the best found configuration on search space 3 for the 50 epochs budget by BOHB when transferred to search spaces 1 and 2. The results show the mean and standard deviation of the architectures found by 6 independent search runs with the respective optimizers and hyperparameters. We can see that there is no clear pattern on what is transferable where.

\begin{table}
\centering
\caption{Results of architectures found on search space 1 and 2 with the best found configuration for 50 epochs by BOHB on search space 3.}
\begin{tabular}{@{}llcc@{}}
\toprule
\multirow{2}{*}{\textbf{Optimizer}} &         & \multicolumn{2}{c}{\textbf{Test regret}} \\
                          &                   & Search space 1     & Search Space 2     \\ \midrule
\multirow{2}{*}{DARTS}    & Default config.   & 1.165e-2 $\pm$  0.162e-2 & 0.933e-2 $\pm$ 0.138e-2 \\
                          & Transferred config. & 1.103e-2 $\pm$ 0.220e-2 &  1.393e-2 $\pm$ 0.635e-2 \\ \midrule
\multirow{2}{*}{GDAS}     & Default config.   & 1.155e-2 $\pm$ 0.0 & 0.871e-2 $\pm$ 0.0                   \\
                          & Transferred config. & 4.619e-2 $\pm$ 4.718e-2 & 1.019e-2 $\pm$ 0.277e-2                   \\ \midrule
\multirow{2}{*}{PC-DARTS} & Default config.   & 1.246e-2 $\pm$ 0.173e-2 & 1.622e-2 $\pm$ 0.364e-2 \\
                          & Transferred config. & 1.107e-2 $\pm$ 0.129e-2 & 1.069e-2 $\pm$ 0.123e-2 \\ \bottomrule
\end{tabular}
\label{tab:transfer}
\end{table}

\section{Hyperparameter importance}
\label{sec: hyperparameter_importance}
To better understand the configuration space which we evaluated with BOHB we use functional analysis of variance (fANOVA) \citep{hutter-icml14a}. The idea is to assess the importance of individual hyperparameters by marginalizing performances over all possible values other hyperparameters could have taken. The marginalization estimates are determined by a random forest model which was trained on all configurations belonging to specific budgets during the BOHB optimization procedure.

\begin{figure}[ht]
\centering
\begin{subfigure}{.33\textwidth}
  \centering
\includegraphics[width=0.99\textwidth]{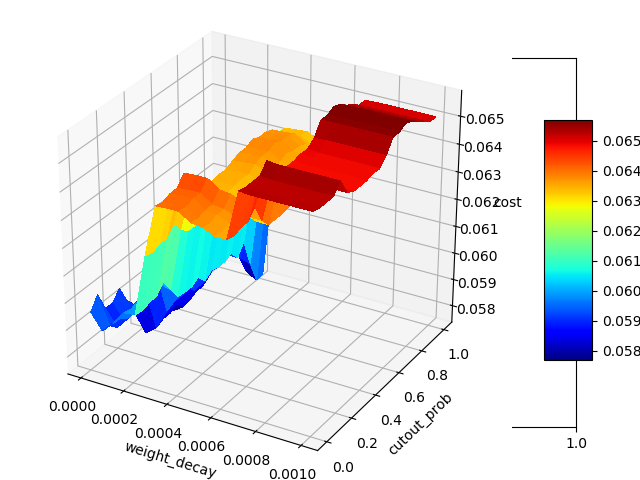}
\caption{SS 1, Budget: 25 epochs}
\includegraphics[width=0.99\textwidth]{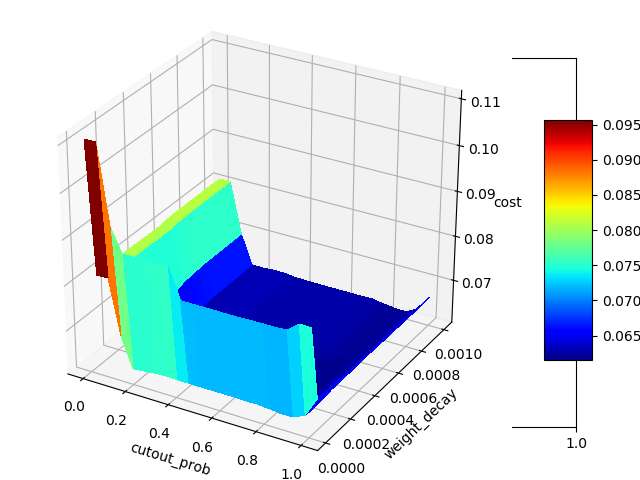}
\caption{SS 1, Budget: 50 epochs}
\includegraphics[width=0.99\textwidth]{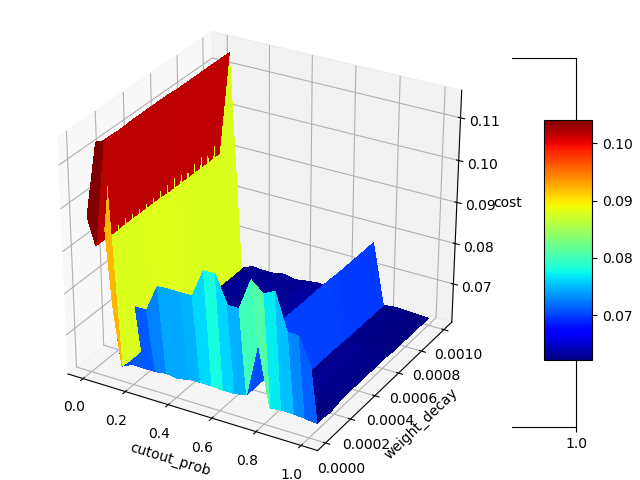}
\caption{SS 1, Budget: 100 epochs}
  \label{fig:fanova_c1_s1_1st}
\end{subfigure}%
\begin{subfigure}{.33\textwidth}
  \centering
\includegraphics[width=0.99\textwidth]{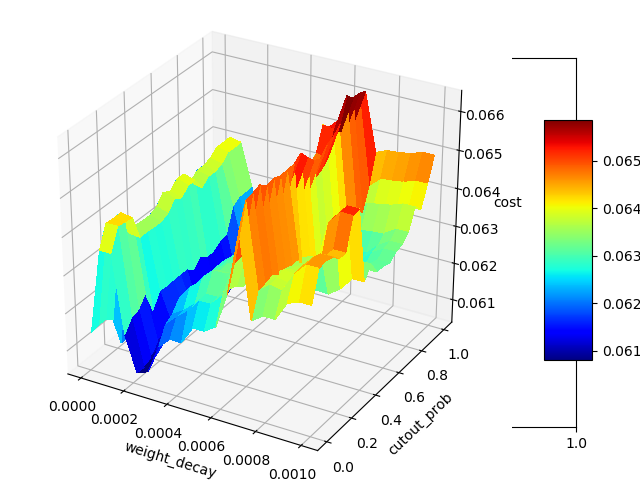}
\caption{SS 2, Budget: 25 epochs}
\includegraphics[width=0.99\textwidth]{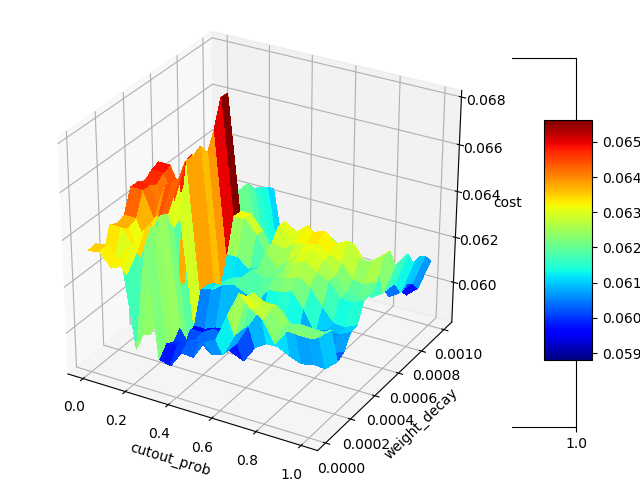}
\caption{SS 2, Budget: 50 epochs}
\includegraphics[width=0.99\textwidth]{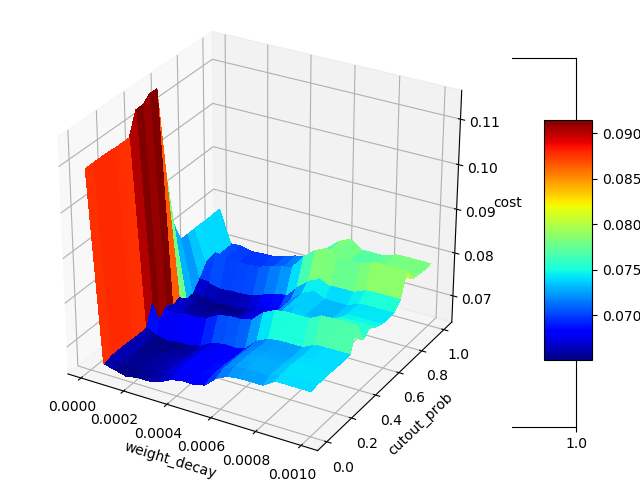} 
\caption{SS 2, Budget: 100 epochs}
  \label{fig:fanova_c1_s2_1st}
\end{subfigure}
\begin{subfigure}{.33\textwidth}
  \centering
\includegraphics[width=0.99\textwidth]{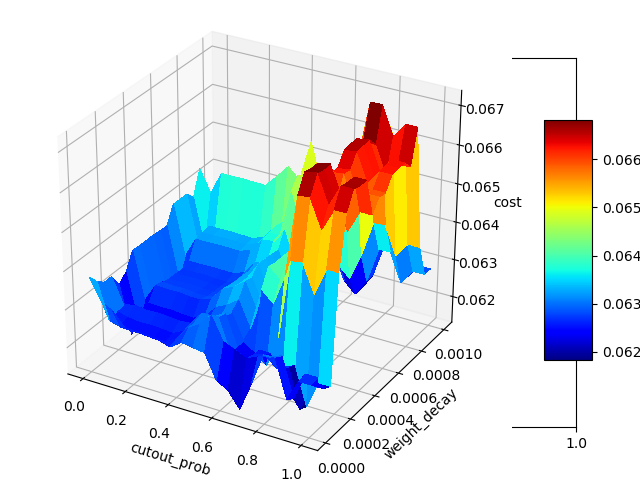}
\caption{SS 3, Budget: 25 epochs}
\includegraphics[width=0.99\textwidth]{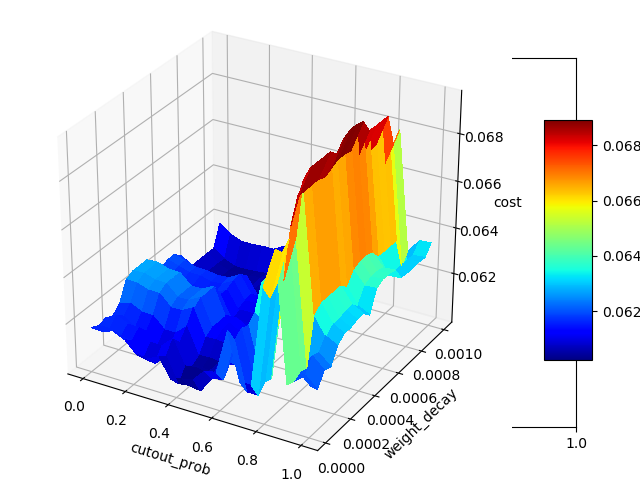}
\caption{SS 3, Budget: 50 epochs}
\includegraphics[width=0.99\textwidth]{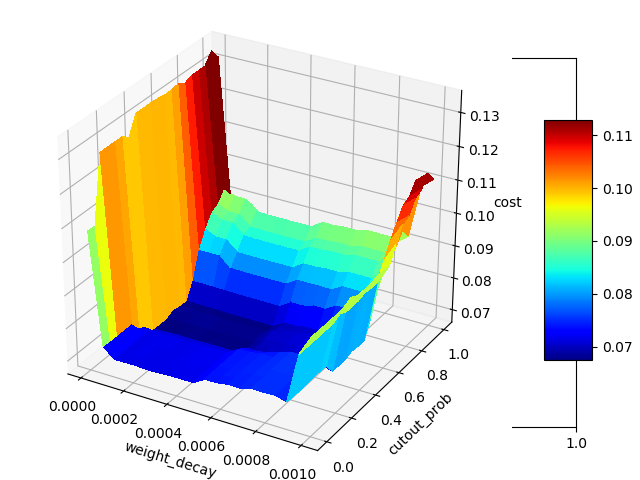}
\caption{SS 3, Budget: 100 epochs}
  \label{fig:fanova_c1_s3_1st}
\end{subfigure}
\caption{Parameter importance for two hyperparameters, Cutout (CO) and $L_2$ regularization (CS2) across different training epochs and search spaces (SS).}
\label{fig:fanova_c1_1st}
\end{figure}

Figure \ref{fig:fanova_c1_1st} shows the interaction between the Cutout (CO) and $L_2$ factor when optimizing on CS2, DARTS 1st order, for search space 1, 2 and 3. Notice that across search spaces there is some correlation between the estimated loss functions from fANOVA, indicating that this might be an interesting future direction for studying hyperparameter transfer across search spaces.

\section{Correlation between the architecture search model and the architecture evaluation model}\label{sec:correlation_proxy}
\begin{figure}[ht]
\centering
\begin{subfigure}{.33\textwidth}
  \centering
\includegraphics[width=0.99\textwidth]{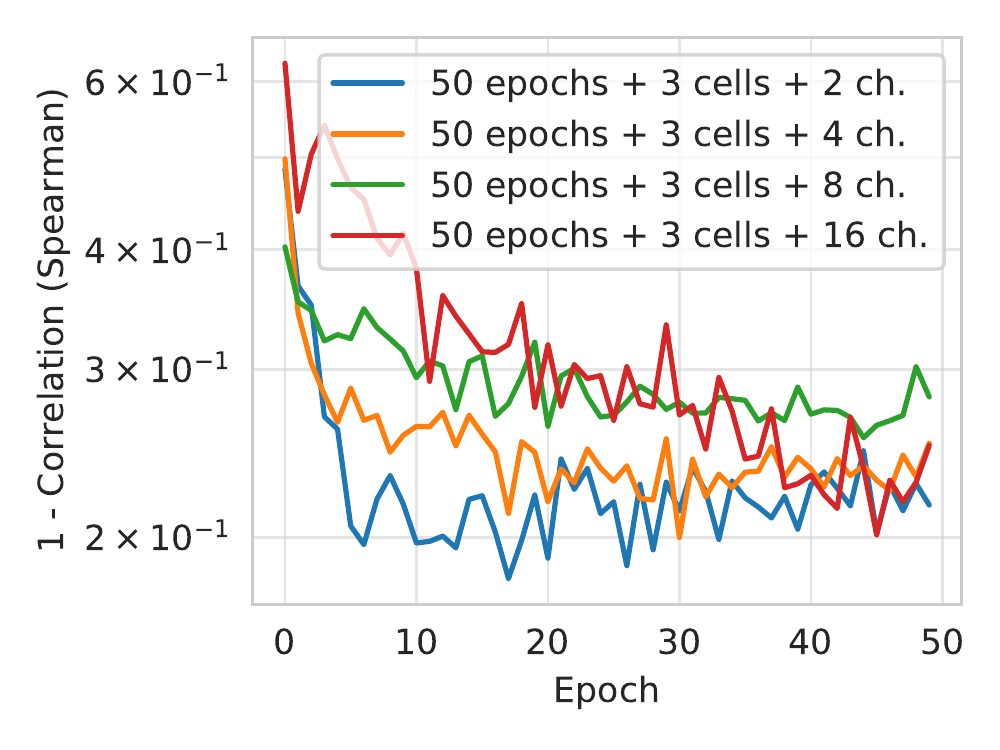}
  \caption{3 cells (1 cell per stack)}
  \label{fig:proxy_3_cells}
\end{subfigure}%
\begin{subfigure}{.33\textwidth}
  \centering
\includegraphics[width=0.99\textwidth]{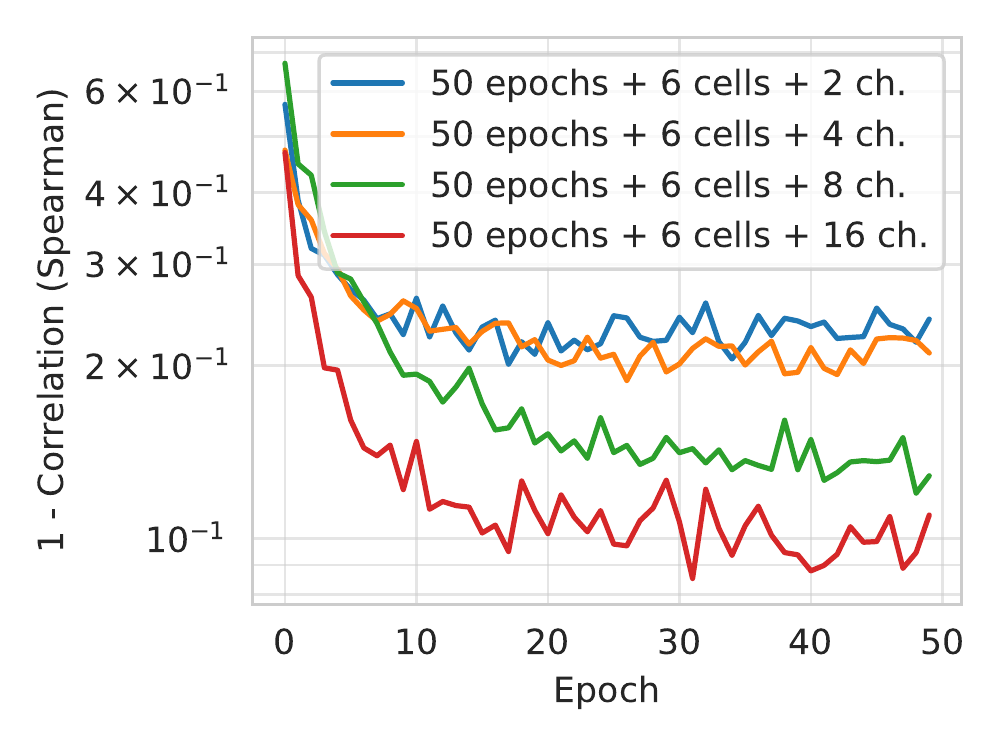}
  \caption{6 cells (2 cells per stack)}
  \label{fig:proxy_6_cells}
\end{subfigure}
\begin{subfigure}{.33\textwidth}
  \centering
\includegraphics[width=0.99\textwidth]{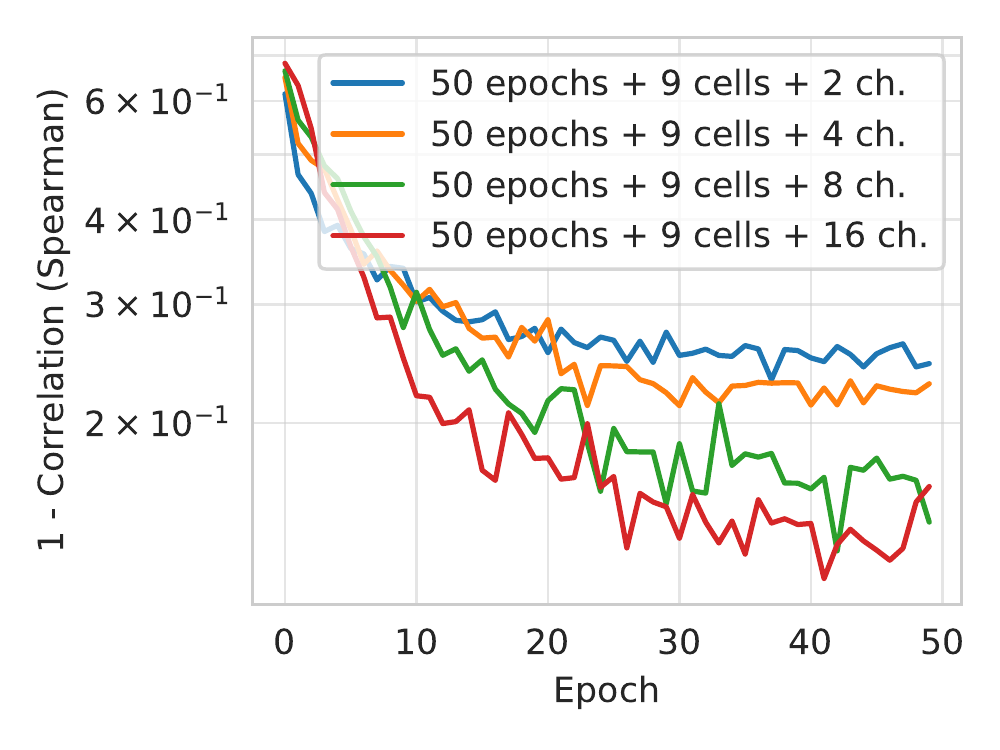}
  \caption{9 cells (3 cells per stack)}
  \label{fig:proxy_9_cells}
\end{subfigure}
\caption{In this experiment we varied the total number of cells within $[3, 6, 9]$ and the number of initial channels of the proxy model within $[2, 4, 8, 16, 36]$. }
\label{fig:proxy_full_model_correlation}
\end{figure}

As a further experiment we wanted to test how strongly the validation error of the models used in architectures search and architecture evaluation are correlated. For this we sampled 150 architectures from search space 3 and trained this architecture in the proxy model. During training we compute the Spearman rank correlation between the validation error of the proxy model and the full architecture evaluation as queried from NAS-Bench-101. The results are shown in Figure~ \ref{fig:proxy_full_model_correlation}. Note that 9 cells in the proxy was used for all of our previous experiments as it is also used by the NAS-Bench-101 models.

For 9 cells (Figure~\ref{fig:proxy_9_cells}) in the proxy model, we find that increasing the total number of channels leads to stronger correlation between the proxy model and the full architecture. However, increasing it beyond 16 channels leads to a decrease in correlation. For 3 cells (Figure~\ref{fig:proxy_3_cells}) the strongest anytime correlation was interestingly found using only 2 initial channels, with more channels leading to worse performance at the beginning and no better final performance. The results suggest that there exists a set of good combinations between the number of cells and the initial number of channels to get the maximum correlation between the search and evaluation model. As a future work we plan to investigate this relationship, which could eventually lead to a more effective bandit-based NAS method.

\end{document}